\documentclass[11pt]{article}
\usepackage[a4paper,margin=1in]{geometry}
\usepackage{amsmath,amssymb,amsthm,mathtools,bm,mathrsfs}
\usepackage{microtype,booktabs,array,enumitem,setspace}
\usepackage[numbers,square,sort&compress]{natbib}
\usepackage[colorlinks=true,citecolor=blue!55!black,linkcolor=blue!55!black,urlcolor=blue!55!black]{hyperref}
\usepackage{xcolor,graphicx,placeins}
\graphicspath{{figures/}{publication_numerics/outputs/}{outputs/}}
\makeatletter
\def\input@path{{figures/}{publication_numerics/outputs/}{outputs/}}
\makeatother
\allowdisplaybreaks[1]
\numberwithin{equation}{section}
\newtheorem{theorem}{Theorem}[section]
\newtheorem{proposition}[theorem]{Proposition}
\newtheorem{lemma}[theorem]{Lemma}
\newtheorem{corollary}[theorem]{Corollary}

\theoremstyle{definition}
\newtheorem{definition}[theorem]{Definition}
\newtheorem{example}[theorem]{Example}
\newtheorem{remark}[theorem]{Remark}

\newcommand{\R}{\mathbb R}
\newcommand{\E}{\mathbb E}
\newcommand{\Prob}{\mathbb P}
\newcommand{\Var}{\operatorname{Var}}
\newcommand{\Cov}{\operatorname{Cov}}
\newcommand{\rank}{\operatorname{rank}}
\newcommand{\tr}{\operatorname{Tr}}
\newcommand{\diag}{\operatorname{diag}}
\newcommand{\sig}{\operatorname{sigmoid}}
\newcommand{\KL}{D_{\mathrm{KL}}}
\newcommand{\CE}{\mathcal L_{\mathrm{CE}}}
\newcommand{\cN}{\mathcal N}
\newcommand{\cC}{\mathcal{C}}
\newcommand{\one}{\mathbf 1}
\newcommand{\Frob}[1]{\lVert #1\rVert_{\mathrm F}}

\newcommand{\dd}{\,\mathrm d}
\title{Sharp Prediction Regret under Information Restriction: Canonical Losses and Singular Limits}
\author{Akeel A. Shah*\\[6pt]
\small Chongqing University\\
\small Shapingba, Chongqing 400044, China\\
\small *akeelshah@cqu.edu.cn}
\date{}
\begin{document}
\maketitle
\begin{abstract}
We investigate the joint singular limit of nonlinear
approximation in which both the magnitude of unresolved
signal fluctuations and the width of a probabilistic
transition layer tend to zero. For binary prediction under
strictly proper canonical losses, we derive an exact
integrated threshold-loss identity whose minimum depends
on the threshold distribution only through its first two
moments, and characterise the loss coordinates admitting
this property. Localisation at the decision boundary
and a separate relative-error analysis establish a sharp
asymptotic equivalent for matched smooth canonical links
along arbitrary joint paths, connecting variance-controlled
and standard-deviation-controlled population regret.
The resulting explicit calibration is asymptotically
optimal without requiring convergence in distribution
of the normalised residual. An exact Wasserstein
representation of integrated regret further distinguishes
the universal total approximation law from the
distribution-dependent contributions of representation
and decoder restriction. For centred Gaussian quadratic
models, spectral alignment and conditional affine
relaxation extend the sharp approximation law to global
minimisation over positive-semidefinite rank-constrained
matrices. Under logarithmic loss we additionally
establish uniform nonasymptotic bounds and sharp
fixed-temperature asymptotics, showing that the
approximation exponent for power-law spectra is halved
in the deterministic-label limit. At finite positive
inverse temperature, the factorised canonical-loss
objective has no nonglobal local minima. We classify
its stationary matrices and determine exactly the
negative Hessian eigenvalues and their multiplicities.
The results establish how distinct approximation
rates arise from the interaction between spectral decay,
loss geometry and the singular limit of the observation
model.
\end{abstract}
\vspace{0.15in}

\section{Introduction}
The approximation error associated with a smooth
loss need not retain its quadratic dependence on
the magnitude of an unresolved signal when the
observation model approaches a discontinuous
decision rule. In this paper we study the joint
asymptotic limit in which both the magnitude of
the omitted fluctuations and the width of the
probabilistic transition layer approach zero.
Our objectives are to determine the optimal
approximation error for arbitrary relative rates
of decay of these two quantities, to identify the
properties of the loss that determine its
dependence on the omitted variance and to establish
whether the resulting sharp approximation law
remains valid under constraints on the admissible
representations. We begin with an abstract
approximation problem involving independent retained
and unresolved signal components and subsequently
extend the analysis to rank-constrained quadratic
models, for which the optimisation over admissible
representations is nonconvex.

Consider a real-valued random signal of the form
\begin{equation}
 F_\varepsilon=H_\varepsilon+\varepsilon Z_\varepsilon,
 \qquad \E\, Z_\varepsilon=0,\qquad \E\, Z_\varepsilon^2=1
\end{equation}
in which $H_\varepsilon$ is the retained signal,
$Z_\varepsilon$ is the normalised unresolved
fluctuation, assumed independent of $H_\varepsilon$,
and $\varepsilon>0$ is the standard deviation of
the unresolved component. We refer to the reference
model, which has access to the full signal
$F_\varepsilon$, as the teacher and to its
approximation based on the retained signal
$H_\varepsilon$ as the student. For a continuous,
strictly increasing distribution function $P$,
the conditional probability of the positive binary
outcome given the full signal is
$P(\tau F_\varepsilon)$, while the student assigns
probability $P(\alpha H_\varepsilon+\beta)$,
in which $\alpha>0$ and $\beta\in\R$.
Here, $P$ is the inverse canonical link,
$\alpha H_\varepsilon+\beta$ is an affine
canonical coordinate and $\tau>0$ is the
inverse temperature. The approximation error
is measured by the regret $D_P$ associated with
the strictly proper loss defined in
Section~\ref{sec:canonical}.
We consider the minimum population regret
\begin{equation}
 E^{P,P}_{\varepsilon,\tau}
 =\inf_{\alpha>0,\,\beta\in\R}
 \E\, D_P\bigl(P(\tau F_\varepsilon),
 \alpha H_\varepsilon+\beta\bigr).
\end{equation}

The two competing scales in this problem are
the standard deviation $\varepsilon$ of the
unresolved component and the width of the
teacher's probabilistic transition layer,
which is proportional to $\tau^{-1}$.
In the zero-temperature limit, the conditional
response probability approaches a deterministic
decision rule with threshold at zero.
At a fixed positive temperature, a local quadratic
expansion of the loss suggests that the regret
is proportional to the variance of the unresolved
component. Such an expansion does not necessarily remain
uniform in the limit $\tau^{-1}\to0$ and does
not determine the approximation error when
$\varepsilon$ is of the same order as, or exceeds, $\tau^{-1}$. Under the matched-link
assumptions specified below, we establish the
sharp asymptotic equivalent
\begin{equation}
 E^{P,P}_{\varepsilon,\tau}\sim
 \frac{\rho_0\tau\varepsilon^2}
 {1+\sqrt{1+\tau^2\varepsilon^2/v_P}}
\end{equation}
in which $\rho_0$ is the limiting value of the
density of the retained signal at zero and $v_P$
is the variance of the centred latent variable
whose distribution function is $P$.
This equivalent holds along arbitrary joint
paths with $\varepsilon\to0$ and $\tau\to\infty$,
yielding a leading-order regret proportional
to $\tau\varepsilon^2$ when
$\tau\varepsilon\to0$ and to $\varepsilon$
when $\tau\varepsilon\to\infty$.
Thus, the dependence of the population regret
on the magnitude of the unresolved component
changes as the conditional response becomes
increasingly deterministic.

The dependence of the approximation error on the
magnitude of the unresolved fluctuations is of
particular interest in the context of neural
scaling laws, which describe how the test losses
of certain neural networks vary with model
capacity, training data volume, computational
resources, or specific combinations of these quantities.
Empirical studies reveal a surprising power-law
dependence of the test loss, or an appropriately
normalised measure of it, on these quantities over
substantial ranges of their values
\citep{kaplan2020,hoffmann2022}.
What is surprising is not that predictive
performance improves as more resources become
available but that the dependence of the loss
on these resources often admits such a  simple
mathematical description.
Identifying the mechanisms responsible for
this behaviour is difficult due to the
interplay between the statistical properties
of the data, the approximation power of the
hypothesis class, the geometry of the loss
function and the optimisation dynamics.
The practical importance of neural scaling laws
has motivated a rapidly growing body of theoretical
work aimed at identifying their origins,
establishing the extent of their universality
and determining whether the observed power laws
describe the genuine asymptotic behaviour
or merely intermediate asymptotics.

From the perspective of classical approximation theory,
increasing model capacity can reduce the magnitude of
the unresolved component of the underlying signal.
The approximation exponent is then determined not only
by the rate at which this component decays but also by
the dependence of the error (measured by the chosen
loss) on its magnitude. A change in this latter dependence
can therefore modify the approximation exponent even
when the underlying spectral decay remains unchanged.
The present work isolates this mechanism by
studying the minimum excess population loss
over a prescribed class of predictors,
independently of statistical estimation
and optimisation errors.

Recent investigations of neural scaling laws include the
effective-frontier viewpoint \citep{zou2026}, spectral
diagnostics of training \citep{nikolaou2026} and analyses
of contextual entropy and language-like generative models
\citep{cagnetta2026,barkeshli2026}. The use of eigenvalue
tails to characterise approximation error is widespread
in kernel learning and high-dimensional statistics, and
\citet{bahri2024} analyse several mechanisms
giving rise to distinct scaling regimes and
explain why the exponent associated with
model size is not necessarily universal. The neural tangent kernel (NTK) and
lazy-training limits \citep{jacot2018,chizat2019} provide
mathematical frameworks for analysing networks in
particular training and parametrisation regimes.
The random-feature dynamical analysis of
\citet{bordelon2024} and the feature-learning theory of
\citet{bordelon2025} examine the dependence of scaling
exponents on the properties and evolution of the feature
representation. The latter study shows
that feature learning can improve scaling exponents
associated with training time and computational resources
for target functions lying outside the reproducing kernel
Hilbert space associated with the initial NTK. 

Recent studies of hierarchical multi-index
teachers and inference in extensive-width
networks identify mechanisms for the recovery
of latent features and examine how their
acquisition affects learning curves
\citep{nguyen2026,defilippis2026,wortsman2026}.
These studies highlight the dependence of scaling
behaviour on the structure of the representation
and the mechanisms by which it is learned.
The question addressed in this paper is
complementary: \emph{can different population
approximation exponents arise within a single
family of Bernoulli observation models solely
by varying the sharpness of the conditional
response, while fixing the signal spectrum,
loss function and admissible class of
predictors?}

Matrix representations of quadratic networks
have been used to investigate empirical loss
landscapes and generalisation in over-parametrised
regimes \citep{dulee2018}.
Recently, \citet{benarous2025} derive width-,
sample- and time-dependent scaling laws for
the squared prediction risk in gradient-based
learning of Gaussian quadratic teachers with
power-law coefficients, while
\citet{defilippis2025spectra} relate spectral
structure to regularisation and sample-dependent
excess risk in quadratic and diagonal networks.
Particularly relevant to our study is the
two-layer quadratic architecture of
\citet{girardin2026}, which admits a
representation in terms of a positive-semidefinite
matrix $S$ whose rank is bounded by the
network width. For centred Gaussian quadratic
forms, the population squared-loss approximation
problem reduces to minimisation of the squared
Frobenius distance between the teacher and
student matrices, whereas its finite-sample
counterpart involves nontrivial regularisation
and random matrix effects.
We retain the positive-semidefinite rank constraint
but consider binary responses and population
canonical losses, including logistic cross-entropy.
An important feature of our analysis is that
the uniform lower bounds established under
logarithmic loss apply to all measurable
predictors based on linear projections of the
Gaussian input whose dimension does not exceed
the prescribed student rank, and hence apply
\emph{a fortiori} to rank-constrained quadratic
logistic students.

In the zero-temperature limit, the probabilistic
response of a nonzero quadratic teacher converges
almost surely to the deterministic decision rule
$\one_{\{F(X)>0\}}$, in which 
$F(X)=X^\top\Lambda X-\tr\Lambda$ is the centred
quadratic signal, $\Lambda$ is a positive-semidefinite
matrix and $X$ is a standard Gaussian random vector.
The resulting decision rule is a degree-two Gaussian
polynomial-threshold function, for which bounds on
Gaussian surface area and noise sensitivity are
established in \citep{kane2010,kane2014}.
Hypercontractive estimates bound higher-order moments
of Gaussian polynomials in terms of their second
moments with dimension-independent constants
\citep{janson1997}, while \citet{carbery2001} establish
polynomial anti-concentration estimates. Whereas many
noise-sensitivity estimates provide upper bounds on
the probability that an input perturbation changes
the binary outcome, our approximation problem requires
a lower bound on the probability of sign disagreement
when the coordinates orthogonal to an arbitrary
$p$-dimensional projection are independently resampled.
Exploiting the positive-semidefinite structure
of the quadratic teacher, we establish a lower
bound that holds uniformly over all admissible
retained subspaces and relates the disagreement
probability to the standard deviation of the
quadratic signal omitted by spectral truncation
to rank $p$, rather than to an externally
prescribed isotropic noise level.

Let $Y\in\{0,1\}$ denote the binary response
variable and let $V$ denote the representation
available to the student, corresponding to
$H_\varepsilon$ in the abstract problem
introduced above. Since cross-entropy is a
strictly proper scoring rule \citep{gneiting2007},
the optimal measurable probability prediction
based on $V$ is $\Prob(Y=1\mid V)$, with
minimum population loss given by the conditional
entropy $H(Y\mid V)$.
The increase in minimum population loss due
to information restriction is therefore the
conditional mutual information between the
response and the full signal given the retained
representation. Restricting the prediction rule
to a prescribed class of decoders generally
incurs additional regret.
For logarithmic loss, the total excess population
loss relative to the full-information Bayes
rule is the expected Bernoulli Kullback-Leibler
(KL) divergence, which admits a Bregman
representation in the natural logit parameter
\citep{banerjee2005,amari2016}.
The local curvature of the log-partition function
explains the weighted quadratic approximation
at finite temperature but does not determine
the singular limit in which that curvature
concentrates near the decision boundary.
Moreover, an upper bound obtained from a
particular calibrated student does not establish
its asymptotic optimality over the entire
rank-constrained class.

The role of the decision boundary is also
central to margin-dependent statistical learning
theory \citep{tsybakov2004}, in which convergence
rates depend on the probability mass in regions
where the conditional probabilities of the
two outcomes are close to one another.
In contrast to the sample-dependent classification
rates considered in this literature, our analysis
relates to the population regret arising from
information restriction.
Under logarithmic loss, the conditional entropy
of the response given the full signal must
be subtracted from the population loss at
each finite temperature. This distinction is
particularly important in deriving sharp
asymptotics in the regular regime of the
joint limit.

The omission of a component of the underlying
signal is closely related to the omission of
covariates in nonlinear regression.
Marginalising over an independent omitted
covariate can change the regression coefficients
associated with the retained covariates in
logistic regression \citep{neuhaus1993}, and
the distinction between conditional and marginal
coefficients in longitudinal binary models
has motivated approximations for their
attenuation \citep{zeger1988}.
\citet{gasparin2025} derive an approximate
marginal logit by matching the variance of
the combined logistic and omitted-variable
noise (Equation~(9) of their paper).
In the independent additive case, this procedure
yields the same algebraic attenuation factor
as our leading-order calibration.
Our contribution is not the attenuation factor
itself but its derivation as the unique
minimiser of an exact threshold-integrated
cross-entropy functional and the proof that
the resulting calibrated predictor attains
the sharp asymptotic equivalent for the
minimum population regret throughout the
joint limit. For Gaussian quadratic prediction
models, we further establish its asymptotic
optimality over the entire positive-semidefinite
rank-constrained matrix class.
In particular, no assumption that the combined
latent fluctuations are approximately logistic
or Gaussian is required.

The construction of proper binary losses from
their Bayes risks and the representation of
regret in terms of convex divergences are
well established in statistical decision theory.
The relationship between proper scoring rules
and entropy is discussed by \citet{gneiting2007},
while \citet{reid2010,reid2011} develop the
theory of composite binary losses and examine
their connections with statistical information
and Bregman divergences.
The convex-potential formulation used in
the present work is also related to the
construction of Fenchel-Young losses
by \citet{blondel2020}.

The variational formulation developed in this
paper is closely related to earlier work on
cumulative distributional entropy.
The bounds on cumulative paired entropy under
variance constraints derived by \citet{klein2016}
include the result that the logistic distribution
maximises the cumulative Shannon entropy under
a prescribed variance (their Corollary~6) and
are related to earlier results in reliability
and uncertainty theory
\citep{drissi2008,chen2011}.
\citet{klein2020} extend this analysis to general
concave entropy generators and characterise
the extremising distributions in terms of
their quantile functions.
Our formulation involves a related variational
problem in which the threshold distribution
is prescribed and the integrated canonical
loss is minimised with respect to affine
calibration parameters.
We derive an exact identity for this integrated
loss and characterise the loss coordinates
for which it depends on the threshold
distribution only through its first two moments.
The minimum of the integrated loss coincides
with the corresponding cumulative entropy
bound, with equality between the integrated
loss and integrated Bayes risk when the
threshold distribution belongs to the
appropriate canonical location-scale family.
The application of the integrated loss identity
to the original population approximation
problem requires localisation of the loss
near the decision boundary, uniform control
of the rescaled objective on compact parameter
sets and control of the minimisers.
Furthermore, in the matched regular regime,
the leading-order terms in the population
loss and the teacher Bayes risk cancel upon
subtraction, requiring a separate analysis
to establish a sharp asymptotic equivalent
for the regret.
Neither the integrated loss identity nor
the corresponding entropy inequality alone
resolves these difficulties.

The representation of cumulative paired entropy
in terms of quantile functions
\citep{klein2016,klein2020} also establishes a
connection with optimal transport, since the
square of the one-dimensional quadratic
Wasserstein distance is the squared $L^2(0,1)$
distance between the corresponding quantile
functions \citep[Chapter~2]{santambrogio2015}.
Although scoring functions have previously been
used to construct transport costs
\citep{pesenti2024}, our result are in relation to the
representation of the integrated canonical
regret itself in terms of the squared
quadratic Wasserstein distance.
This representation allows us to quantify the
additional regret incurred by restricting the
decoder to affine canonical coordinates in
terms of the distance between the effective
threshold distribution and a location-scale
transformation of the distribution associated
with the inverse canonical link.
In the crossover and singular regimes, the
individual contributions of information loss
and decoder restriction can depend on the
distribution of the omitted component beyond
its variance, even though their sum depends
at leading order only on this variance for
given teacher and student links, inverse
temperature and limiting boundary density.

For matched teacher and student links, the
matrix formulation places the quadratic
approximation problem within the theory of
matrix nearness under Bregman divergences.
The framework of \citet{dhillon2007} and
the low-rank constructions of \citet{kulis2009}
provide a natural setting for matrix approximation
using convex potentials, while the theory of
spectral functions developed by
\citet{lewis1996} provides mathematical tools
for analysing the dependence of the objective
on the eigenvalues and eigenvectors of the
student matrix.
In our formulation, the matrix Bregman
divergence is generated by the population
canonical loss.
The principal approximation-theoretic question
is therefore how the optimal rank-constrained
Bregman approximation behaves as the conditional
response approaches a discontinuous threshold.

Attainment of the population optimum by a
spectrally aligned matrix does not, however,
resolve the behaviour of the nonconvex
factorised objective.
Existing analyses of low-rank matrix optimisation,
notably that of \citet{li2019}, establish
strict-saddle properties under restricted
conditioning and rank hypotheses, and
over-parametrisation can remove undesirable
stationary configurations in quadratic networks
\citep{dulee2018}.
The population gradient-flow analysis of
\citet{martin2024} concerns quadratic activation
and squared loss, while the extensive-width
extension of \citet{martin2026} addresses
regularised empirical gradient flow and
recovery thresholds.
\citet{gamarnik2025} relate approximate
stationarity and rank-dependent energy barriers
to optimisation and generalisation in quadratic
networks with full-rank planted weights.

Of particular relevance to the present work
is the analysis of \citet{arjevani2026}, who
study the geometry and optimisation of shallow
polynomial networks and establish extensions
of the Eckart-Young theorem for symmetric
low-rank matrix approximation, including
the characterisation of critical points and
their Hessian signatures under Gaussian-induced
norms.
Here we consider the corresponding landscape
problem for nonlinear canonical binary losses,
allowing the prescribed student rank to be
strictly smaller than the teacher rank.
At finite positive inverse temperature, the
centred Gaussian parametrisation yields an
exact identity for the gradient in inactive
spectral directions, allowing us to characterise
all stationary matrices without imposing a
restricted condition-number assumption.
Furthermore, for a given factor width, the
negative eigenvalues of the Hessian of the
factorised objective and their multiplicities
are determined entirely by the spectrum of
the teacher moment matrix and the active
spectral subspace, without requiring explicit
knowledge of the stationary coefficients.
The resulting landscape characterisation
complements the sharp approximation theory
by establishing the absence of nonglobal
local minima even when the optimal
approximation error is strictly positive.

\subsection{Summary of the contributions of this work}

Let $v_P$ denote the variance of the centred distribution
with cumulative distribution function $P$ and let $\rho_0>0$
denote the limiting value of the density of the retained
signal $H_\varepsilon$ at zero. We assume independence of
$H_\varepsilon$ and $Z_\varepsilon$, uniform integrability
of $Z_\varepsilon^2$ and uniform boundedness of the retained
densities, which converge locally uniformly near zero to
a function continuous and strictly positive there.
For matched inverse links with $P\in C^2(\R)$ and
$P'>0$, under the canonical moment assumptions of
Section~\ref{sec:canonical}, Theorem~\ref{thm:abstract-sharp}
establishes the sharp asymptotic equivalent
\begin{equation}
 E^{P,P}_{\varepsilon,\tau}\sim
 \rho_0\frac{\tau\varepsilon^2}
 {1+\sqrt{1+\tau^2\varepsilon^2/v_P}}
 \qquad (\varepsilon\to0,\ \tau\to\infty)
\end{equation}
along arbitrary joint asymptotic paths. In particular,
in the regular ($\tau\varepsilon\to0$) and singular
($\tau\varepsilon\to\infty$) regimes we obtain
\begin{equation}
 E^{P,P}_{\varepsilon,\tau}\sim
 \begin{cases}
 \rho_0\tau\varepsilon^2/2,&\tau\varepsilon\to0,\\
 \rho_0\sqrt{v_P}\,\varepsilon,&\tau\varepsilon\to\infty
 \end{cases}
\end{equation}
with an asymptotically optimal calibration throughout
the joint limit given by
$\alpha=(\tau^{-2}+\varepsilon^2/v_P)^{-1/2}$ and $\beta=0$.
Neither Gaussianity nor convergence in distribution of
$Z_\varepsilon$ is required.

We derive an exact identity for the threshold-integrated
canonical loss and characterise the loss coordinates
for which the integrated loss depends on the threshold
distribution only through its first two moments.
Furthermore, we obtain an exact representation of the
integrated canonical regret in terms of the squared
quadratic Wasserstein distance, which allows us to
characterise separately the contributions to the total
regret arising from the loss of information and the
restriction of the decoder to affine canonical coordinates.
In the crossover and singular regimes, the individual
contributions can depend on the distribution of the
unresolved fluctuations beyond their variance, even
though their sum does not at leading order for given
teacher and student links, inverse temperature and
limiting boundary density. In the matched regular regime,
even the optimal unrestricted measurable decoder incurs
the same leading-order regret as the optimal affine
canonical decoder.

For the Gaussian quadratic teacher introduced above,
we define the optimal rank-constrained population regret
\begin{equation}
 \mathcal E^P_{p,\tau}
 =\inf_{\substack{S\succeq0\\\rank S\le p}}
 \E\, D_P\bigl(P(\tau F(X)),X^\top SX-\tr S\bigr)
\end{equation}
and denote by $s_p=(2\sum_{j=p+1}^{d}\lambda_j^2)^{1/2}$
the standard deviation of the unresolved component
associated with truncation to the leading $p$ spectral
directions, in which $\lambda_j$ are the eigenvalues of
$\Lambda$, arranged in nonincreasing order.
In addition to uniform nonasymptotic bounds under
logarithmic loss, we establish the sharp asymptotic
equivalent
\begin{equation}
 \mathcal E^P_{p,\tau}\sim
 \rho_0\frac{\tau s_p^2}
 {1+\sqrt{1+\tau^2s_p^2/v_P}}
\end{equation}
under the same matched-link regularity assumptions
when the teacher spectra are truncations of a fixed
nonincreasing nonnegative square-summable sequence
with infinitely many positive eigenvalues, as
$p\to\infty$, $d=d(p)>p$ and $\tau\to\infty$.
Here $\rho_0$ is the density at zero of the limiting
quadratic teacher signal. This result establishes the
asymptotic optimality of calibrated spectral truncation
over the entire positive-semidefinite rank-constrained
matrix class without imposing proportionality between
the retained student eigenvalues and the corresponding
teacher eigenvalues. Under logarithmic loss, we also
establish the exact leading-order coefficient at fixed
positive inverse temperature. For spectra satisfying
$\lambda_j\sim Cj^{-\gamma}$ with $C>0$, $\gamma>1/2$
and $d/p\to\infty$, the exponents governing the decay
of the population regret with increasing student rank
are $2\gamma-1$ at fixed positive inverse temperature
and $\gamma-1/2$ for deterministic labels. Thus, the
approximation exponent is halved despite the underlying
teacher spectrum, student model class and loss function
remaining unchanged.

Finally, we characterise the optimisation landscape
associated with the factorisation $S=WW^\top$, in which
$W\in\R^{d\times p}$. At finite positive inverse
temperature, for continuous strictly increasing teacher
inverse links $G$ and canonical student inverse links
$P\in C^1(\R)$ with bounded $P'$, every local minimum
of the factorised objective is global and every
nonglobal stationary point is a strict saddle.
The teacher and student links need not coincide.
We classify all stationary matrices and enumerate
them in the case of a simple positive teacher spectrum.
Moreover, for a given factor width, we determine
the negative eigenvalues of the Hessian of the
factorised objective and their multiplicities explicitly
in terms of the teacher moment spectrum and the active
spectral subspace, without requiring knowledge of
the stationary coefficients. The Hessian
characterisation also covers zero teacher eigenvalues
and rank-deficient stationary students, including
the case in which the admissible student rank is
smaller than the teacher rank.

\subsection{Notation and conventions}
Probability, variance and covariance are denoted by $\Prob$, $\Var$ and $\Cov$. We write $\one_A$ for the indicator of an event $A$, $F_V$ for a distribution function and $\rho_V$ for its density when it exists. The generalised quantile is denoted by $Q_V$. The notation $\Rightarrow$ signifies convergence in distribution. The Euclidean and Frobenius norms are denoted by $\|\cdot\|_2$ and $\|\cdot\|_{\mathrm F}$. For $x\in\R$ we set $x_+=\max\{x,0\}$ and $x_-=\max\{-x,0\}$. For positive quantities $a,b$, $a\lesssim b$ means $a\le Cb$ with a constant independent of the parameters for which the estimate is uniform, and $a\asymp b$ means that both comparisons hold. The notation $a\sim b$ means $a/b\to1$ in the specified limit. Constants may change between occurrences. More specific definitions are introduced at appropriate places. 

\subsection{Organisation of the paper}
Section~\ref{sec:canonical} develops the integrated variational identities and the characterisation of moment dependence. Section~\ref{sec:abstract-approx} establishes the abstract population asymptotics, and Section~\ref{sec:decomposition} examines the representation and decoder contributions. Section~\ref{sec:model} develops the Gaussian quadratic formulation and uniform deterministic bounds, while Section~\ref{sec:temperature} treats finite-temperature bounds, exact spectral alignment and sharp global asymptotics. The factorised optimisation landscape is studied in Section~\ref{sec:landscape}. Numerical illustrations are presented in Section~\ref{sec:experiments}, and Section~\ref{sec:discussion} discusses the implications and limitations. Supporting Gaussian identities and the Fourier quadrature methodology are provided in the appendices.

\section{Integrated variational principles for canonical losses}\label{sec:canonical}
\subsection{Canonical losses and convex potentials}
We first isolate the variational mechanism underlying the spectral approximation problem. We recall the basic terminology of proper composite losses.
\begin{definition}[Strictly proper binary loss and canonical link]
For outcome losses $\ell_0,\ell_1$, the expected binary loss is $L(q,r)=(1-q)\ell_0(r)+q\ell_1(r)$. It is strictly proper if its unique minimum over predictions $r\in[0,1]$ occurs at $r=q$, for every $q\in[0,1]$. The Bayes risk is $B(q)=L(q,q)$; endpoint products with zero probability are interpreted by the convention $0\cdot\infty=0$. For a differentiable strictly concave Bayes risk, the canonical link is $-B'$ on $(0,1)$. When its range is $\R$, its inverse is a continuous strictly increasing distribution function. The centring and moment assumptions below specify the subclass used here.
\end{definition}
Let $P$ be a continuous strictly increasing distribution function on $\R$ whose associated random variable $L_P$ is centred and has variance $v_P\in(0,\infty)$. Define
\begin{equation}
 A(u)=\E\,[(u-L_P)_+],\qquad A'(u)=P(u),\qquad L_P^{\rm loss}(q,u)=A(u)-qu
\label{eq:canonical-potential}
\end{equation}
in which the superscript distinguishes the expected binary loss from the latent variable. The two outcome losses $A(u)$ and $A(u)-u=\E\,[(L_P-u)_+]$ are nonnegative. For $0<q<1$ the unique minimum over $u$ is attained at $P^{-1}(q)$, so that the corresponding Bayes risk is
\begin{equation}
 B_P(q)=A(P^{-1}(q))-qP^{-1}(q),\qquad -B_P'(q)=P^{-1}(q)
\end{equation}
with $B_P(0)=B_P(1)=0$ by continuous extension. The derivative identity can be obtained without differentiating the inverse link. Indeed, define $A^*(q)=\sup_u(qu-A(u))$; then for $q,q+h\in(0,1)$ the unique maximisers are $u_q=P^{-1}(q)$ and $u_{q+h}=P^{-1}(q+h)$. Comparison of the two variational problems yields, for $h>0$,
\begin{equation*}
 u_q\le\frac{A^*(q+h)-A^*(q)}{h}\le u_{q+h}
\end{equation*}
Continuity of $P^{-1}$ proves that
$(A^*)'(q)=P^{-1}(q)$ and since $B_P=-A^*$,
it follows that $B_P'(q)=-P^{-1}(q)$. This is the canonical proper composite loss construction \citep{reid2010,reid2011,blondel2020}. We use its nonnegative regret $D_P(q,u)=A(u)-qu-B_P(q)$ throughout. For $P=\sig$, $A(u)=\log(1+e^u)$, $v_P=\pi^2/3$, and $B_P$ is the binary Shannon entropy. The Gaussian distribution function is another admissible canonical inverse link, with $A(u)=u\Phi_{\rm N}(u)+\phi_{\rm N}(u)$ and $v_P=1$, but its corresponding loss is not ordinary logarithmic loss.

\subsection{Integrated moment identity}
\begin{definition}[Integrated canonical threshold loss]\label{def:integrated-threshold}
For the canonical potential $A$ and a threshold variable with distribution function $Q$, the integrated canonical threshold loss is
\begin{equation}
 \mathscr C_{Q,P}(b,c)=\int_\R [A(bx+c)-Q(x)(bx+c)]\dd x,
 \qquad b>0,\quad c\in\R.
\end{equation}
\end{definition}
\begin{proposition}[Integrated canonical loss]\label{prop:canonical-integrated}
Let $T$ have distribution function $Q$, mean $\mu$ and finite variance $v\geq0$. For $b>0$ and $c\in\R$,
\begin{equation}
 \mathscr C_{Q,P}(b,c)
 =\frac{v_P}{2b}+\frac{bv}{2}+\frac{(b\mu+c)^2}{2b}
\label{eq:canonical-integrated}
\end{equation}
If $v>0$, the unique minimiser is $(b,c)=(\sqrt{v_P/v},-b\mu)$ and the minimum value is $\sqrt{v_Pv}$. If $v=0$, the infimum is zero and is approached as $b\to\infty$ with $c=-b\mu$.
\end{proposition}
\begin{proof}
For a deterministic threshold $t\in\R$, define
\begin{equation*}
 J(t)=\int_\R[A(x)-x\one_{\{t\le x\}}]\dd x=\int_{-\infty}^t A(x)\dd x+\int_t^\infty[A(x)-x]\dd x
\end{equation*}
in which both integrands on the right are nonnegative since $A(x)\ge x_+$. 
To establish finiteness, first consider $t=0$. Centring of the latent variable yields $A(x)-x=\E\,[(L_P-x)_+]$, so that Tonelli's theorem applied to the two nonnegative integrands yields
\begin{equation}\label{prop23}
\begin{aligned}
 J(0)&=\E\,\left[\int_{-\infty}^0(x-L_P)_+\dd x+\int_0^\infty(L_P-x)_+\dd x\right]\\
 &=\frac12\E\,[(L_P)_-^2+(L_P)_+^2]=\frac{v_P}{2}
 \end{aligned}
\end{equation}
Changing the threshold from 0 to any finite $t$ changes only an integral over a bounded interval. Thus, $J(t)$ is finite and, by continuity of $A$, differentiation yields $J'(t)=A(t)-[A(t)-t]=t$. Integration from 0 to $t$ then gives $J(t)=(v_P+t^2)/2$, using Equation \eqref{prop23}.

For a random threshold $T$ with distribution function $Q$, Tonelli's theorem and the positive affine change of variables $y=bx+c$ yield
\begin{align*}
 \mathscr C_{Q,P}(b,c)&=\E\,\int_\R[A(bx+c)-(bx+c)\one_{\{T\le x\}}]\dd x\\
 &=b^{-1}\E\, J(bT+c)\\
 &=\frac{v_P+\E\,[(bT+c)^2]}{2b}
\end{align*}
and writing the second moment in terms of the mean and variance leads to \eqref{eq:canonical-integrated}. For each fixed $b>0$ the last term in \eqref{eq:canonical-integrated} is uniquely minimised by $c=-b\mu$. For $v>0$, the remaining term $v_P/(2b)+bv/2$ is strictly convex, diverges at both endpoints of the positive half-line and has derivative $-v_P/(2b^2)+v/2$. Its unique minimum is therefore attained at $b=\sqrt{v_P/v}$, with the value $\sqrt{v_Pv}$. For $v=0$, the remaining term is $v_P/(2b)$, whose infimum is zero as $b\to\infty$ but is not attained at any finite slope.
\end{proof}
\subsection{Cumulative entropy inequalities}
\begin{definition}[Cumulative Bayes-risk functional]\label{def:cumulative-bayes}
For a random variable $T$ with distribution function $F_T$, the cumulative Bayes-risk functional relative to $P$ is defined by
\begin{equation}
 \mathscr H_P(T)=\int_\R B_P(F_T(x))\dd x
\end{equation}
\end{definition}
Nonnegativity of the proper-loss regret and Proposition~\ref{prop:canonical-integrated} imply that
\begin{equation}
 0\le\mathscr H_P(T)\le\sqrt{v_P\Var(T)},\qquad \mathscr H_P(L_P)=v_P
\label{eq:cumulative-entropy-bound}
\end{equation}
If $\Var(T)=0$, then $T$ is almost surely constant
and its distribution function only takes  values in $\{0,1\}$. Since $B_P(0)=B_P(1)=0$, the cumulative
Bayes risk is zero. Suppose now that $\Var(T)>0$ and let $(b_*,c_*)$
denote the minimising parameters in
\eqref{eq:canonical-integrated}. By the definition
of the Bayes risk,
\begin{equation}
 B_P(F_T(x))
 \leq A(b_*x+c_*)-F_T(x)(b_*x+c_*)
\end{equation}
and integrating both sides over $\R$ yields the
desired inequality. Equality holds if and only if
\begin{equation}
 \int_{\R}D_P(F_T(x),b_*x+c_*)\,\dd x=0
\end{equation}
and since the regret is nonnegative, this requires
$D_P(F_T(x),b_*x+c_*)=0$ for a.e.\ $x\in\R$.
Strict propriety then implies that
\begin{equation}
 F_T(x)=P(b_*x+c_*)\quad \text{for a.e. }x\in\R
\end{equation}
By right continuity of the two
distribution functions, the equality holds for
every $x\in\R$. Hence equality in the cumulative
Bayes-risk inequality requires $T$ to have the
same distribution as $(L_P-c_*)/b_*$. Conversely, if $T$ has the same distribution as a
positive location-scale transformation of $L_P$,
then its distribution function is of the form
$P(b_*x+c_*)$, in which $(b_*,c_*)$ are the
minimising parameters in
\eqref{eq:canonical-integrated}. The regret
therefore vanishes identically and the cumulative
Bayes-risk inequality reduces to an equality.

This recovers the cumulative entropy extremisation
results of \citep{klein2016,klein2020}, including
the optimality of the logistic distribution for
cumulative paired Shannon entropy under a fixed
variance constraint. We include this known extremisation
result to make explicit the connection between the
integrated threshold-loss minimum and the cumulative
entropy maximum. In the subsequent analysis we use
\eqref{eq:canonical-integrated} to determine the optimal
student calibration and derive sharp population
approximation asymptotics, rather than to obtain a new
variance-constrained entropy inequality.

\subsection{Characterisation of moment-dependent losses}

The integrated loss identity \eqref{eq:canonical-integrated}
naturally leads to the converse question of which combinations of
loss and link functions yield an integrated loss that
depends on the threshold distribution only through
its first two moments. The following proposition
characterises such combinations.

\begin{proposition}[Characterisation by moment dependence]\label{prop:canonical-characterisation}
Let $\ell_0,\ell_1$ be continuous nonnegative outcome losses on $(0,1)$ and let $P$ be a continuous strictly increasing distribution function taking values in $(0,1)$, and suppose the integrated losses for deterministic thresholds are finite. The integrated loss against every finitely supported threshold distribution depends only on the first two moments if and only if $x\mapsto\ell_0(P(x))-\ell_1(P(x))$ is affine. If the losses are strictly proper with differentiable strictly concave Bayes risk $B$, this implies that $-B'\circ P$ is affine with positive slope.
\end{proposition}
\begin{proof}
For a deterministic threshold $t$ put
\begin{equation*}
 J(t)=\int_{-\infty}^t\ell_0(P(x))\dd x+\int_t^\infty\ell_1(P(x))\dd x.
\end{equation*}
By nonnegativity of the losses and Tonelli's theorem,
the integrated loss associated with a random threshold
$T$ can be written as $\E \,J(T)$. Continuity
of the losses allows us to differentiate $J$ with
respect to the threshold location, yielding
\begin{equation}
 J'(t)=\ell_0(P(t))-\ell_1(P(t))
\end{equation}
We first establish that the dependence of $\E \,J(T)$
on the distribution of $T$ only through its first two
moments is equivalent to $J$ being a polynomial of
degree at most two.

Choose four distinct real numbers $x_0,\ldots,x_3$
and consider the matrix
\begin{equation*}
 V=\begin{pmatrix}
 1&1&1&1\\
 x_0&x_1&x_2&x_3\\
 x_0^2&x_1^2&x_2^2&x_3^2
 \end{pmatrix}
\end{equation*}
which is of rank three, by the Vandermonde determinant
formula. Let $a=(a_0,\ldots,a_3)^\top$ be a nonzero
vector in $\mathrm{ker}\, V$, then the condition $Va=0$
implies that
\begin{equation*}
 \sum_{i=0}^3 a_i=0,\qquad
 \sum_{i=0}^3 a_ix_i=0,\qquad
 \sum_{i=0}^3 a_ix_i^2=0
\end{equation*}
Since $a$ is nonzero and its components sum to zero,
the positive and negative parts of $a$ have equal
strictly positive mass
\begin{equation*}
 m=\sum_{i=0}^3(a_i)_+
  =\sum_{i=0}^3(a_i)_-
\end{equation*}
and we can therefore define two probability measures
\begin{equation*}
 \nu_\pm=\frac{1}{m}
 \sum_{i=0}^3(a_i)_\pm\delta_{x_i}
\end{equation*}
in which  $\delta_{x_i}$ denotes the Dirac measure
concentrated at $x_i$. The remaining two conditions
in $Va=0$ ensure that these measures have identical
first and second moments. By the assumed dependence
of the integrated loss on these moments alone, we
must have
\begin{equation*}
 \int_{\R}J\,\dd\nu_+
 =
 \int_{\R}J\,\dd\nu_-
\end{equation*}
or equivalently
\begin{equation*}
 \sum_{i=0}^3a_iJ(x_i)=0
\end{equation*}
Thus, $(J(x_0),\ldots,J(x_3))^\top$ is
orthogonal to $\mathrm{ker}\,V$ and hence belongs
to its row space. Consequently, there exist constants
$c_0,c_1,c_2$ such that
\begin{equation*}
 J(x_i)=c_0+c_1x_i+c_2x_i^2,
 \qquad i=0,\ldots,3
\end{equation*}
Fixing the first three points uniquely determines
these coefficients through quadratic interpolation,
and since the fourth point can be chosen arbitrarily
outside this set of three points, it follows that
$J(x)=c_0+c_1x+c_2x^2$ for every $x\in\R$.
Conversely, if $J$ is a polynomial of degree at
most two, then
\begin{equation*}
 \E\, J(T)=c_0+c_1\E\, T+c_2\E\, T^2
\end{equation*}
which depends on the distribution of $T$ only
through its first two moments. Since $J$ is
continuously differentiable, this moment dependence
is therefore equivalent to an affine $J'$.

It remains to relate this condition to the Bayes
risk. For a strictly proper binary loss we have
\begin{equation*}
 B(p)\leq(1-p)\ell_0(q)+p\ell_1(q)
\end{equation*}
with equality at $p=q$. For any fixed $q\in(0,1)$,
the difference between the right-hand side and
$B(p)$ is thus a nonnegative function of $p$ that
attains its minimum at $p=q$. Differentiating
with respect to $p$ and evaluating at this interior
minimum yields
\begin{equation*}
 B'(q)=\ell_1(q)-\ell_0(q)
\end{equation*}
and hence
\begin{equation*}
 J'(t)=-B'(P(t))
\end{equation*}
The integrated loss therefore depends on the
threshold distribution only through its first two
moments if and only if $-B'\circ P$ is affine.
Finally, strict concavity of the differentiable
Bayes risk implies that $-B'$ is strictly
increasing. Since $P$ is also strictly increasing, 
$-B'\circ P$ is strictly increasing, and so 
if $-B'\circ P$ is affine, its slope must be
strictly positive.
\end{proof}
\begin{remark}
For a logarithmic loss, the requirement that the
integrated threshold loss depends on the threshold
distribution only through its first two moments
is equivalent to the condition
\begin{equation*}
 -B'(P(x))=\log\left(\frac{P(x)}{1-P(x)}\right)
 =ax+b
\end{equation*}
for some constants $a>0$ and $b\in\R$. Solving for
$P(x)$ shows that this property is satisfied only
by the location-scale logistic family. More
generally, since a strictly increasing affine
function on $\R$ is unbounded in both directions,
the canonical coordinate $-B'(q)$ must be unbounded
as $q$ approaches either endpoint of $(0,1)$.
This excludes, for example, the Brier loss, for
which the canonical coordinate is $2q-1$, and is
therefore bounded. Extending its inverse canonical
link to the whole of $\R$ by clipping does not
recover the required affine dependence on $x$.
\end{remark}

\section{Sharp asymptotics for population approximation}\label{sec:abstract-approx}
\subsection{Formulation and assumptions}
Let $H_\varepsilon$ and $Z_\varepsilon$ denote the
retained signal and the normalised unresolved
fluctuation, respectively, and suppose that they are
independent with $\E\, Z_\varepsilon=0$ and
$\E\, Z_\varepsilon^2=1$. We further assume that the
family of squared normalised unresolved fluctuations
$Z_\varepsilon^2$ is uniformly integrable, i.e.,
\begin{equation}
 \lim_{M\to\infty}\sup_\varepsilon
 \E\,\left[Z_\varepsilon^2
 \one_{\{|Z_\varepsilon|>M\}}\right]=0
\end{equation}
Let $\rho_\varepsilon$ denote the probability density
of $H_\varepsilon$. We assume that these densities are
uniformly bounded on $\R$ and converge uniformly as
$\varepsilon\to0$ in a fixed neighbourhood of zero
to a function $\rho$ that is continuous at 0 and
satisfies $\rho(0)=\rho_0>0$.

Let $L_G$ be a centred random variable, independent
of both $H_\varepsilon$ and $Z_\varepsilon$, with
cumulative distribution function $G$ and finite
positive variance $v_G$. The probability assigned
by the teacher to the positive binary outcome is
\begin{equation*}
 q=G(\tau F_\varepsilon)
  =G(\tau(H_\varepsilon+\varepsilon Z_\varepsilon))
\end{equation*}
in which $\tau>0$ is the inverse temperature.
\begin{definition}[Optimal canonical approximation error]\label{def:population-error}
The optimal population regret over affine canonical decoders is defined as
\begin{equation}
 E^{G,P}_{\varepsilon,\tau}=\inf_{\alpha>0,\beta\in\R}\E\, D_P(q,\alpha H_\varepsilon+\beta)
\label{eq:abstract-risk}
\end{equation}
\end{definition}
The assumptions on $H_\varepsilon$ relate to its behaviour near the decision boundary rather than to its global law; no distributional limit of the residuals is required. Set
\begin{equation}
 \delta=(\varepsilon^2+v_G\tau^{-2})^{1/2},\qquad T=\frac{\tau^{-1}L_G-\varepsilon Z_\varepsilon}{\delta},\qquad b=\delta\alpha,\quad c=\beta
\label{eq:abstract-threshold}
\end{equation}
then the conditional teacher probability at $H_\varepsilon=\delta x$ is $F_T(x)$, and the family $T$ is centred with unit variance and uniformly integrable squares. The latter follows from $T^2\le 2L_G^2/v_G+2Z_\varepsilon^2$ and uniform integrability of the terms on the right.

\subsection{Localisation and convergence of minimisers}
\begin{theorem}[Localisation and calibration]\label{thm:abstract-localisation}
As $\varepsilon\to0$ and $\tau\to\infty$ along arbitrary joint sequences
\begin{equation}
 E^{G,P}_{\varepsilon,\tau}=\rho_0\left[\sqrt{v_P}\delta-\tau^{-1}\mathscr H_P(L_G)\right]+o(\delta)
\label{eq:abstract-additive}
\end{equation}
The calibration $\alpha=\sqrt{v_P}/\delta$, $\beta=0$ attains this expansion. Every sequence with objective within $o(\delta)$ of the infimum satisfies $\delta\alpha\to\sqrt{v_P}$ and $\beta\to0$.
\end{theorem}
\begin{proof}
By conditioning on the retained signal and exploiting
the independence of $H_\varepsilon$ and $Z_\varepsilon$,
we obtain the conditional probability of the positive
binary outcome given $H_\varepsilon=h$ as
\begin{equation*}
 \Prob(Y=1\mid H_\varepsilon=h)
 =\E\,G(\tau(h+\varepsilon Z_\varepsilon))
\end{equation*}
in which the expectation is with respect to the
unresolved fluctuation $Z_\varepsilon$. By the
definition of the effective threshold $T$ and the
independence of $L_G$ and $Z_\varepsilon$, this
probability is $F_T(x)$ in the rescaled
coordinate $x=h/\delta$. Introducing the rescaled
calibration parameters $b=\delta\alpha$ and
$c=\beta$, conditioning on $H_\varepsilon$ and
changing variables from $h$ to $x$ yields the
expected population loss, prior to subtraction
of the teacher Bayes risk, in the form
$\delta C_\delta(b,c)$, in which
\begin{equation}
 C_\delta(b,c)=\int_\R\rho_\varepsilon(\delta x)[A(bx+c)-F_T(x)(bx+c)]\dd x
\end{equation}
which converges locally uniformly on $(0,\infty)\times\R$ to
\begin{equation}
 C_*(b,c)=\rho_0\left(\frac{v_P}{2b}+\frac b2+\frac{c^2}{2b}\right)
\label{eq:abstract-limit}
\end{equation}
To establish this convergence, we first decompose
the loss integrand into two nonnegative terms.
Since $F_T(x)=\E\,\one_{\{T\leq x\}}$, the integrand
excluding the density factor can be written as
\begin{equation*}
 \E\,\left[A(bx+c)-\one_{\{T\leq x\}}(bx+c)\right]
\end{equation*}
For a fixed realisation $T=t$, introducing
$r_P(u)=A(u)-u_+$ and using the fact that $b>0$
gives
\begin{align*}
 A(bx+c)-\one_{\{x\geq t\}}(bx+c)=r_P(bx+c)+|bx+c|\one_{I_{t,b,c}}(x)
\end{align*}
for a.e.\ $x\in\R$, in which $I_{t,b,c}=(\min\{t,-c/b\},\max\{t,-c/b\})$. The function $r_P$
is integrable, with $\int_{\R}r_P(u)\,\dd u=\frac{v_P}{2}$, as established in the proof of
Proposition~\ref{prop:canonical-integrated}.
On a compact parameter set $K\subset(0,\infty)
\times\R$, we can choose constants $b_-,b_+$ and
$A_+$ such that $0<b_-\leq b\leq b_+$ and
$|c/b|\leq A_+$ for every $(b,c)\in K$.
For $R>A_+$, the substitution $y=bx+c$ yields
\begin{equation*}
 \int_{|x|>R}r_P(bx+c)\dd x\le b_-^{-1}\int_{|y|>b_-(R-A_+)}r_P(y)\dd y
\end{equation*}
which converges to 0 uniformly on $K$ since $r_P$ is integrable. Now set $a=-c/b$ and consider the contribution of
the second term in the decomposition to the integral
over $\{|x|>R\}$. Since $|a|\leq A_+<R$, the
interval $I_{t,b,c}$ can intersect this region only
if $|t|>R$. Moreover, the integral over this region
is bounded above by the integral over the entire
interval $I_{t,b,c}$, yielding
\begin{align*}
 \int_{|x|>R}|bx+c|\one_{I_{t,b,c}}(x)\,\dd x\leq \frac{b_+}{2}(|t|+A_+)^2
 \one_{\{|t|>R\}}
\end{align*}
Taking the expectation with respect to the random
threshold $T$ yields
\begin{equation*}
 \E\,\left[\int_{|x|>R}|bx+c|
 \one_{I_{T,b,c}}(x)\,\dd x\right]
 \leq
 \frac{b_+}{2}\E\,\left[
 (|T|+A_+)^2\one_{\{|T|>R\}}\right]
\end{equation*}
which tends uniformly to 0 by virtue of uniform integrability. To see the density replacement explicitly, denote the nonnegative unweighted kernel by $K_T(x;b,c)$ and let $C$ be a uniform bound for $\rho_\varepsilon$. The difference from the integral with constant density $\rho_0$ is bounded by
\begin{align*}
 \sup_{|x|\le R}|\rho_\varepsilon(\delta x)-\rho_0|\int_{-R}^R K_T(x;b,c)\dd x+(C+\rho_0)\int_{|x|>R}K_T(x;b,c)\dd x.
\end{align*}
The integrated moment identity bounds the first integral uniformly on $K$ since $T$ is centred with a unit variance. At a fixed $R$, the density difference converges to zero, and the second term converges uniformly to 0 in the limit  $R\to\infty$. Proposition~\ref{prop:canonical-integrated} proves \eqref{eq:abstract-limit}.

The limiting objective is coercive within the positive-slope half-plane and has the unique minimiser $(\sqrt{v_P},0)$. Choose a compact rectangle in that half-plane containing this minimiser in its interior. Its boundary has a strictly positive limiting objective gap, which persists for $C_\delta$ by compact convergence. The original objectives are convex, even if assigned the value $+\infty$ outside their finite domain. More explicitly, set $z_*=(\sqrt{v_P},0)$ and let $z$ be an exterior point of the finite domain. Then there exists $0<\theta<1$ such that $z_\theta=(1-\theta)z_*+\theta z$ lies on the boundary of the rectangle. If the boundary gap relative to $z_*$ is at least $\kappa>0$, convexity yields
\begin{equation*}
 C_\delta(z_*)+\kappa\le C_\delta(z_\theta)\le(1-\theta)C_\delta(z_*)+\theta C_\delta(z)
\end{equation*}
and it  follows that $C_\delta(z)\ge C_\delta(z_*)+\kappa/\theta\ge C_\delta(z_*)+\kappa$. Thus all $o(1)$-minimising sequences for $C_\delta$ eventually belong to the rectangle, proving convergence of their parameters and of the infimum to $\rho_0\sqrt{v_P}$.

Finally the expected teacher Bayes risk is
\begin{equation}
 \frac1\tau\int_\R B_P(G(u))\E\,[\rho_\varepsilon(u/\tau-\varepsilon Z_\varepsilon)]\dd u
 =\frac{\rho_0}{\tau}\mathscr H_P(L_G)+o(\tau^{-1})
\end{equation}
For each fixed $u$, the argument of the density converges in probability to zero, by virtue of the fact that $\Prob(|\varepsilon Z_\varepsilon|>r)\le\varepsilon^2/r^2$ for every $r>0$. Local uniform density convergence and continuity at zero, therefore, provide convergence in probability of the density values to $\rho_0$. The uniform bound on the densities yields convergence of their expectations and integrability follows from \eqref{eq:cumulative-entropy-bound}. Dominated convergence and $\tau^{-1}=O(\delta)$ then complete the proof.
\end{proof}
\begin{remark}
When $\eta=\tau\varepsilon$ is bounded away from
zero, the additive expansion \eqref{eq:abstract-additive}
yields the sharp asymptotic equivalent
\begin{equation*}
 E^{G,P}_{\varepsilon,\tau}\sim
 \rho_0\varepsilon\left(
 \sqrt{v_P}\sqrt{1+v_G/\eta^2}
 -\frac{\mathscr H_P(L_G)}{\eta}
 \right)
\end{equation*}
The deterministic-label problem can be treated
directly using the same argument with
$T=-Z_\varepsilon$, $\delta=\varepsilon$ and zero
teacher Bayes risk, yielding the asymptotic
equivalent $\rho_0\sqrt{v_P}\varepsilon$. If, on the other hand, $\eta\to0$ and $G$ does not
belong to the location-scale family associated with
$P$, then
\begin{equation*}
 E^{G,P}_{\varepsilon,\tau}\sim
 \frac{\rho_0}{\tau}
 \left(\sqrt{v_Pv_G}-\mathscr H_P(L_G)\right)
\end{equation*}
In this regime the leading-order approximation error
arises from the mismatch between the teacher and
student inverse links rather than from the omission
of the unresolved fluctuations. Although subtracting
this leading-order misspecification error from the
expansion \eqref{eq:abstract-additive} formally
isolates the contribution of the unresolved
fluctuations, the remainder in the expansion need
not be asymptotically smaller than this contribution.
As a result, the expansion alone does not establish
a sharp equivalent for the error attributable to
the unresolved fluctuations in this regime.
\end{remark}

\subsection{Sharp matched-link asymptotics}
\begin{theorem}[Sharp matched canonical crossover]\label{thm:abstract-sharp}
Adopt  the assumptions of
Theorem~\ref{thm:abstract-localisation} and suppose
further that $G=P$, that $P\in C^2(\R)$ and that
$P'>0$ on $\R$. Along every joint asymptotic path
with $\varepsilon\to0$ and $\tau\to\infty$
\begin{equation}
 E^{P,P}_{\varepsilon,\tau}\sim\rho_0\frac{\tau\varepsilon^2}{1+\sqrt{1+\tau^2\varepsilon^2/v_P}}.
\label{eq:abstract-sharp}
\end{equation}
The student with $\alpha=(\tau^{-2}+\varepsilon^2/v_P)^{-1/2}$ and $\beta=0$ is asymptotically optimal in relative regret. When $\tau\varepsilon\to0$ the infimum over all measurable probability decoders based on $H_\varepsilon$ has the same leading term $\rho_0\tau\varepsilon^2/2$.
\end{theorem}
\begin{proof}
Theorem~\ref{thm:abstract-localisation} and $\mathscr H_P(L_P)=v_P$ prove the assertion on subsequences with $\eta=\tau\varepsilon$ approaching a positive or infinite limit. For $\eta\to0$ set $u=\tau H_\varepsilon$, $a_\eta=(1+\eta^2/v_P)^{-1/2}$ and let
\begin{equation}
 k_\eta(u)=\E\, D_P(P(u+\eta Z_\varepsilon),a_\eta u)
\end{equation}
The averaged teacher probability
$\E\, P(u+\eta Z_\varepsilon)$ is the distribution
function of $L_P-\eta Z_\varepsilon$ evaluated at
$u$. By independence, this random variable has
zero mean and variance $v_P+\eta^2$.
Proposition~\ref{prop:canonical-integrated}
therefore supplies $a_\eta$ as the optimal integrated
slope, with the corresponding integrated expected
loss given by
\begin{equation*}
 \int_\R
 \left[A(a_\eta u)-a_\eta u\,
 \E\, P(u+\eta Z_\varepsilon)\right]\dd u
 =\sqrt{v_P(v_P+\eta^2)}
\end{equation*}
The integrated teacher Bayes risk, on the other
hand, can be evaluated using Tonelli's theorem
and translation invariance to yield
\begin{equation*}
 \int_\R\E\,B_P(P(u+\eta Z_\varepsilon))\dd u
 =\int_\R B_P(P(u))\dd u=v_P
\end{equation*}
Subtracting the integrated teacher Bayes risk
from the integrated expected loss of the
calibrated student then gives
\begin{equation}
 \int_\R k_\eta(u)\dd u
 =\sqrt{v_P(v_P+\eta^2)}-v_P
 =\eta^2/2+O(\eta^4)
 \label{eq:regular-kernel-mass}
\end{equation}
in which the final equality follows from a Taylor
expansion as $\eta\to0$.

To obtain a lower bound that is valid for every measurable decoder, set $q_u=P(u+\eta Z_\varepsilon)$ and $j_\eta(u)=B_P(\E\, q_u)-\E\, B_P(q_u)$. Strict propriety and linearity in the teacher probability imply that $j_\eta$ is the minimum conditional regret. Let $A_M=\{|Z_\varepsilon|\le M\}$, $a_M=\Prob(A_M)$ and, further, let $m_1,m_2$ denote the conditional means of $q_u$ on $A_M$ and its complement, respectively. If both events have positive probability, concavity yields $B_P(a_Mm_1+(1-a_M)m_2)\ge a_MB_P(m_1)+(1-a_M)B_P(m_2)$. Subtracting the expectation of $B_P(q_u)$ on the two events yields
\begin{align*}
 j_\eta(u)\ge a_M[B_P(m_1)-\E\,(B_P(q_u)\mid A_M)] 
 +(1-a_M)[B_P(m_2)-\E\,(B_P(q_u)\mid A_M^c)]
\end{align*}
The second bracketed term is nonnegative by Jensen's inequality and may therefore be discarded. When $a_M=1$ the first bracketed terms equals the full gap. For fixed values of $M$ and $R$, a Taylor expansion on $|u|\le R$ yields,  uniformly
\begin{equation}
 j_\eta(u)\ge\frac{\eta^2}{2}P'(u)\Prob(A_M)\Var(Z_\varepsilon\mid A_M)+o_{M,R}(\eta^2)
\label{eq:regular-jensen}
\end{equation}
Indeed, differentiating $-B_P'(P(u))=u$ yields $-B_P''(P(u))=1/P'(u)$. For $|u|\le R$ and $|z|\le M$, the Taylor expansion $P(u+\eta z)=P(u)+\eta P'(u)z+O_{M,R}(\eta^2)$ holds uniformly. Let $\bar q_{u,M}=\E\,[q_u\mid A_M]$ denote the
conditional mean of $q_u$. Expanding $B_P(q_u)$
about $\bar q_{u,M}$ and taking the expectation
conditional on $A_M$, the first-order term
vanishes since $\E\,[q_u-\bar q_{u,M}\mid A_M]=0$.
Thus, the conditional Jensen gap satisfies
\begin{align*}
 B_P(\bar q_{u,M})-\E\,[B_P(q_u)\mid A_M]
=-\frac12 B_P''(\bar q_{u,M})
 \Var(q_u\mid A_M)+o_{M,R}(\eta^2)
\end{align*}
Since $\bar q_{u,M}=P(u)+O_{M,R}(\eta)$ and
$\Var(q_u\mid A_M)=O_{M,R}(\eta^2)$,
continuity of $B_P''$ yields
\begin{align*}
 B_P(\bar q_{u,M})-\E\,[B_P(q_u)\mid A_M]=-\frac12 B_P''(P(u))
 \Var(q_u\mid A_M)+o_{M,R}(\eta^2)
\end{align*}
The Taylor expansion of $P(u+\eta Z_\varepsilon)$
further implies that
\begin{equation*}
 \Var(q_u\mid A_M)
 =\eta^2P'(u)^2\Var(Z_\varepsilon\mid A_M)
 +o_{M,R}(\eta^2)
\end{equation*}
The Taylor remainders are uniform since all the probabilities in question lie in a compact subset of $(0,1)$ for small $\eta$. For sufficiently large $M$, $\Prob(A_M)$ is bounded away from zero. Moreover
\begin{equation*}
\Prob(A_M)\Var(Z_\varepsilon\mid A_M)
=\E\,[Z_\varepsilon^2\one_{A_M}]-\frac{(\E\,[Z_\varepsilon\one_{A_M}])^2}{\Prob(A_M)}
\end{equation*}
in which first term converges uniformly to 1 by uniform integrability, and the second converges uniformly to 0 by centredness and Cauchy-Schwarz on the complementary event.

Since $j_\eta$ is nonnegative, the minimum population
regret over all measurable probability predictions
based on $H_\varepsilon$ is bounded below by
\begin{equation*}
 \frac{1}{\tau}\int_{-R}^R
 \rho_\varepsilon(u/\tau)j_\eta(u)\dd u
\end{equation*}
Multiplying \eqref{eq:regular-jensen} by
$\rho_\varepsilon(u/\tau)$, integrating over $[-R,R]$
and using the uniform convergence of
$\rho_\varepsilon(u/\tau)$ to $\rho_0$ on this
interval results in
\begin{align*}
 \liminf_{\substack{\varepsilon\to0,\ \tau\to\infty\\
                     \eta\to0}}
 \frac{1}{\eta^2}\int_{-R}^R
 \rho_\varepsilon(u/\tau)j_\eta(u)\dd u\geq
 \frac{\rho_0}{2}
 \left(\int_{-R}^R P'(u)\dd u\right)
 \liminf_{\varepsilon\to0}
 \left[\Prob(A_M)\Var(Z_\varepsilon\mid A_M)\right]
\end{align*}
Now take the limit as $M\to\infty$, followed by the limit  $R\to\infty$, and recall that
$\int_\R P'(u)\dd u=1$; then  the lower
limit of the unrestricted population regret
normalised by $\eta^2/\tau$ is at least $\rho_0/2$.

To obtain anupper bound, we now consider
the regret of the calibrated student. Since
$k_\eta\geq j_\eta\geq0$, the preceding local lower
bound, applied without the density factor, also yields
\begin{equation*}
 \liminf_{\eta\to0}\frac{1}{\eta^2}
 \int_{-R}^R k_\eta(u)\dd u
 \geq\frac{1}{2}\int_{-R}^R P'(u)\dd u
\end{equation*}
On the other hand, \eqref{eq:regular-kernel-mass}
implies that
\begin{equation*}
 \lim_{\eta\to0}\frac{1}{\eta^2}
 \int_\R k_\eta(u)\dd u=\frac{1}{2}
\end{equation*}
Subtracting the contribution from $[-R,R]$ from
that over $\R$ therefore yields
\begin{equation*}
 \limsup_{\eta\to0}\frac{1}{\eta^2}
 \int_{|u|>R}k_\eta(u)\dd u
 \leq\frac{1}{2}\left(1-\int_{-R}^R P'(u)\dd u\right)
\end{equation*}
and hence
\begin{equation}
 \lim_{R\to\infty}\limsup_{\eta\to0}
 \frac{1}{\eta^2}\int_{|u|>R}k_\eta(u)\dd u=0
\end{equation}
The contribution from outside $[-R,R]$ to the
normalised integral can thus be made arbitrarily
small by choosing $R$ sufficiently large. Combining
this tail estimate with the local uniform convergence
and uniform boundedness of the densities gives
\begin{equation*}
 \frac{1}{\eta^2}\int_\R
 \rho_\varepsilon(u/\tau)k_\eta(u)\dd u
 \longrightarrow\frac{\rho_0}{2}
\end{equation*}
The population regret of the calibrated student
is therefore asymptotic to $\rho_0\eta^2/(2\tau)$,
which matches the lower bound established for
all measurable probability predictions based on
the retained signal. This completes the proof
for the regular regime.

It remains to establish \eqref{eq:abstract-sharp}
along arbitrary joint asymptotic paths. Suppose instead that the ratio of the two sides
of \eqref{eq:abstract-sharp} does not converge
to one. There then exists a subsequence along
which the absolute difference between this ratio
and 1 is bounded below by a strictly positive
constant. By compactness of the extended half-line
$[0,\infty]$, we can extract a further subsequence
along which $\eta$ converges to 0, $\infty$ or a finite
positive constant. Since convergence
of the ratio to 1 has already been established
in each of these three regimes, we obtain a
contradiction.
\end{proof}
\begin{remark}
The dependence of the crossover law
\eqref{eq:abstract-sharp} on the choice of canonical
loss can be eliminated by an appropriate
normalisation. Introducing the dimensionless
parameter $\xi=\tau\varepsilon/\sqrt{v_P}$, we obtain
\begin{equation*}
 \frac{E^{P,P}_{\varepsilon,\tau}}
 {\rho_0\varepsilon\sqrt{v_P}}
 \sim\frac{\xi}{1+\sqrt{1+\xi^2}}
\end{equation*}
The functional form of the normalised crossover law
is therefore universal within the class of canonical
losses considered here. In particular, the leading-order
regret is independent of $v_P$ in the regular regime,
while in the singular regime its dependence on the
choice of loss enters through the factor $\sqrt{v_P}$.
The optimality established in the theorem relates to
affine canonical coordinates based on a prescribed
retained signal, with the stronger optimality over
all measurable probability decoders holding in the
regular regime. The optimisation over admissible
representations is  subsequently addressed in the
context of Gaussian quadratic models.
\end{remark}

\section{Representation errors and optimal transport}\label{sec:decomposition}
\subsection{Conditional Bayes-risk decomposition}
The sharp equivalent \eqref{eq:abstract-sharp} depends on the variance of the unresolved component but not on its distribution, but this does not imply that the information lost by discarding this component is independent of its distribution. To examine this distinction, we separate the regret incurred by restricting the information available to the student from that incurred by restricting its canonical coordinate to an affine function of the retained signal.
\begin{definition}[Representation and decoder errors]\label{def:representation-decoder}
For $q=G(\tau F_\varepsilon)$, let $m(h)$ denote the conditional probability of the positive binary outcome given $H_\varepsilon=h$
\begin{equation*}
 m(h)=\E\,[G(\tau(h+\varepsilon Z_\varepsilon))]
\end{equation*}
in which the expectation is with respect to $Z_\varepsilon$. The representation and decoder errors are, respectively,
\begin{align}
 I_P&=\E\,[B_P(m(H_\varepsilon))-B_P(q)]\\
 D_P^{\rm dec}&=\inf_{\substack{\alpha>0\\\beta\in\R}}\E\, D_P(m(H_\varepsilon),\alpha H_\varepsilon+\beta)
\end{align}
\end{definition}
\begin{proposition}[Exact representation-decoder decomposition]\label{prop:error-decomposition}
Under the assumptions of Section~\ref{sec:abstract-approx}, for any measurable student coordinate $u=u(H_\varepsilon)$,
\begin{equation}
 \E\, D_P(q,u)=I_P+\E\, D_P(m(H_\varepsilon),u)
 \label{eq:decomp-general}
\end{equation}
The representation error is the minimum regret over all measurable probability decoders of the retained signal, with endpoint predictions interpreted by limits. In particular
\begin{equation}
 E^{G,P}_{\varepsilon,\tau}=I_P+D_P^{\rm dec}
\end{equation}
\end{proposition}
\begin{proof}
Since the student coordinate $u$ is a measurable function of $H_\varepsilon$, it can be taken outside the conditional expectation given $H_\varepsilon$. Using $\E\,[q\mid H_\varepsilon]=m(H_\varepsilon)$ in the definition of the regret yields
\begin{align*}
 \E\,[D_P(q,u)\mid H_\varepsilon]&=A(u)-m(H_\varepsilon)u-\E\,[B_P(q)\mid H_\varepsilon]\\
 &=D_P(m(H_\varepsilon),u)+B_P(m(H_\varepsilon))-\E\,[B_P(q)\mid H_\varepsilon]
\end{align*}
By concavity of $B_P$, Jensen's inequality implies that $B_P(m(H_\varepsilon))-\E\,[B_P(q)\mid H_\varepsilon]\ge0$ and this difference is integrable since $B_P$ is bounded. Taking expectations therefore gives \eqref{eq:decomp-general}, even if the expected regret is infinite. Strict propriety implies that the second term on the right-hand side of \eqref{eq:decomp-general} vanishes when the student predicts $m(H_\varepsilon)$, with predictions at $0,1$  interpreted by limits. Thus, the smallest regret attainable by any measurable probability prediction based on $H_\varepsilon$ is exactly $I_P$. Restricting the student coordinate to $u=\alpha H_\varepsilon+\beta$ and taking the infimum over $\alpha>0$ and $\beta\in\R$ yields the claimed decomposition.
\end{proof}
\begin{remark}
For la ogarithmic loss, $B_P$ is the binary Shannon
entropy and hence the representation error can
be expressed as
\begin{equation*}
 I_P=H(Y\mid H_\varepsilon)
 -H(Y\mid H_\varepsilon,Z_\varepsilon)
 =I(Y;Z_\varepsilon\mid H_\varepsilon)
\end{equation*}
in which $I(Y;Z_\varepsilon\mid H_\varepsilon)$ denotes
the conditional mutual information between the
binary response and the unresolved fluctuation
given the retained signal. For general proper losses,
$I_P$ measures the increase in the minimum expected
loss resulting from the omission of the unresolved
component and need not coincide with Shannon
conditional mutual information \citep{reid2011}.
\end{remark}

\subsection{Asymptotic decomposition}
\begin{theorem}[Asymptotic error decomposition]\label{thm:decomposition}
Under the assumptions of Theorem~\ref{thm:abstract-localisation}, with $\delta$ and $T$ given by \eqref{eq:abstract-threshold},
\begin{align}
 I_P&=\rho_0\left[\delta\mathscr H_P(T)-\tau^{-1}\mathscr H_P(L_G)\right]+o(\delta)\label{eq:info-asymptotic}\\
 D_P^{\rm dec}&=\rho_0\delta\left[\sqrt{v_P}-\mathscr H_P(T)\right]+o(\delta)\label{eq:decoder-asymptotic}
\end{align}
These expansions do not require convergence in law of $Z_\varepsilon$. If $G=P$ and the additional assumptions of Theorem~\ref{thm:abstract-sharp} hold, in the regular regime $I_P\sim\rho_0\tau\varepsilon^2/2$ and $D_P^{\rm dec}=o(\tau\varepsilon^2)$.
\end{theorem}
\begin{proof}
By the threshold representation \eqref{eq:abstract-threshold}, $m(\delta x)=F_T(x)$. In the expectation of $B_P(m(H_\varepsilon))$ we change variables from $h$ to $x=h/\delta$. In the expectation of $B_P(q)$ we first condition on $Z_\varepsilon$ and then use the variable $u=\tau(h+\varepsilon Z_\varepsilon)$. The definition of $I_P$ then  gives
\begin{align*}
 I_P=\delta\int_\R\rho_\varepsilon(\delta x)B_P(F_T(x))\dd x-\tau^{-1}\int_\R B_P(G(u))\E\,[\rho_\varepsilon(u/\tau-\varepsilon Z_\varepsilon)]\dd u
\end{align*}
By the definition of the Bayes risk we have $0\le B_P(F_T(x))\le A(x)-xF_T(x)$. The tail estimates in the proof of Theorem~\ref{thm:abstract-localisation}, applied with $b=1$ and $c=0$, show that the integral of the upper bound over $|x|>R$ approaches 0 uniformly along the asymptotic sequence as $R\to\infty$. In particular, uniform boundedness of $\rho_\varepsilon$ controls the contribution from this region, while on $[-R,R]$ the density $\rho_\varepsilon(\delta x)$ converges uniformly to $\rho_0$. It follows that the first term in the expression for $I_P$ is $\rho_0\delta\mathscr H_P(T)+o(\delta)$. The second term is the integrated teacher Bayes risk already evaluated in Theorem~\ref{thm:abstract-localisation}, which yields \eqref{eq:info-asymptotic}. Finally, by Proposition~\ref{prop:error-decomposition}, the decoder error is the difference between the optimal affine regret and $I_P$. Subtracting \eqref{eq:info-asymptotic} from \eqref{eq:abstract-additive} gives \eqref{eq:decoder-asymptotic}. In the matched regular regime, Proposition~\ref{prop:error-decomposition} identifies $I_P$ with the minimum population regret over all measurable probability predictions based on the retained signal. Theorem~\ref{thm:abstract-sharp} establishes that both $I_P$ and $E^{P,P}_{\varepsilon,\tau}$ are asymptotic to $\rho_0\tau\varepsilon^2/2$. Their nonnegative difference is consequently $o(\tau\varepsilon^2)$.
\end{proof}
\subsection{Transport representation of the decoder error}
For a real-valued random variable $V$ we use the generalised quantile $Q_V(u)=\inf\{x:F_V(x)\ge u\}$ for $0<u<1$. Distances between random variables below denote distances between their laws. The quadratic Wasserstein distance is the square root of the infimum of the expected squared difference over all couplings of the two laws. In one dimension it is given by the monotone quantile coupling \citep[Proposition~2.17]{santambrogio2015}
\begin{equation}
 W_2^2(V,W)=\int_0^1|Q_V(u)-Q_W(u)|^2\dd u
\end{equation}
\begin{proposition}[Quantile representation and integrated regret]\label{prop:transport}
If $T$ has finite second moment, then
\begin{equation}
 \mathscr H_P(T)=\int_0^1Q_{L_P}(u)Q_T(u)\dd u
 \label{eq:entropy-quantile}
\end{equation}
For $b>0$ and $c\in\R$ the integrated regret satisfies
\begin{equation}
 \int_\R D_P(F_T(x),bx+c)\dd x
 =\frac b2 W_2^2\left(T,\frac{L_P-c}{b}\right)
 \label{eq:transport-regret}
\end{equation}
\end{proposition}
\begin{proof}
The centring of $L_P$ and the canonical relation give
\begin{equation}
 B_P(q)=-\int_0^q Q_{L_P}(u)\dd u=\int_q^1Q_{L_P}(u)\dd u
\end{equation}
Using the first expression on $x<0$ and the second on $x\ge0$ yields
\begin{align*}
 \mathscr H_P(T)=-\int_{-\infty}^0\int_0^{F_T(x)}Q_{L_P}(u)\dd u\dd x+\int_0^\infty\int_{F_T(x)}^1Q_{L_P}(u)\dd u\dd x
\end{align*}
By the definition of the generalised quantile, $u\le F_T(x)$ if and only if $Q_T(u)\le x$. For a fixed $u$, the range of integration in the first integral is therefore $Q_T(u)\le x<0$ when $Q_T(u)<0$, and is empty otherwise. The range in the second integral is $0\le x<Q_T(u)$ when $Q_T(u)>0$, and is empty otherwise, with endpoints immaterial to the integration. The sum of the absolute integrals is bounded by
\begin{equation*}
 \int_0^1|Q_{L_P}(u)Q_T(u)|\dd u\le\sqrt{v_P\E\, T^2}
\end{equation*}
by Cauchy-Schwarz. Fubini's theorem thus permits interchange of the integrals, giving
\begin{equation}
 \mathscr H_P(T)=-\int_0^1 Q_{L_P}(u)(Q_T(u))_-\dd u
 +\int_0^1 Q_{L_P}(u)(Q_T(u))_+\dd u
\end{equation}
which proves \eqref{eq:entropy-quantile}, including for atomic laws. By Proposition~\ref{prop:canonical-integrated}, subtraction of $\mathscr H_P(T)$ yields
\begin{align}
 \int_\R D_P(F_T(x),bx+c)\dd x
 &=\frac{v_P+\E\,(bT+c)^2}{2b}-\mathscr H_P(T)\\
 &=\frac1{2b}\int_0^1|bQ_T(u)+c-Q_{L_P}(u)|^2\dd u
\end{align}
To obtain the second equality, we expand the square, use $\int_0^1 Q_{L_P}(u)\dd u=0$ and $\int_0^1 Q_{L_P}(u)^2\dd u=v_P$, and identify the cross term using \eqref{eq:entropy-quantile}. Since $b>0$, the quantile of $(L_P-c)/b$ is $(Q_{L_P}(u)-c)/b$, including at discontinuities of the quantile. Factoring $b^2$ out of the square and applying the one-dimensional transport formula proves \eqref{eq:transport-regret}.
\end{proof}
\begin{remark}
For logarithmic loss, the left-hand side of
\eqref{eq:transport-regret} is the cumulative paired
Shannon divergence with second argument $\sig(bx+c)$,
as considered by \citet{klein2016}. The restriction
of the second argument to the location-scale logistic
family is essential, since the cumulative divergence
is not in general symmetric for arbitrary pairs of
distribution functions. The quantile representation \eqref{eq:entropy-quantile}
is a form of the established cumulative-entropy
representation \citep{klein2016,klein2020} and is
included here to give a direct derivation of
\eqref{eq:transport-regret}. The latter expresses
the integrated regret for an affine canonical coordinate
as a quadratic Wasserstein distance. Our use of these
identities is to characterise the cumulative entropy
deficit and, through the population localisation
argument, the distribution-dependent leading-order
contribution of the decoder error to the population
regret.
\end{remark}
\begin{corollary}[Transport deficit and population decoder error]\label{cor:transport-decoder}
If $T$ has mean $\mu$ and variance $v>0$, then
\begin{equation}
 \sqrt{v_Pv}-\mathscr H_P(T)
 =\frac12\sqrt{\frac{v_P}{v}}W_2^2\left(T,\mu+\sqrt{v/v_P}L_P\right)
\end{equation}
Under the assumptions of Theorem~\ref{thm:decomposition}, its standardised threshold $T$ satisfies
\begin{equation}
 D_P^{\rm dec}=\frac{\rho_0\delta\sqrt{v_P}}2 W_2^2(T,L_P/\sqrt{v_P})+o(\delta)
 \label{eq:transport-decoder}
\end{equation}
If $\tau\varepsilon\to\infty$, then
\begin{equation}
 D_P^{\rm dec}=\frac{\rho_0\varepsilon\sqrt{v_P}}2 W_2^2(-Z_\varepsilon,L_P/\sqrt{v_P})+o(\varepsilon)
\end{equation}
In particular $D_P^{\rm dec}=o(\varepsilon)$ in this regime if and only if $W_2(-Z_\varepsilon,L_P/\sqrt{v_P})\to0$.
\end{corollary}
\begin{proof}
By Proposition~\ref{prop:canonical-integrated}, the minimising parameters are $b_*=\sqrt{v_P/v}$ and $c_*=-b_*\mu$, with minimum integrated loss $\sqrt{v_Pv}$. Subtracting the cumulative Bayes risk gives the minimum integrated regret $\sqrt{v_Pv}-\mathscr H_P(T)$. Evaluating \eqref{eq:transport-regret} at $(b_*,c_*)$ yields the first identity. The threshold $T$ in Theorem~\ref{thm:decomposition} has zero mean and unit variance, so substitution of the deficit identity with $\mu=0$ and $v=1$ into \eqref{eq:decoder-asymptotic} yields \eqref{eq:transport-decoder}. In the singular regime we have $\delta\sim\varepsilon$, and, recalling $T=(\tau^{-1}L_G-\varepsilon Z_\varepsilon)/\delta$ from \eqref{eq:abstract-threshold}, the coupling using the same $Z_\varepsilon$ and independent $L_G$ yields
\begin{equation}
 W_2^2(T,-Z_\varepsilon)\le \frac{v_G}{\tau^2\delta^2}+(1-\varepsilon/\delta)^2\to0
\end{equation}
Set $L_*=L_P/\sqrt{v_P}$, then each distance to $L_*$ is at most two by the triangle inequality through the point mass at zero, since all the normalised laws have unit second moment. The reverse triangle inequality thus yields
\begin{equation*}
 |W_2^2(T,L_*)-W_2^2(-Z_\varepsilon,L_*)|\le4W_2(T,-Z_\varepsilon)\to0
\end{equation*}
The remaining statements follow via substitution.
\end{proof}
\begin{remark}
The expansions above are additive. If the coefficient of $\delta$ (or of $\varepsilon$ in the singular regime) tends to zero, the remainder need not be asymptotically smaller than the leading term. Thus, a relative equivalent for the decoder contribution does not follow along such paths.
\end{remark}

\subsection{Limiting error coefficients}
\begin{lemma}[Continuity of cumulative Bayes risk]\label{lem:entropy-continuity}
For any two laws with finite second moments we have
\begin{equation*}
 |\mathscr H_P(V)-\mathscr H_P(W)|\le\sqrt{v_P}W_2(V,W)
\end{equation*}
In particular, if $V_n\Rightarrow V$ and the squares $V_n^2$ are uniformly integrable, then $\mathscr H_P(V_n)\to\mathscr H_P(V)$.
\end{lemma}
\begin{proof}
By Proposition~\ref{prop:transport} and Cauchy-Schwarz
\begin{equation*}
 |\mathscr H_P(V)-\mathscr H_P(W)|\le\|Q_{L_P}\|_2\|Q_V-Q_W\|_2=\sqrt{v_P}W_2(V,W)
\end{equation*}
The weak convergence and uniform integrability of the squares imply convergence of second moments by truncation. Hence, the characterisation of convergence in the
quadratic Wasserstein distance
\citep[Theorem~6.9]{villani2009} yields
$W_2(V_n,V)\to0$. It then follows from the Lipschitz
estimate
\begin{equation*}
 |\mathscr H_P(V_n)-\mathscr H_P(V)|
 \leq\sqrt{v_P}\,W_2(V_n,V)
\end{equation*}
that $\mathscr H_P(V_n)\to\mathscr H_P(V)$.
\end{proof}

Suppose now that $Z_\varepsilon\Rightarrow Z$ and
$\eta=\tau\varepsilon\to\eta_0\in(0,\infty)$.
Since
\begin{equation*}
 \frac{\delta T}{\varepsilon}
 =\frac{L_G}{\eta}-Z_\varepsilon
\end{equation*}
independence and uniform integrability of
$Z_\varepsilon^2$ imply convergence in quadratic
Wasserstein distance to $L_G/\eta_0-Z$, with
$L_G$ is independent of $Z$. Furthermore, a change
of variables in the defining integral of the
cumulative Bayes risk suggests
$\mathscr H_P(aV)=a\mathscr H_P(V)$ for any $a>0$.
It therefore follows from
Lemma~\ref{lem:entropy-continuity} that the limiting
coefficients of $\rho_0\varepsilon$ in the
representation and decoder error expansions
\eqref{eq:info-asymptotic} and
\eqref{eq:decoder-asymptotic} are, respectively
\begin{align}
 &\mathscr H_P(L_G/\eta_0-Z)
 -\mathscr H_P(L_G)/\eta_0\\
 &\sqrt{v_P}\sqrt{1+v_G/\eta_0^2}
 -\mathscr H_P(L_G/\eta_0-Z)
\end{align}
In the singular regime $\eta\to\infty$, the
corresponding coefficients are
$\mathscr H_P(-Z)$ and
$\sqrt{v_P}-\mathscr H_P(-Z)$, respectively.

\begin{remark}
The distinction between $Z$ and $-Z$ is important
when the canonical loss is not symmetric under
interchange of the two binary outcomes. Since $Z$
has zero mean and unit variance, the coefficient
of the decoder error in the singular regime
vanishes if and only if
\begin{equation*}
 -Z\overset{\mathrm d}=L_P/\sqrt{v_P}
\end{equation*}
by the equality condition in
\eqref{eq:cumulative-entropy-bound}.
For logarithmic loss, the cumulative Bayes risk
is invariant under reflection of its argument and
hence the singular decoder coefficient vanishes
when the limiting residual has a
unit-variance logistic distribution.

To illustrate the dependence of the representation
and decoder errors on the residual distribution,
consider a symmetric unit-variance residual supported
on $\{-1,1\}$ and a residual with a centred
unit-variance logistic distribution. Their cumulative
paired Shannon entropies are $2\log 2$ and
$\pi/\sqrt{3}$, respectively. Although the total
leading-order regret is identical for the two
residual distributions, their representation and
decoder contributions differ. In particular, the
decoder contribution vanishes to leading order
for the logistic residual but not for the
two-point residual.
\end{remark}

\section{Rank-constrained quadratic approximation}\label{sec:model}
\subsection{Gaussian quadratic teacher and rank-constrained student}
Let $d\ge4$ be an integer, let $X\sim\cN(0,I_d)$, set
\begin{equation}
\Lambda=\diag(\lambda_1,\ldots,\lambda_d),\qquad
\lambda_1\ge\lambda_2\ge\cdots\ge\lambda_d\ge0
\label{eq:teacher-spectrum}
\end{equation}
and define the centred quadratic teacher signal
\begin{equation}
F(X)=X^\top\Lambda X-\tr\Lambda
=\sum_{j=1}^d\lambda_j(X_j^2-1)
\label{eq:teacher}
\end{equation}
We first treat the zero-temperature problem, with deterministic teacher label
\begin{equation}
Y=\one_{\{F(X)>0\}}
\label{eq:label}
\end{equation}
and the student class consists of centred quadratic logits
\begin{equation}
f_S(X)=X^\top SX-\tr S,\qquad
S\succeq0,\quad\rank S\le p
\label{eq:student}
\end{equation}
with predicted conditional probability $\sig(f_S(X))$, in which $\sig(u)=(1+e^{-u})^{-1}$. The factorisation $S=WW^\top$ with $W\in\R^{d\times p}$ shows that this is an exactly rank-constrained quadratic network. Fixed architecture-dependent prefactors can be absorbed into $W$ and therefore do not change the present population class. The intercept is not separately trained. The subtraction of $\tr S$ matches the centring of the teacher signal and is part of the model definition. The upper-bound construction below is admissible in precisely this class.

The population binary cross-entropy of the student is
\begin{equation}
\begin{aligned}
\CE(S)&=\E\,\bigl[-Y\log\sig(f_S(X))-(1-Y)\log(1-\sig(f_S(X)))\bigr]\\
&=\E\,\log\bigl(1+e^{-(2Y-1)f_S(X)}\bigr)
\end{aligned}
\label{eq:loss}
\end{equation}
We study the optimal risk
\begin{equation}
\mathcal E_p(\Lambda)=\inf_{\substack{S\succeq0\\\mathrm{rank} S\le p}}\CE(S)
\label{eq:optimal-risk}
\end{equation}
which is an approximation quantity, not a generalisation bound for a finite training set. Let us define the spectral-tail standard deviation and the full Frobenius norm by
\begin{equation}
s_p=\left(2\sum_{j>p}\lambda_j^2\right)^{1/2},\qquad
M=\Frob{\Lambda}=\left(\sum_{j=1}^d\lambda_j^2\right)^{1/2}
\label{eq:tail}
\end{equation}
For $p<d$ we shall assume $s_p>0$. Some statements below are trivial if $s_p=0$.

\subsection{Arbitrary Gaussian projections and conditional information}
Let $U\in\R^{d\times p}$ have orthonormal columns
spanning the retained subspace and let
$V\in\R^{d\times(d-p)}$ have orthonormal columns
spanning its orthogonal complement, so that
$(U,V)$ is an orthogonal matrix. In order to
quantify the effect of discarding the coordinates
in the orthogonal complement, we construct two
Gaussian inputs with identical retained coordinates
but independently sampled unresolved coordinates.
To this end, let $Z\sim\cN(0,I_p)$ and
$W,W'\sim\cN(0,I_{d-p})$ be mutually independent
Gaussian random vectors and define
\begin{equation}
 X=UZ+VW,\qquad X'=UZ+VW'
 \label{eq:coupling}
\end{equation}
Both $X$ and $X'$ are standard Gaussian random
vectors in $\R^d$. Furthermore, since $U^\top U=I_p$
and $U^\top V=0$, their representations available
to the student are identical, namely
$U^\top X=U^\top X'=Z$. However, the teacher,
which has access to the full input, may assign
different binary outcomes to $X$ and $X'$.

We define the probability of disagreement between
the deterministic teacher decisions for these
two inputs by
\begin{equation}
 D(U)=\Prob\bigl(\one_{\{F(X)>0\}}\ne\one_{\{F(X')>0\}}\bigr)
 \label{eq:disagreement}
\end{equation}
Let $Y=\one_{\{F(X)>0\}}$ and
$Y'=\one_{\{F(X')>0\}}$ denote the corresponding
binary responses and let
$\eta_U(Z)=\Prob(Y=1\mid Z)$ denote the conditional
probability of the positive binary outcome given
the retained coordinates. Since $W$ and $W'$ are
independent of each other and of $Z$, the binary
responses $Y$ and $Y'$ are conditionally independent
given $Z$, with the same conditional probability
$\eta_U(Z)$ of taking the value one. The conditional
probability of disagreement is therefore
\begin{equation*}
 \Prob(Y\ne Y'\mid Z)=2\eta_U(Z)(1-\eta_U(Z))
\end{equation*}
Taking the expectation with respect to $Z$ yields
\begin{equation}
 D(U)=2\E\,\bigl[\eta_U(Z)(1-\eta_U(Z))\bigr]
 \label{eq:disagreement-eta}
\end{equation}

The minimum population cross-entropy over all
measurable probability predictions based on $Z$
is the conditional entropy
$H(Y\mid Z)=\E\,h(\eta_U(Z))$, in which
$h(t)=-t\log t-(1-t)\log(1-t)$ is the binary
Shannon entropy. Since $-\log t\geq1-t$ and
$-\log(1-t)\geq t$ for $0<t<1$, it follows that
$h(t)\geq2t(1-t)$, with the inequality also holding
at the endpoints by continuity. Hence, by virtue of
\eqref{eq:disagreement-eta}
\begin{equation}
 H(Y\mid Z)\geq D(U)
 \label{eq:entropy-disagreement}
\end{equation}
To relate this bound to the rank-constrained
approximation problem, consider any admissible
student matrix $S$. Since $\rank S\leq p$, there
exists a $p$-dimensional subspace containing the
range of $S$. Let $U$ be a matrix whose columns
form an orthonormal basis for this subspace.
By symmetry of $S$, we have
$S=U(U^\top SU)U^\top$ and hence
\begin{equation*}
 f_S(X)=Z^\top(U^\top SU)Z-\tr S
\end{equation*}
in which $Z=U^\top X$. The student probability
prediction is therefore a measurable function of
the retained coordinates $Z$ alone and its
population cross-entropy cannot be smaller than
$H(Y\mid Z)$. Consequently, any lower bound on
$D(U)$ that holds uniformly over all admissible
choices of $U$ also provides a lower bound on
the optimal rank-constrained population regret
in \eqref{eq:optimal-risk}, since the teacher
Bayes risk is zero in the deterministic case.
\subsection{Conditional Gaussian quadratic forms and unresolved fluctuations}\label{sec:results}
\begin{proposition}[Conditional decomposition]\label{prop:decomp}
Let $U,V$ be as above and let
\begin{equation}
\begin{pmatrix}A&B\\B^\top&C\end{pmatrix}
=(U,V)^\top\Lambda(U,V)
\end{equation}
Then
\begin{equation}
F(X)=m(Z)+R_Z(W)
\end{equation}
in which
\begin{equation}
m(z)=z^\top Az-\tr A,\qquad
R_z(w)=2z^\top Bw+w^\top Cw-\tr C
\label{eq:conditional-decomposition}
\end{equation}
Moreover, we have
\begin{equation}
\E\,[R_z(W)]=0,\qquad
\Var(R_z(W))=4\lVert B^\top z\rVert_2^{2}+2\Frob{C}^2
\label{eq:conditional-variance}
\end{equation}
\end{proposition}
\begin{proof}
Substituting $X=Uz+Vw$ into the centred quadratic
teacher signal and using the definitions of the
matrices $A$, $B$ and $C$, we obtain
\begin{equation*}
 F(Uz+Vw)=z^\top Az+2z^\top Bw+w^\top Cw-\tr\Lambda
\end{equation*}
Since $(U,V)$ is an orthogonal matrix, invariance
of the trace under orthogonal transformations gives
$\tr\Lambda=\tr A+\tr C$, from which the stated
decomposition follows.

For fixed $z$, the linear term $2z^\top BW$ and
the centred quadratic term $W^\top CW-\tr C$
both have zero mean. Moreover, their covariance
vanishes since the product
$(z^\top Bw)(w^\top Cw-\tr C)$ changes sign under
the transformation $w\mapsto-w$, whereas the
Gaussian distribution of $W$ is invariant under
this transformation. Using $\E\,WW^\top=I_{d-p}$,
we obtain
\begin{equation*}
 \E\,[(2z^\top BW)^2]=4z^\top B\E\,[WW^\top]B^\top z=4\|B^\top z\|_2^2
\end{equation*}
By \eqref{eq:gaussian-quadratic-variance},
the variance of the centred quadratic term
is $2\Frob{C}^2$. Since the linear
and quadratic terms are uncorrelated, the
conditional variance of $F(X)$ given $Z=z$ is
the sum of their respective variances, i.e.,
\begin{equation*}
 \Var(F(X)\mid Z=z)=4\|B^\top z\|_2^2+2\Frob{C}^2
\end{equation*}
which proves the assertion.
\end{proof}

\begin{lemma}[Unavoidable conditional quadratic variance]\label{lem:interlace}
For every $p$-dimensional subspace $U$ and every $z\in\R^p$, we have
\begin{equation}
\Var(R_z(W))\ge s_p^2
\label{eq:unavoidable}
\end{equation}
\end{lemma}
\begin{proof}
Since $C=V^\top\Lambda V$ is a compression of the symmetric matrix $\Lambda$ to a subspace of codimension $p$, the Poincar\'e separation theorem (the compression interlacing theorem; see \citet[Corollary~III.1.5]{bhatia1997}) yields $\lambda_i(C)\ge\lambda_{i+p}(\Lambda)$ for $1\le i\le d-p$. Both matrices are positive semidefinite, so the eigenvalue inequalities can be squared term by term. Therefore
\begin{equation}
\Frob{C}^2=\sum_{i=1}^{d-p}\lambda_i(C)^2\ge\sum_{j=p+1}^d\lambda_j^2
\end{equation}
and the claim follows from~\eqref{eq:conditional-variance}.
\end{proof}

\begin{remark}
The bound \eqref{eq:unavoidable} holds for every value of the retained coordinates rather than only after averaging over their distribution. Positive semidefiniteness is essential to the step in the proof in which the interlacing inequalities are squared.
\end{remark}

\subsection{Two-sided Gaussian polynomial fluctuations}
\begin{lemma}[Uniform two-sided tails]\label{lem:twosided}
There exist numerical constants $a,c_0>0$ such that every nonzero centred Gaussian polynomial $R$ of degree at most two with variance $v^2$ satisfies
\begin{equation}
\Prob(R\ge av)\ge c_0,\qquad
\Prob(R\le-av)\ge c_0
\label{eq:twosided}
\end{equation}
One admissible choice is $a=1/36$ and $c_0=1/1296$.
\end{lemma}
\begin{proof}
Gaussian hypercontractivity for Gaussian polynomials of degree at most two \citep[Chapter~5]{janson1997} yields $\lVert R\rVert_4\le3\lVert R\rVert_2=3v$, and thus  $\E\, R^4\le81v^4$. In the usual Gaussian hypercontractive inequality $\|R\|_q\le(q-1)^{k/2}\|R\|_2$ for polynomials of degree at most $k$, these constants correspond to $k=2$, $q=4$. H\"older's inequality applied to $|R|^{2/3}|R|^{4/3}$ yields $\E\,|R|^2\le(\E\,|R|)^{2/3}(\E\,|R|^4)^{1/3}$, and rearranging yields
\begin{equation}
\E\,|R|\ge\frac{(\E\, R^2)^{3/2}}{(\E\, R^4)^{1/2}}\ge\frac v9
\end{equation}
Since $\E\, R=0$, each of $\E\, R_+$ and $\E\,(-R)_+$ is at least $v/18$. With $a=1/36$, the contribution from $0<R\le av$ to $\E\, R_+$ is at most $av$, whereupon
\begin{equation}
\E\,[R\one_{\{R>av\}}]\ge v/36
\end{equation}
By Cauchy-Schwarz,
\begin{equation*}
 v/36\le\E\,[R\one_{\{R>av\}}]\le v\Prob(R>av)^{1/2}
\end{equation*}
which yields $\Prob(R>av)\ge1/1296$. We can apply the argument to $-R$ for the other tail.
\end{proof}

\subsection{Small-ball probabilities for centred positive quadratic forms}
\begin{lemma}[Uniform central small-ball estimate]\label{lem:smallball}
Let $Z\sim\cN(0,I_m)$ and $A\succeq0$ be nonzero. Define $Q(Z)=Z^\top AZ-\tr A$ and $\sigma_Q^2=2\Frob{A}^2$. There exists a numerical constant $c_*>0$, independent of $m$ and $A$, such that for every $r>0$
\begin{equation}
\Prob(|Q|\le r)\ge c_*\min\left\{1,\frac r{\sigma_Q}\right\}
\label{eq:smallball}
\end{equation}
For $A=0$, the probability equals one.
\end{lemma}
\begin{proof}
Let $\gamma_m$ denote the standard Gaussian measure on $\R^m$ and let $\varphi_m$ denote its density. Further, let $\varphi$ and $\Phi$ denote the one-dimensional standard Gaussian density and distribution function, respectively. Let $K_t=\{z:Q(z)\le t\}$, and denote its Gaussian perimeter by $P_\gamma(K_t)$. Applying Lemma~\ref{lem:twosided} to $Q$, we can find universal constants $a,c_0>0$ such that for $|t|\le a\sigma_Q$
\begin{equation}
c_0\le\gamma_m(K_t)\le1-c_0
\label{eq:central-cdf}
\end{equation}
The Gaussian isoperimetric inequality \citep{borell1975,sudakov1978,ledoux2001} therefore yields
\begin{equation}
P_\gamma(K_t)\ge\varphi\bigl(\Phi^{-1}(\gamma_m(K_t))\bigr)\ge c_1>0
\label{eq:perimeter-lower}
\end{equation}
for a numerical constant $c_1$ independent of the dimension.

The Ornstein-Uhlenbeck operator $\mathscr L=\Delta-z\cdot\nabla$ satisfies $\mathscr LQ=-2Q$, and since $A\succeq0$ and $A\ne0$, the only critical value of $Q$ is $-\tr A$; indeed, $\nabla Q(z)=2Az=0$ implies $Q(z)=-\tr A$. The critical value lies outside the central interval since $\tr A\ge\Frob{A}=\sigma_Q/\sqrt2$ and we may take $a=1/36$. Thus, every level in that interval is regular, and its Gaussian perimeter equals the surface integral of the Gaussian density over the boundary. The critical set $\ker A$ is a proper linear subspace and has Gaussian measure zero. The distribution of $Q$ has no atoms, since after diagonalising $A$ it is the convolution of the non-atomic law of one positive multiple of a centred $\chi_1^2$ variable with an independent probability measure.

The weighted divergence identity is
\begin{equation*}
 \operatorname{div}(\varphi_m\nabla Q)=\varphi_m(\Delta Q-z\cdot\nabla Q)=\varphi_m\mathscr LQ
\end{equation*}
and on a regular level the outward unit normal to the sublevel set is $\nabla Q/|\nabla Q|$. For any regular value $t$, apply the divergence theorem to the vector field $\varphi_m\nabla Q$ on $K_t\cap B_R$, in which $B_R$ is the Euclidean ball of radius $R$. The boundary term on $\partial B_R\cap K_t$ vanishes as $R\to\infty$, since $|\nabla Q(z)|\le2\|A\|_{\mathrm{op}}|z|$ and the absolute Gaussian-weighted flux on the truncating sphere is bounded by $C_m\|A\|_{\mathrm{op}}R^m e^{-R^2/2}$, which approaches 0 for each fixed dimension. The dimension-dependent constant enters only this truncation argument and disappears before the dimension-independent estimate below is derived. We may take radii tending to $\infty$,  for which the truncating sphere meets the regular level transversely, or use smooth radial cutoffs. The boundary integral on $\partial K_t\cap B_R$ converges by monotone convergence, since its integrand is nonnegative. Thus the Gaussian divergence theorem yields
\begin{equation}
\int_{\partial K_t}|\nabla Q|\,\varphi_m(z)\dd S(z)
=\int_{K_t}\mathscr LQ\,\dd\gamma_m
=-2\E\,[Q\one_{\{Q\le t\}}]
\le2\E\,|Q|\le2\sigma_Q
\label{eq:gradient-surface}
\end{equation}
in which the integral on the left-hand side is nonnegative. Its upper bound, which is independent of the dimension, is the quantity needed to convert the lower bound on Gaussian perimeter into a lower bound on the density of $Q$. To apply the coarea formula \citep[Section~3.4]{evansgariepy2015} away from the critical
set $\cC=\{z\in\R^m:\nabla Q(z)=0\}$, introduce
the increasing sequence of sets
\begin{equation*}
 \Omega_n=\{z\in\R^m:|z|\leq n,\ |\nabla Q(z)|\geq n^{-1}\}\qquad \bigcup_n\Omega_n=\R^m\setminus\cC
\end{equation*}
Now, for any nonnegative measurable
function $g$, applying the
coarea formula 
on $\Omega_n$ yields
\begin{equation*}
 \int_{\Omega_n}g(Q(z))\varphi_m(z)\dd z
 =\int_\R g(t)\int_{Q^{-1}(t)\cap\Omega_n}
 \frac{\varphi_m(z)}{|\nabla Q(z)|}\dd S(z)\dd t
\end{equation*}
Since the integrands are nonnegative, we can apply
the monotone convergence theorem to both sides
as $n\to\infty$. Moreover, since the critical set
has Gaussian measure zero, its exclusion does not
affect the integral on the left-hand side.
Consequently, for a.e.\ regular value $t$, the
density of the pushforward law is given by
\begin{equation}
 \rho_Q(t)=\int_{\partial K_t}
 \frac{\varphi_m(z)}{|\nabla Q(z)|}\dd S(z)
\end{equation}
By Cauchy-Schwarz on $\partial K_t$,
\begin{equation}
P_\gamma(K_t)^2\le
\left(\int_{\partial K_t}|\nabla Q|\varphi_m\dd S\right)
\left(\int_{\partial K_t}\frac{\varphi_m}{|\nabla Q|}\dd S\right)
\end{equation}
and together with~\eqref{eq:perimeter-lower} and~\eqref{eq:gradient-surface}, this implies
\begin{equation}
\rho_Q(t)\ge\frac{c_1^2}{2\sigma_Q}
\quad\text{for almost every }|t|\le a\sigma_Q
\end{equation}
For $r\le a\sigma_Q$, integration on $[-r,r]$ yields the required lower bound proportional to $r/\sigma_Q$. For $r>a\sigma_Q$, monotonicity yields a fixed positive lower bound from the case $r=a\sigma_Q$. The case $A=0$ is immediate.
\end{proof}

\subsection{Uniform threshold disagreement}
\begin{theorem}[Projection-resampling disagreement]\label{thm:disagreement}
Let $\Lambda\succeq0$ with $M=\Frob{\Lambda}>0$ and let $1\le p<d$. For every $p$-dimensional subspace $U$, let $(X,X')$ be coupled as in \eqref{eq:coupling}. Then there exists an absolute constant $c>0$ such that
\begin{equation}
D(U)\ge c\frac{s_p}{M}
\label{eq:main-disagreement}
\end{equation}
with  $s_p$ defined as in~\eqref{eq:tail}.
\end{theorem}
\begin{proof}
The assertion is immediate when $s_p=0$, so suppose that $s_p>0$. Using the decomposition $F(X)=m(Z)+R_Z(W)$ of Proposition~\ref{prop:decomp}, and setting $v(z)^2=\Var(R_z(W))$, Lemma~\ref{lem:interlace} shows that $v(z)\ge s_p$. Lemma~\ref{lem:twosided}, applied conditionally at each fixed $z$, shows that $R_z(W)$ has probability at least $c_0$ of being above $av(z)$, as well as probability at least $c_0$ of being below $-av(z)$.

To obtain a lower bound on disagreement we restrict attention to retained coordinates for which the conditional mean signal is small relative to the unresolved standard deviation. Define $E=\{|m(Z)|\le as_p/2\}$. For each $z$ satisfying $|m(z)|\le as_p/2$, the event $R_z(W)\ge av(z)$ implies $F(X)\ge as_p/2>0$, and $R_z(W)\le-av(z)$ implies $F(X)\le-as_p/2<0$. Each event has conditional probability at least $c_0$, so $c_0\le\eta_U(z)\le1-c_0$. Hence $2\eta_U(Z)(1-\eta_U(Z))\ge2c_0^2$ on $E$, and therefore
\begin{equation}
D(U)\ge2c_0^2\Prob(E)
\label{eq:disagreement-event}
\end{equation}
The matrix $A=U^\top\Lambda U$ is positive semidefinite, and $m(Z)=Z^\top AZ-\tr A$ has standard deviation $\sigma_m=\sqrt2\Frob{A}\le\sqrt2M$. If $A\ne0$, Lemma~\ref{lem:smallball} with $r=as_p/2$ yields
\begin{equation}
\Prob(E)\ge c_*\min\left\{1,\frac{as_p}{2\sigma_m}\right\}
\end{equation}
Since $s_p\le\sqrt2M$, the right-hand side is bounded below by $c_2s_p/M$ for an absolute $c_2>0$. If $A=0$, then $m\equiv0$ and $\Prob(E)=1$, which satisfies the same bound after reducing the constant. Substitution into~\eqref{eq:disagreement-event} completes the proof.
\end{proof}

\begin{remark}
Theorem \ref{thm:disagreement} bounds the disagreement probability uniformly over retained subspaces. It does not identify the subspace that minimises conditional entropy among arbitrary measurable predictors, although no such predictor can improve the order of the spectral-tail lower bound. Within the quadratic logistic class, Proposition~\ref{prop:alignment} will show that rotation alone cannot improve the optimum.
\end{remark}

\subsection{Global population cross-entropy bounds}
\begin{corollary}[Information-theoretic rank lower bound]\label{cor:entropy-lower}
Under the assumptions of Theorem~\ref{thm:disagreement}, every predictor measurable with respect to a $p$-dimensional orthogonal Gaussian projection has cross-entropy at least $cs_p/M$, and in particular
\begin{equation}
\mathcal E_p(\Lambda)\ge c\frac{s_p}{M}
\label{eq:risk-lower}
\end{equation}
\end{corollary}
\begin{proof}
The minimal cross-entropy among all predictions measurable with respect to $Z=U^\top X$ is $H(Y\mid Z)$. By~\eqref{eq:entropy-disagreement} and Theorem~\ref{thm:disagreement}, $H(Y\mid Z)\ge D(U)\ge cs_p/M$. For any $S\succeq0$ with rank at most $p$, choose a $p$-dimensional subspace containing $\operatorname{range}S$; the logistic predictor $\sig(f_S(X))$ is measurable with respect to the corresponding Gaussian projection. Taking the infimum over all admissible matrices $S$ gives \eqref{eq:risk-lower}.
\end{proof}

\begin{lemma}[Uniform density of the retained quadratic component]
\label{lem:density}
Suppose that the two largest eigenvalues of the teacher
matrix $\Lambda$ satisfy $\lambda_1\geq a_1>0$ and
$\lambda_2\geq a_2>0$, in which $a_1$ and $a_2$ are
constants independent of the ambient dimension $d$.
For every $p\geq2$, the random variable
\begin{equation}
H_p=\sum_{j=1}^p\lambda_j(X_j^2-1)
\end{equation}
has a density $\rho_p$ satisfying
\begin{equation}
\|\rho_p\|_\infty\le\frac{1}{2\sqrt{\lambda_1\lambda_2}}\le\frac{1}{2\sqrt{a_1a_2}}=:K
\label{eq:density-explicit}
\end{equation}
with constant $K$ independent of $p$ and $d$.
\end{lemma}
\begin{proof}
For $i\in\{1,2\}$, the positive variable $V_i=\lambda_iX_i^2$ has density
\begin{equation}
f_i(t)=\frac{\exp(-t/(2\lambda_i))}{\sqrt{2\pi\lambda_i t}}\,\one_{\{t>0\}}
\end{equation}
The convolution density of $V_1+V_2$ satisfies, for $t>0$
\begin{equation}
\begin{aligned}
(f_1*f_2)(t)
\le\frac{1}{2\pi\sqrt{\lambda_1\lambda_2}}
\int_0^t\frac{\dd u}{\sqrt{u(t-u)}}=\frac{1}{2\sqrt{\lambda_1\lambda_2}}
\end{aligned}
\end{equation}
in which the substitution $u=tv$ reduces the integral to $B(1/2,1/2)=\pi$. Translating by $-\lambda_1-\lambda_2$ does not change the $L^\infty$ norm. The remaining centred summands are independent; convolution with their joint probability law cannot increase the supremum of a bounded density. This proves~\eqref{eq:density-explicit} for any $p\ge2$.
\end{proof}

\begin{proposition}[Explicit logistic upper bound]\label{prop:upper}
Under the assumptions of Lemma~\ref{lem:density}, for $p\ge2$ with $s_p>0$, the positive-semidefinite rank-$p$ matrix
\begin{equation}
S_p=\frac1{s_p}\diag(\lambda_1,\ldots,\lambda_p,0,\ldots,0)
\label{eq:upper-student}
\end{equation}
satisfies
\begin{equation}
\CE(S_p)\le K\left(1+\frac{\pi^2}{6}\right)s_p
\label{eq:upper-risk}
\end{equation}
\end{proposition}
\begin{proof}
First set $F=H_p+R_p$, in which $R_p$ is independent of $H_p$, is centred, and has variance $s_p^2$. The student logit associated with~\eqref{eq:upper-student} is $H_p/s_p$ and for any nonzero $H_p$ and binary label $Y=\one_{\{F>0\}}$, the logistic loss can be decomposed as follows
\begin{equation}
\log(1+e^{-(2Y-1)H_p/s_p})
=\log(1+e^{-|H_p|/s_p})
+\frac{|H_p|}{s_p}\one_{\{\operatorname{sign}H_p\ne\operatorname{sign}F\}}
\label{eq:loss-decomposition}
\end{equation}
Now, since the density of $H_p$ is bounded by $K$
\begin{equation}
\E\,\log(1+e^{-|H_p|/s_p})
\le Ks_p\int_\R\log(1+e^{-|u|})\dd u
=\frac{\pi^2}{6}Ks_p
\end{equation}
in which the last step (for the logistic canonical potential) follows from  the deterministic threshold calculation at 0 in Proposition~\ref{prop:canonical-integrated}. If $H_p$ and $F=H_p+R_p$ have different signs, then the magnitude of the unresolved component must be at least that of the retained signal, namely, $|H_p|\le|R_p|$. By independence, the conditional density of $H_p$ given $R_p$ is $\rho_p$, so using its uniform bound yields
\begin{equation}
\E\,\bigl[|H_p|\one_{\{|H_p|\le|R_p|\}}\mid R_p\bigr]
\le K\int_{-|R_p|}^{|R_p|}|h|\dd h=KR_p^2
\end{equation}
Therefore, the expectation of the second term of~\eqref{eq:loss-decomposition} is at most $K\E\, R_p^2/s_p=Ks_p$. Adding together the two bounds proves~\eqref{eq:upper-risk}.
\end{proof}

\begin{theorem}[Matching width-dependent cross-entropy bounds]\label{thm:main}
Consider a family of dimensions $d$ and teacher matrices $\Lambda_d\succeq0$ with decreasing eigenvalues $\lambda_{1,d}\ge\cdots\ge\lambda_{d,d}\ge0$. Assume that there exist constants $0<M_-\le M_+<\infty$ and $a_1,a_2>0$, independent of $d$, such that
\begin{equation}
M_-\le\|\Lambda_d\|_{\mathrm F}\le M_+
\qquad
\lambda_{1,d}\ge a_1
\qquad
\lambda_{2,d}\ge a_2
\label{eq:uniform-spectral-assumptions}
\end{equation}
Then for every $d$ and every $2\le p<d$ with
\begin{equation}
s_{p,d}^2=2\sum_{j=p+1}^d\lambda_{j,d}^2>0
\end{equation}
there are constants $0<c<C<\infty$, depending only on $M_+$, $a_1$ and $a_2$ but not on $p$ or $d$, such that the optimal population cross-entropy satisfies
\begin{equation}
c\,s_{p,d}\le\mathcal E_p(\Lambda_d)\le C\,s_{p,d}
\label{eq:main-bound}
\end{equation}
\end{theorem}
\begin{proof}
By Corollary~\ref{cor:entropy-lower}, $\mathcal E_p(\Lambda_d)\ge c_0s_{p,d}/\|\Lambda_d\|_{\mathrm F}\ge(c_0/M_+)s_{p,d}$. Proposition~\ref{prop:upper} and the density bound $K\le(2\sqrt{a_1a_2})^{-1}$ provide an admissible student with risk at most $K(1+\pi^2/6)s_{p,d}$. Both constants are independent of $p$ and $d$.
\end{proof}

\begin{corollary}[Power-law width exponent]\label{cor:powerlaw}
Suppose $\lambda_j=C_0j^{-\gamma}$ for $1\le j\le d$, with $C_0>0$, $\gamma>1/2$, and $p\to\infty$, $d/p\to\infty$. Then
\begin{equation}
\mathcal E_p(\Lambda)\asymp p^{1/2-\gamma}
\label{eq:powerlaw}
\end{equation}
in which the implicit constants are independent of $p$ and $d$.
\end{corollary}
\begin{proof}
The Frobenius norm is bounded above and away from 0 by virtue of $\sum_{j\ge1}j^{-2\gamma}<\infty$. The first two eigenvalues are bounded below uniformly by positive constants. Since $x^{-2\gamma}$ is decreasing, its sum from $p+1$ to $d$ lies between the integrals over $[p+1,d+1]$ and $[p,d]$. Then evaluating both integrals and using $d/p\to\infty$ yields
\begin{equation}
\sum_{j=p+1}^d\lambda_j^2
\sim\frac{C_0^2}{2\gamma-1}p^{1-2\gamma}
\end{equation}
when $d/p\to\infty$. The proof is completed by applying Theorem~\ref{thm:main}.
\end{proof}

\begin{remark}[Comparison with squared-error approximation]
For the same quadratic teacher, Gaussian fourth-moment identities imply that the squared error of an eigen-aligned rank-$p$ truncation is exactly $2\sum_{j>p}\lambda_j^2=s_p^2$, and the Frobenius/Eckart-Young argument shows that this is the best squared-logit approximation among positive-semidefinite rank-$p$ matrices. Hence, the corresponding power-law exponent is $2\gamma-1$, compared with $\gamma-\frac12$ for a deterministic-label logistic cross-entropy. This difference is related to two distinct observation and loss models, and  should not be interpreted as a universal claim about all classification and regression tasks.
\end{remark}

\begin{remark}[Parameter count]
A factorisation $S=WW^\top$, $W\in\R^{d\times p}$, has $dp$ entries before quotienting orthogonal factorisation redundancies. Corollary~\ref{cor:powerlaw} is therefore a statement concerning the width $p$. Translating this  result  into a parameter-count law would require the specification of  how the input dimension $d$ changes, or would require holding $d$ fixed over a finite range. In the joint limit $d/p\to\infty$, it is not mathematically correct to replace $p$ by the total parameter count without a separate architecture-scaling relation.
\end{remark}

\section{Finite-temperature bounds and sharp spectral asymptotics}\label{sec:temperature}
We now retain the same student class while introducing a probabilistic teacher with inverse temperature $\tau>0$
\begin{equation}
q_\tau(Y=1\mid X)=\sig(\tau F(X))
\label{eq:finite-temp-teacher}
\end{equation}
If we write $h(q)=-q\log q-(1-q)\log(1-q)$, $H_\tau=\E\, h(q_\tau)$ and ${\CE}_\tau(S)=\E\,[\psi(f_S)-q_\tau f_S]$, in which $\psi(u)=\log(1+e^u)$, the excess population risk is
\begin{equation}
\begin{aligned}
\mathcal E_{p,\tau}&=\inf_{\substack{S\succeq0\\\rank S\le p}}\E\,\KL\!\left(\operatorname{Bern}(\sig(\tau F(X)))\,\middle\|\,\operatorname{Bern}(\sig(f_S(X)))\right)
\\ &=\inf_S{\CE}_\tau(S)-H_\tau
\end{aligned}
\label{eq:finite-temp-risk}
\end{equation}
in which the second infimum is over the same admissible class. At $\tau=\infty$ we use the deterministic teacher and set $H_\infty=0$. Note that the subtraction of the teacher entropy is essential when comparing approximation errors at different temperatures.

\subsection{Uniform crossover bounds}
\begin{theorem}[Uniform finite-temperature crossover]\label{thm:uniform-temp}
Under the assumptions of Theorem~\ref{thm:main}, uniformly for $\tau>0$ and $2\le p<d$ with $s_p>0$, we have
\begin{equation}
\mathcal E_{p,\tau}\asymp\min\left\{\tau^2s_p^2,\frac{\tau s_p^2}{M},\frac{s_p}{M}\right\}
\label{eq:uniform-temp}
\end{equation}
The comparison constants depend only on the uniform spectral bounds. The lower bound holds for all measurable predictors based on an arbitrary $p$-dimensional Gaussian projection.
\end{theorem}
\begin{proof}
Let $Z=U^\top X$ denote the retained coordinates
and let $Q=\sig(\tau F(X))$ and $Q'=\sig(\tau F(X'))$
be the teacher probabilities associated with the
Gaussian inputs $X$ and $X'$ constructed in
\eqref{eq:coupling}. Conditional on $Z$, the random
variables $Q$ and $Q'$ are independent and identically
distributed. We denote their common conditional
mean by $\bar Q=\E\,[Q\mid Z]$, which is the conditional
probability of the positive binary outcome given
the retained coordinates.

Consider any measurable probability prediction
$r(Z)$ based on the retained coordinates. Since
the cross-entropy is affine in the teacher
probability $Q$ for a fixed prediction $r(Z)$,
its conditional expectation given $Z$ is
\begin{equation*}
 -\bar Q\log r(Z)-(1-\bar Q)\log(1-r(Z))
\end{equation*}
The population regret of this prediction can
therefore be decomposed into two contributions,
as in Proposition~\ref{prop:error-decomposition}.
The first is the increase in the Bayes risk
resulting from the restriction to the retained
coordinates, while the second is the excess
loss of the prediction $r(Z)$ relative to the
optimal prediction $\bar Q$ based on these
coordinates. For logarithmic loss, this
decomposition takes the form
\begin{align*}
 \E\,\KL(\operatorname{Bern}(Q)\|\operatorname{Bern}(r(Z)))=\E\,\KL(\operatorname{Bern}(Q)\|\operatorname{Bern}(\bar Q))
 +\E\,\KL(\operatorname{Bern}(\bar Q)\|\operatorname{Bern}(r(Z)))
\end{align*}
in which the first term is the conditional mutual information and the second term is nonnegative. Conditional independence and the identical conditional laws of $Q,Q'$ mean that $\E\,[(Q-Q')^2\mid Z]=2\Var(Q\mid Z)$. Thus,  Pinsker's inequality (with natural logarithms) \citep[Lemma~2.5]{tsybakov2009} yields, for any predictor $r(Z)$
\begin{equation}
\E\,\KL(\operatorname{Bern}(Q)\|\operatorname{Bern}(r(Z)))
\ge I(Y;X\mid Z)\ge2\E\,\Var(Q\mid Z)=\E\,(Q-Q')^2
\end{equation}
By Proposition~\ref{prop:decomp} and
Lemmas~\ref{lem:interlace} and~\ref{lem:twosided},
both of the conditionally independent residuals
has a conditional probability of at least $c_0$
of exceeding $as_p$ and a conditional probability
of at least $c_0$ of being smaller than $-as_p$.
By conditional independence, the probability that
the first residual exceeds $as_p$ and the second
is smaller than $-as_p$ is therefore at least
$c_0^2$. Restricting the conditional expectation
to this event and using the monotonicity of
$\sig$, we obtain the lower bound
\begin{equation}
 c_0^2\E\,\left[
 \left(\sig(\tau(m(Z)+as_p))-
 \sig(\tau(m(Z)-as_p))\right)^2\right]
\end{equation}

Suppose first that $\tau s_p\leq1$ and consider
the event $\{|m(Z)|\leq1/\tau\}$. By the fundamental
theorem of calculus
\begin{equation*}
 \sig(\tau(m(Z)+as_p))-\sig(\tau(m(Z)-as_p))
 =\int_{\tau(m(Z)-as_p)}^{\tau(m(Z)+as_p)}
 \sig'(v)\dd v
\end{equation*}
On this event, both limits of integration belong
to the interval $[-1-a,1+a]$. Since $\sig'$ is
continuous and strictly positive on this compact
interval, it is bounded below by a positive
constant. The length of the interval of integration
is $2a\tau s_p$ and hence the difference between
the two sigmoid functions is bounded below by
$c\tau s_p$. Consequently, restricting the
expectation to the event $\{|m(Z)|\leq1/\tau\}$
gives a lower bound of the form
\begin{equation*}
 c\tau^2s_p^2\Prob(|m(Z)|\leq1/\tau)
\end{equation*}
Since $\Var(m(Z))\leq2M^2$, the small-ball estimate
in Lemma~\ref{lem:smallball} implies that
\begin{equation*}
 \Prob(|m(Z)|\leq1/\tau)\geq c\min\{1,1/(\tau M)\}
\end{equation*}
and we therefore obtain the lower bound
\begin{equation*}
 c\tau^2s_p^2\min\{1,1/(\tau M)\}
\end{equation*}

Suppose now that $\tau s_p\geq1$ and consider
instead the event $\{|m(Z)|\leq as_p/2\}$. On
this event, the arguments of the two sigmoid
functions satisfy
\begin{equation*}
 \tau(m(Z)+as_p)\geq a/2,
 \qquad \tau(m(Z)-as_p)\leq-a/2
\end{equation*}
and by monotonicity of $\sig$ and the identity
$\sig(x)-\sig(-x)=\tanh(x/2)$, it follows that
\begin{equation*}
 \sig(\tau(m(Z)+as_p))-\sig(\tau(m(Z)-as_p))
 \geq\tanh(a/4)
\end{equation*}
Restricting the expectation to this event and
applying Lemma~\ref{lem:smallball} provides a lower
bound of the form
\begin{equation*}
 c\Prob(|m(Z)|\leq as_p/2)\geq c'\min\{1,s_p/M\}
\end{equation*}
Since $s_p\leq\sqrt{2}M$, this quantity is bounded
below by a positive constant times $s_p/M$.
If $A=0$, then $m(Z)=0$ a.s. and the
events considered in both regimes have probability
one, so that the same lower bounds hold.
These estimates establish the required lower
bound in the two temperature regimes.

To establish the upper bound, set $F=H_p+R_p$,
in which $H_p$ and $R_p$ are the independent retained
and unresolved components associated with spectral
truncation (as in Proposition~\ref{prop:upper}).
Let $\Lambda_p=\diag(\lambda_1,\ldots,\lambda_p,0,\ldots,0)$
denote the rank-$p$ spectral truncation of $\Lambda$
in its eigenbasis and recall that the KL divergence
between Bernoulli distributions with logits $u$
and $v$ admits the Bregman representation
\begin{equation*}
 \KL(\operatorname{Bern}(\sig(u))\|\operatorname{Bern}(\sig(v)))
 =\psi(v)-\psi(u)-\psi'(u)(v-u)
\end{equation*}
For the admissible student $S=\tau\Lambda_p$,
the teacher and student logits are $u=\tau F$
and $v=\tau H_p$, respectively, so that
$v-u=-\tau R_p$. Then applying Taylor's formula with
integral remainder to the Bregman divergence
and taking the expectation yields
\begin{equation}
 \mathcal E_{p,\tau}\leq\tau^2\E\,\left[
 R_p^2\int_0^1(1-t)
 \psi''(\tau(H_p+(1-t)R_p))\dd t\right]
 \label{eq:temp-bregman}
\end{equation}
Using $\psi''\leq1/4$ and $\E\, R_p^2=s_p^2$
we obtain  the upper bound
$\tau^2s_p^2/8$.

A second upper bound can be obtained by exploiting
the uniform density bound for $H_p$. Since $H_p$
and $R_p$ are independent, conditioning on $R_p$
and applying Lemma~\ref{lem:density} leads to
\begin{align*}
 \E\,[\psi''(\tau(H_p+(1-t)R_p))\mid R_p]
 &=\int_\R
 \psi''(\tau(h+(1-t)R_p))\rho_p(h)\dd h\\
 &\leq\frac{K}{\tau}\int_\R\psi''(v)\dd v\\&
 =\frac{K}{\tau}
\end{align*}
in which  we have used the uniform density bound
$\rho_p(h)\leq K$ and the change of variables
$v=\tau(h+(1-t)R_p)$. The final equality follows
from $\psi'=\sig$ and the limiting values of
$\sig$ at $\pm\infty$.
Substituting this conditional expectation bound
into \eqref{eq:temp-bregman}, integrating $1-t$
over $[0,1]$ and using $\E\, R_p^2=s_p^2$ yields
the upper bound $K\tau s_p^2/2$.

To obtain an upper bound appropriate for the
regime $\tau s_p\geq1$, we represent the
probabilistic teacher in terms of a random
perturbation of the deterministic decision
boundary. Let $L$ be a standard logistic random
variable independent of $X$, with zero mean
and variance $\pi^2/3$. Defining the binary
response by
\begin{equation*}
 Y=\one_{\{F+L/\tau>0\}}
\end{equation*}
we recover the probabilistic teacher, since
\begin{equation*}
 \Prob(Y=1\mid X)=1-\sig(-\tau F(X))=\sig(\tau F(X))
\end{equation*}
The binary response can therefore be represented by
a deterministic threshold applied to the augmented
signal $H_p+(R_p+L/\tau)$. The augmented unresolved
component $R_p+L/\tau$ is independent of $H_p$,
has zero mean and variance
\begin{equation*}
 \Var(R_p+L/\tau)=s_p^2+\frac{\pi^2}{3\tau^2}
\end{equation*}
We can therefore apply the argument in the proof
of Proposition~\ref{prop:upper} with $R_p$
replaced by $R_p+L/\tau$. For the admissible
student $S=\Lambda_p/s_p$, this yields
\begin{equation}
 \mathcal E_{p,\tau}\leq{\CE}_\tau(\Lambda_p/s_p)
 \leq K\left[\left(1+\frac{\pi^2}{6}\right)s_p+
 \frac{\pi^2}{3\tau^2s_p}\right]
\end{equation}
in which the first inequality follows from the
nonnegativity of the teacher Bayes risk.
Since $\tau s_p\geq1$, we have
$1/(\tau^2s_p)\leq s_p$ and hence we obtain
the upper bound $K(1+\pi^2/2)s_p$.

It remains to express the upper and lower bounds
in terms of the three scales appearing in
\eqref{eq:uniform-temp}. For $\tau s_p\leq1$,
the two upper bounds obtained from
\eqref{eq:temp-bregman}, together with the
assumed uniform upper bound on $M$, give an
upper bound of the form
\begin{equation*}
 C\min\{\tau^2s_p^2,\tau s_p^2/M\}
\end{equation*}
which agrees with the form of the lower bound
established in this regime. Moreover, since
$\tau s_p\leq1$, we have
$\tau s_p^2/M\leq s_p/M$ and hence
\begin{equation*}
 \min\{\tau^2s_p^2,\tau s_p^2/M,s_p/M\}
 =\min\{\tau^2s_p^2,\tau s_p^2/M\}
\end{equation*}
For $\tau s_p\geq1$, we have matching upper and
lower bounds proportional to $s_p/M$, in which
we have again used the uniform upper bound on
$M$ in the upper estimate. In this regime,
$\tau s_p^2/M\geq s_p/M$. Furthermore, since
$s_p\leq\sqrt{2}M$
\begin{equation*}
 \tau^2s_p^2\geq1\geq\frac{s_p}{\sqrt{2}M}
\end{equation*}
As a consequence,
$\min\{\tau^2s_p^2,\tau s_p^2/M,s_p/M\}$
is comparable to $s_p/M$. Combining the upper
and lower bounds in the two regimes establishes
\eqref{eq:uniform-temp}.
\end{proof}

\subsection{Optimal orientation of the student}
The lower bound in Theorem~\ref{thm:uniform-temp} holds regardless of the orientation of the student eigenspaces relative to those of the teacher. To determine the exact leading-order coefficient, we first establish that for any prescribed student eigenvalues an optimal orientation is obtained by aligning the corresponding eigenvectors with those of the teacher in decreasing eigenvalue order.
\begin{proposition}[Optimal spectral orientation]\label{prop:alignment}
For a continuous strictly increasing teacher link $G$ at finite positive $\tau$, or for the deterministic threshold at $\tau=\infty$, and any canonical student potential $A$, let $\mathcal F_{P,G}(S)=\E\, A(f_S)-\tr(ST_{\tau,G})$, in which $T_{\tau,G}=\E\,[q(XX^\top-I_d)]$ and $q=G(\tau F)$ (or the threshold). For every admissible $S$ there exists a matrix $S^\sharp=\diag(\mu_1,\ldots,\mu_p,0,\ldots,0)$, with the same ordered nonnegative eigenvalues as $S$, such that $\mathcal F_{P,G}(S^\sharp)\le\mathcal F_{P,G}(S)$. Thus the global infimum equals the infimum over nonnegative coefficients in the leading teacher eigenspace.
\end{proposition}
\begin{proof}
For convenience, write $T_\tau=T_{\tau,G}$ and
$q_\tau=q$. We first establish that the term
$\E\, A(f_S)$ in the objective function
$\mathcal F_{P,G}$ depends only on the eigenvalues
of $S$. For any orthogonal matrix $O$, invariance
of the trace under orthogonal conjugation gives $f_{OSO^\top}(X)=f_S(O^\top X)$. Since the standard Gaussian distribution is
invariant under orthogonal transformations,
$O^\top X$ has the same distribution as $X$
and hence
\begin{equation*}
 \E\, A(f_{OSO^\top}(X))=\E\, A(f_S(X))
\end{equation*}
and so the first term in $\mathcal F_{P,G}$ is
independent of the orientation of the eigenvectors
of $S$.

We next establish that $T_\tau$ is diagonal in
the eigenbasis of the teacher matrix $\Lambda$
and that its diagonal entries inherit the ordering
of the teacher eigenvalues. In this basis the
teacher probability $q_\tau$ depends on the input
coordinates only through their squares. For
$i\ne j$, the transformation $X_i\mapsto-X_i$
leaves both the Gaussian distribution and the
teacher probability unchanged but reverses the
sign of $X_iX_j$. It follows that
$\E\,[q_\tau X_iX_j]=0$ and hence $T_\tau$ is
diagonal in the teacher eigenbasis, with diagonal
entries
\begin{equation*}
 t_i=\E\,[q_\tau(X_i^2-1)]
\end{equation*}

To establish the ordering of these entries, fix
$i<j$ and write $U_i=X_i^2-1$, $U_j=X_j^2-1$ and
\begin{equation*}
 C=\sum_{k\notin\{i,j\}}\lambda_k(X_k^2-1)
\end{equation*}
so that $F=C+\lambda_iU_i+\lambda_jU_j$.
Let $g_\tau(x)=G(\tau x)$ for a finite inverse
temperature and $g_\infty(x)=\one_{\{x>0\}}$
for the deterministic teacher. The difference
between the corresponding diagonal entries of
$T_\tau$ is given by
\begin{equation*}
 t_i-t_j=\E\,[(U_i-U_j)g_\tau(C+\lambda_iU_i+\lambda_jU_j)]
\end{equation*}
Since $U_i$ and $U_j$ are i.i.d. and independent of
$C$, interchanging $U_i$ and $U_j$ does not
change their joint distribution with $C$.
Consequently, we also obtain
\begin{equation*}
 t_i-t_j=-\E\,[(U_i-U_j)g_\tau(C+\lambda_iU_j+\lambda_jU_i)]
\end{equation*}
Taking the arithmetic mean of these two
expressions yields
\begin{equation}
 t_i-t_j=\frac12\E\,\big[(U_i-U_j)
 \{g_\tau(C+\lambda_iU_i+\lambda_jU_j)
 -g_\tau(C+\lambda_iU_j+\lambda_jU_i)\}\big]
\end{equation}
The difference between the two arguments of
$g_\tau$ is $(\lambda_i-\lambda_j)(U_i-U_j)$.
Since $\lambda_i\geq\lambda_j$ and $g_\tau$ is
nondecreasing, the integrand is nonnegative.
It follows that $t_i\geq t_j$ and hence the
diagonal entries of $T_\tau$ have the same
nonincreasing ordering as the eigenvalues of
$\Lambda$.

For prescribed eigenvalues
$\mu_1\geq\cdots\geq\mu_d\geq0$ of $S$, the first
term in $\mathcal F_{P,G}$ is independent of the
orientation of $S$. Minimising the objective
function over all orientations is therefore
equivalent to maximising $\tr(ST_\tau)$. The
symmetric matrix trace inequality
\citep{bhatia1997} gives
\begin{equation*}
 \tr(ST_\tau)\leq\sum_{i=1}^d\mu_i t_i
\end{equation*}

This inequality can be
established directly in the present setting.
Writing the spectral decomposition of $S$ in the
teacher eigenbasis as
\begin{equation*}
 S=O\diag(\mu_1,\ldots,\mu_d)O^\top
\end{equation*}
the diagonal entries of $S$ are given by
\begin{equation*}
 a_i=\sum_{j=1}^dO_{ij}^2\mu_j
\end{equation*}
Since $O$ is orthogonal, the matrix with entries
$O_{ij}^2$ is doubly stochastic. In particular,
for every $k=1,\ldots,d$, the weights
$\omega_j=\sum_{i=1}^kO_{ij}^2$ satisfy
$0\leq\omega_j\leq1$ and
$\sum_{j=1}^d\omega_j=k$. Since the eigenvalues
$\mu_j$ are ordered in nonincreasing order,
it follows that
\begin{equation*}
 \sum_{i=1}^k a_i=\sum_{j=1}^d\omega_j\mu_j
 \leq\sum_{j=1}^k\mu_j
\end{equation*}
with equality when $k=d$. Since $T_\tau$ is diagonal in the teacher
eigenbasis, $\tr(ST_\tau)=\sum_{i=1}^da_it_i$.
Then summation by parts and equality of the sums
of the $a_i$ and $\mu_i$ yields
\begin{equation*}
 \sum_{i=1}^d(\mu_i-a_i)t_i
 =\sum_{k=1}^{d-1}(t_k-t_{k+1})
 \sum_{i=1}^k(\mu_i-a_i)\geq0
\end{equation*}
in which the inequality follows from the ordering
of the $t_i$ and the earlier bounds on the
partial sums. This establishes the trace inequality,
with equality attained when $S$ is diagonal in
the teacher eigenbasis and its eigenvalues are
arranged in nonincreasing order.

Since $S$ is positive semidefinite and has rank
at most $p$, we have $\mu_i=0$ for $i>p$.
Thus, for any admissible matrix $S$,
there exists a matrix with the same eigenvalues
whose range is contained in the leading
$p$-dimensional eigenspace of $\Lambda$ and
whose objective function is no greater than
that of $S$. The remaining optimisation is
therefore over the nonnegative eigenvalues
associated with these leading teacher
eigenvectors, which completes the proof.
\end{proof}

\begin{remark}
Proposition \ref{prop:alignment} determines an optimal orientation
of the student eigenspaces but does not determine
the magnitudes of the corresponding student
eigenvalues. The reduced objective is strictly
convex in the retained student eigenvalues.
Indeed, by the Gaussian moment identity \eqref{eq:gaussian-quadratic-variance}, the difference between
two centred Gaussian quadratic forms associated
with distinct coefficient vectors has a strictly
positive second moment and is therefore nonzero
on a set of positive probability. Strict convexity
of $A$ then implies strict convexity of the expected
potential, which is preserved upon subtraction
of the linear term in the reduced objective.
Furthermore, for finite positive $\tau$, strict
monotonicity of $G$ implies that $t_i>t_j$ whenever
$\lambda_i>\lambda_j$, as follows directly from
the exchangeability argument in the proof.
In deriving the sharp asymptotic result below,
we do not impose any proportionality constraint
between the retained teacher eigenvalues and
the corresponding student eigenvalues.
\end{remark}

\subsection{Fixed spectra and the central density}
The uniform estimates above involve the spectral scale
$M=\Frob{\Lambda}$ and do not require a limiting density.
In contrast, the sharp asymptotic coefficient below involves
the limiting density at zero of the leading spectral signal,
not the density of the conditional mean associated with an
arbitrary retained subspace.

Having established uniform two-sided bounds on the
optimal rank-constrained population regret, we now
seek to determine its precise leading-order
behaviour as the rank of the student increases
and the variance of the unresolved component
tends to zero. Of particular interest is the
dependence of the leading-order regret on the
relative magnitudes of the unresolved fluctuations
and the temperature of the teacher. To derive
sharp asymptotic equivalents in this joint limit,
we impose a stronger spectral assumption
than~\eqref{eq:uniform-spectral-assumptions}.
Specifically, we assume that the finite-dimensional
teacher matrices are spectral truncations of a
fixed nonincreasing nonnegative sequence
$(\lambda_j)_{j\geq1}$ satisfying
$\sum_{j\geq1}\lambda_j^2<\infty$ and having
infinitely many positive eigenvalues.

Throughout the following analysis we consider
the limit $p\to\infty$ with $d=d(p)>p$ and write
$s_p=s_{p,d}$ for the standard deviation of the
unresolved component. Under the above spectral
assumption
\begin{equation*}
 0<s_p^2=2\sum_{j=p+1}^{d(p)}\lambda_j^2
 \leq2\sum_{j=p+1}^{\infty}\lambda_j^2\to0
\end{equation*}
Furthermore, since the random variables
$\lambda_j(X_j^2-1)$ are independent and centred
and the sum of their variances is finite, the
infinite-dimensional teacher signal
\begin{equation*}
 F_\infty=\sum_{j\geq1}\lambda_j(X_j^2-1)
\end{equation*}
converges a.s. and in $L^2$ by the
convergence theorem for independent centred
square-integrable series
\citep[Section~22]{billingsley1995}.
Note that square summability of the eigenvalues
is sufficient for this convergence and we do not
require $\sum_{j\geq1}\lambda_j<\infty$.

\begin{lemma}[Central density convergence]\label{lem:central-density}
The densities $\rho_p$ of $H_p$ converge uniformly on $\R$ to a bounded continuous density $\rho_\infty$ of $F_\infty$, and $\rho_0:=\rho_\infty(0)>0$.
\end{lemma}
\begin{proof}
Independence of the Gaussian coordinates and the characteristic function of a centred chi-squared variable lead to
\begin{equation*}
 |\varphi_p(t)|=\prod_{j\le p}(1+4t^2\lambda_j^2)^{-1/4}\le\prod_{j=1}^3(1+4t^2\lambda_j^2)^{-1/4},\qquad p\ge3
\end{equation*}
The majorant is bounded near 0 and decays as $|t|^{-3/2}$ at infinity, since the first three eigenvalues are fixed and positive. It therefore belongs to $L^1(\R)$. Convergence of $H_p$ to $F_\infty$ in $L^2$ implies pointwise convergence of the characteristic functions to $\varphi_\infty$. The same majorant bounds the limit, so dominated convergence yields $\|\varphi_p-\varphi_\infty\|_{L^1}\to0$. By the characteristic-function inversion theorem for integrable characteristic functions \citep[Section~26]{billingsley1995}, these laws admit continuous densities and
\begin{equation*}
 \sup_x|\rho_p(x)-\rho_\infty(x)|\le(2\pi)^{-1}\|\varphi_p-\varphi_\infty\|_{L^1}\longrightarrow0
\end{equation*}
The proof of Lemma~\ref{lem:smallball} yields a uniform positive density lower bound a.e. in a fixed neighbourhood of zero, since the variances of $H_p$ are bounded above and away from zero. Continuity and uniform convergence yield $\rho_0>0$.
\end{proof}

We can now apply the abstract approximation results
of Section~\ref{sec:abstract-approx} to the spectral
decomposition $F=H_p+R_p$ by making the identifications
\begin{equation*}
 H_\varepsilon=H_p,\qquad \varepsilon=s_p,\qquad
 Z_\varepsilon=R_p/s_p
\end{equation*}
The retained signal and the normalised unresolved
fluctuation are independent since they depend on
disjoint sets of independent Gaussian coordinates.
Furthermore, the fourth-moment estimate established
in the proof of Lemma~\ref{lem:twosided} gives
$\E\,[(R_p/s_p)^4]\leq81$, which ensures uniform
integrability of the squared normalised unresolved
fluctuations. Lemma~\ref{lem:central-density}
establishes the remaining assumptions regarding 
the uniform boundedness and local convergence of
the retained signal densities, with a strictly
positive limiting density at zero. Under the
general canonical assumptions, we can therefore
apply Theorem~\ref{thm:abstract-localisation} to
obtain an additive asymptotic expansion of the
minimum population regret over students whose
canonical coordinates are affine functions of
$H_p$. When the teacher and student inverse links
coincide and the additional smoothness assumptions
of Theorem~\ref{thm:abstract-sharp} are satisfied,
we also obtain a sharp asymptotic equivalent for
the minimum regret and an asymptotically optimal
calibration within this class of students.
\begin{remark}
The representation and decoder error decomposition
in Theorem~\ref{thm:decomposition} also applies
to the representation consisting of the leading
$p$ Gaussian coordinates. Indeed, since $R_p$ is
independent of these coordinates, they influence the conditional
teacher probability depends only through
the scalar retained signal $H_p$. The decomposition
therefore characterises the information loss
associated with spectral truncation and the
additional regret resulting from restricting the
decoder to an affine canonical coordinate based
on $H_p$. It does not, however, establish that
spectral truncation minimises the representation
error among all $p$-dimensional linear projections
of the input.
\end{remark}
It remains to determine whether the additional
freedom afforded by allowing the student eigenvalues
to vary independently can reduce the leading-order
regret. Although the spectral alignment result
allows us to restrict the optimisation to matrices
supported on the leading teacher eigenspace, it
does not require the student eigenvalues to be
proportional to those of the teacher. To establish
the sharp asymptotic minimum over the full admissible
matrix class, we therefore develop a conditional
argument based on Proposition~\ref{prop:canonical-integrated}.
This requires a version of the integrated variational
identity adapted to conditional densities that are not necessarily 
 uniformly bounded.
\subsection{Integrated boundary-layer loss}
Let $G_p=R_p/s_p$, $\eta_p=\tau_ps_p$ and $T_p=G_p-L_P/\eta_p$, with $1/\eta_p=0$ for deterministic labels and $L_P$ independent of the Gaussian variables. Then $\E \,T_p=0$ and $V_p:=\E \,T_p^2=1+v_P/\eta_p^2$. If $\eta_p$ is bounded away from zero, the finite latent variance and Lemma~\ref{lem:twosided} imply uniform integrability of $T_p^2$. Set $u=H_p/s_p$, then for an affine canonical coordinate $bu+c=b(u-a)$ with $b>0$ and $a=-c/b$, define the expected loss (before subtracting the teacher Bayes risk)
\begin{equation}
\ell_p(u;b,c)=\E\,[A(bu+c)-(bu+c)\one_{\{u+T_p>0\}}]
\end{equation}
\begin{lemma}[Integrated affine loss]\label{lem:integrated-loss}
For every $b>0$ and $c\in\R$,
\begin{equation}
C_p(b,c):=\int_\R\ell_p(u;b,c)\dd u=\frac{v_P}{2b}+\frac{bV_p}{2}+\frac{c^2}{2b}
\label{eq:integrated-loss}
\end{equation}
\end{lemma}
\begin{proof}
The result  follows from Proposition~\ref{prop:canonical-integrated} for the canonical link and the threshold $-T_p$, whose mean is 0 and whose variance is $V_p$. 
\end{proof}
If $\eta_p\to\eta\in(0,\infty]$, the functions $C_p$ converge locally uniformly on $(0,\infty)\times\R$ to $C_\eta$, obtained by replacing $V_p$ by $V_\eta=1+v_P/\eta^2$. The unique minimum is attained at
\begin{equation}
b_*(\eta)=\sqrt{v_P/V_\eta},\qquad c_*=0,\qquad C_\eta^*=\sqrt{v_PV_\eta}
\label{eq:optimal-scale}
\end{equation}
The absence of a distributional assumption on $G_p$ is important: the integrated loss depends on the residual only through its first two moments.

\begin{lemma}[Uniform truncated optimisation]\label{lem:truncated}
Suppose $\eta_p\to\eta\in(0,\infty]$. Let $C_{p,R}(b,c)=\int_{-R}^R\ell_p(u;b,c)\dd u$ and $m_{p,R}=\inf_{b\ge0,c\in\R}C_{p,R}(b,c)$. Then
\begin{equation}
\lim_{R\to\infty}\liminf_{p\to\infty}m_{p,R}=C_\eta^*
\label{eq:truncated}
\end{equation}
\end{lemma}
\begin{proof}
We first establish that the error introduced by
truncating the domain of integration tends to zero
uniformly on compact subsets of the parameter
space $(0,\infty)\times\R$. On any such compact
subset we can choose constants $0<b_-\leq b_+$
and $A_+<\infty$ such that
\begin{equation*}
 b_-\leq b\leq b_+,\qquad |a|\leq A_+
\end{equation*}
in which $a=-c/b$ is the student decision boundary.
Using the threshold-loss decomposition established
in the proof of Theorem~\ref{thm:abstract-localisation},
we can write
\begin{equation*}
 A(bu+c)-\one_{\{u>-T_p\}}(bu+c)
 =r_P(bu+c)+|bu+c|\one_{\{u\in I_p\}}\quad \text{for a.e. }u
\end{equation*}
in which $r_P(x)=A(x)-x_+$
and $I_p=\bigl(\min\{-T_p,a\},\max\{-T_p,a\}\bigr)$. We estimate separately the contributions
of these two terms to the integral over $|u|>R$.

For $R>A_+$, the change of variables
$x=bu+c=b(u-a)$ leads to
\begin{equation*}
 \int_{|u|>R}r_P(bu+c)\dd u
 \leq\frac{1}{b_-}
 \int_{|x|>b_-(R-A_+)}r_P(x)\dd x
\end{equation*}
since $|u-a|\geq|u|-|a|$. The right-hand side
tends to 0 as $R\to\infty$ by the integrability
of $r_P$. To estimate the contribution of the second term,
note that since $|a|\leq A_+<R$, the interval
$I_p$ can intersect the region $|u|>R$ only if
$|T_p|>R$. Furthermore, integrating $|bu+c|$
over the entire interval $I_p$ gives
\begin{equation*}
 \int_{I_p}|bu+c|\dd u
 =\frac{b}{2}|T_p+a|^2
 \leq\frac{b_+}{2}(|T_p|+A_+)^2
\end{equation*}
 and consequently
\begin{equation*}
 \int_{|u|>R}|bu+c|\one_{\{u\in I_p\}}\dd u
 \leq\frac{b_+}{2}(|T_p|+A_+)^2
 \one_{\{|T_p|>R\}}
\end{equation*}
Taking expectations and using the uniform
integrability of $T_p^2$, we conclude that the
contribution of the second term also tends to
zero as $R\to\infty$, uniformly in $p$ and in
the parameters $(b,c)$ belonging to the compact
set under consideration. Combining the two
estimates, we obtain
\begin{equation*}
 \lim_{R\to\infty}\sup_p\sup_{(b,c)\in\mathcal K}
 |C_{p,R}(b,c)-C_p(b,c)|=0
\end{equation*}
for every compact subset $\mathcal K$ of
$(0,\infty)\times\R$.

Although we have established uniform approximation
on compact subsets of the parameter space, this
does not by itself ensure convergence of the
global minimum values, since minimising sequences
could leave every compact subset. We therefore
use convexity to establish that, for sufficiently
large $R$ and $p$, the global minimum of $C_{p,R}$
is attained in a fixed compact neighbourhood of
the unique minimiser of the limiting objective
$C_\eta$. Let $z_*=(b_*(\eta),0)$ denote the unique minimiser
of $C_\eta$ and choose a closed ball $\mathcal K$
centred at $z_*$ and contained in the open
half-plane $b>0$. By continuity of $C_\eta$,
compactness of $\partial\mathcal K$ and uniqueness
of the minimiser $z_*$
\begin{equation*}
 \kappa=\min_{z\in\partial\mathcal K}
 \bigl(C_\eta(z)-C_\eta(z_*)\bigr)>0
\end{equation*}
By the uniform truncation
estimate established above and the uniform
convergence of $C_p$ to $C_\eta$ on $\mathcal K$,
we can choose $R$ sufficiently large and subsequently
$p$ sufficiently large such that
\begin{equation*}
 \sup_{z\in\mathcal K}|C_{p,R}(z)-C_\eta(z)|
 <\frac{\kappa}{3}
\end{equation*}
It follows that for every $z\in\partial\mathcal K$
\begin{equation*}
 C_{p,R}(z)-C_{p,R}(z_*)
 \geq C_\eta(z)-C_\eta(z_*)-\frac{2\kappa}{3}
 \geq\frac{\kappa}{3}>0
\end{equation*}
so the value of $C_{p,R}$ at every point on
the boundary of $\mathcal K$ is strictly greater
than its value at the centre $z_*$.

Since the integrand defining $C_{p,R}$ is convex
in $(b,c)$, the truncated objective is convex on
the closed half-plane $b\geq0$. Suppose that there
exists a point $z$ outside $\mathcal K$ satisfying
$C_{p,R}(z)\leq C_{p,R}(z_*)$. The line segment
joining $z_*$ and $z$ would intersect
$\partial\mathcal K$ at a point
$z_\theta=(1-\theta)z_*+\theta z$ for some
$\theta\in(0,1)$. By convexity
\begin{equation*}
 C_{p,R}(z_\theta)
 \leq(1-\theta)C_{p,R}(z_*)+\theta C_{p,R}(z)
 \leq C_{p,R}(z_*)
\end{equation*}
which contradicts the strict inequality 
on $\partial\mathcal K$. It therefore follows that no point
outside $\mathcal K$ can have an objective value
less than or equal to that at $z_*$. Continuity
of $C_{p,R}$ on the compact set $\mathcal K$
ensures attainment of its minimum on this set
and the strict boundary inequality ensures that
the minimum is attained in its interior. This
minimum is thus also a global minimum over
the closed half-plane $b\geq0$.

Since both the limiting and truncated objectives
attain their global minima in $\mathcal K$, their
minimum values satisfy
\begin{equation*}
 |m_{p,R}-C_\eta^*|
 \leq\sup_{z\in\mathcal K}
 |C_{p,R}(z)-C_\eta(z)|
\end{equation*}
and using the triangle inequality, we obtain
\begin{equation*}
 |m_{p,R}-C_\eta^*|
 \leq\sup_{z\in\mathcal K}|C_{p,R}(z)-C_p(z)|
 +\sup_{z\in\mathcal K}|C_p(z)-C_\eta(z)|
\end{equation*}
The second term approaches 0 as $p\to\infty$
by virtue of uniform convergence on compact subsets,
while the first term tends to 0 uniformly
in $p$ as $R\to\infty$ by the truncation estimate.
Taking the limits in this order gives
\begin{equation*}
 \lim_{R\to\infty}\limsup_{p\to\infty}|m_{p,R}-C_\eta^*|=0
\end{equation*}
which establishes \eqref{eq:truncated}, also with the
inner $\liminf$ replaced by $\limsup$.
\end{proof}

\subsection{Exact global crossover}
For a matched canonical pair as in Section~\ref{sec:canonical}, we define the optimal population regret over the full matrix class by
\begin{equation}
 \mathcal E^P_{p,\tau}=\inf_{\substack{S\succeq0\\\rank S\le p}}\E\, D_P(q,f_S)
\end{equation}
in which $q=P(\tau F)$ for finite $\tau$ and $q=\one_{\{F>0\}}$ otherwise. When $P=\sig$, this is the excess cross-entropy introduced previously. No differentiability beyond the canonical assumptions is required for the following theorem.
\begin{theorem}[Exact crossover at positive or infinite rescaled inverse temperature]\label{thm:exact-crossover}
Under the fixed-spectrum assumptions and the canonical assumptions of Section~\ref{sec:canonical}, if $\eta_p=\tau_ps_p\to\eta\in(0,\infty]$, then
\begin{equation}
\mathcal E^P_{p,\tau_p}\sim\rho_0s_p\Phi_P(\eta),\qquad
\Phi_P(\eta)=\frac{\eta}{1+\sqrt{1+\eta^2/v_P}},\qquad \Phi_P(\infty)=\sqrt{v_P}
\label{eq:exact-crossover}
\end{equation}
The scaled spectral truncation $S=\alpha_p^P\Lambda_p$
is asymptotically optimal in relative excess risk
when
\begin{equation}
 \alpha_p^P=\left(\tau_p^{-2}+s_p^2/v_P\right)^{-1/2}
 \label{eq:canonical-spectral-calibration}
\end{equation}
For the deterministic teacher, corresponding to
$\tau_p=\infty$, the calibration reduces to
$\alpha_p^P=\sqrt{v_P}/s_p$.
\end{theorem}
\begin{proof}
We first establish an upper bound on the minimum
population regret by constructing a student that
attains the minimum of the limiting integrated
loss. Consider the scaled spectral truncation
$S=b_*(\eta_p)\Lambda_p/s_p$, in which
$b_*(\eta_p)\to b_*(\eta)>0$. The canonical
coordinate of this student is
$f_S=b_*(\eta_p)H_p/s_p$ and hence the coefficient
of the retained signal $H_p$ is
\begin{equation*}
 \frac{b_*(\eta_p)}{s_p}
 =\frac{1}{s_p}\sqrt{\frac{v_P}{1+v_P/\eta_p^2}}
 =\left(\tau_p^{-2}+s_p^2/v_P\right)^{-1/2}
\end{equation*}
which gives the calibration in
\eqref{eq:canonical-spectral-calibration}.
By conditioning on $H_p$ and introducing the
rescaled retained signal $u=H_p/s_p$, the expected
canonical loss of this student, before subtracting
 the teacher Bayes risk, can be written as
\begin{equation*}
 \frac{\mathcal F_{P,P}(S)}{s_p}
 =\int_\R\rho_p(s_pu)\ell_p(u;b_*(\eta_p),0)\dd u
\end{equation*}
By Lemma~\ref{lem:central-density},
$\rho_p(s_pu)\to\rho_0$ uniformly on bounded
intervals of $u$. Moreover, since
$b_*(\eta_p)\to b_*(\eta)>0$, the slopes
$b_*(\eta_p)$ belong to a compact interval
contained in $(0,\infty)$ for all sufficiently
large $p$. The tail estimates established in
the proof of Lemma~\ref{lem:truncated}, together
with the uniform density bound, therefore imply
that
\begin{equation*}
 \lim_{R\to\infty}\limsup_{p\to\infty}
 \int_{|u|>R}\rho_p(s_pu)
 \ell_p(u;b_*(\eta_p),0)\dd u=0
\end{equation*}
For any fixed $R$, the error incurred by replacing
$\rho_p(s_pu)$ by $\rho_0$ in the integral over
$[-R,R]$ is bounded by
\begin{equation*}
 \sup_{|u|\leq R}|\rho_p(s_pu)-\rho_0|\,
 C_p(b_*(\eta_p),0)
\end{equation*}
and since $C_p(b_*(\eta_p),0)\to C_\eta^*$ by
\eqref{eq:integrated-loss}, the above expression
tends to 0 as $p\to\infty$. Combining the
local density convergence with the uniform
tail estimates furnishes
\begin{equation*}
 \lim_{p\to\infty}
 \frac{\mathcal F_{P,P}(b_*(\eta_p)\Lambda_p/s_p)}{s_p}
 =\rho_0 C_\eta^*
\end{equation*}
Since the scaled spectral truncation is an
admissible student matrix, it follows that
\begin{equation*}
 \limsup_{p\to\infty}
 \frac{\inf_S\mathcal F_{P,P}(S)}{s_p}
 \leq\rho_0 C_\eta^*
\end{equation*}

To establish a matching lower bound over the
full admissible matrix class, we first use
Proposition~\ref{prop:alignment} to restrict
the optimisation to student matrices that
are diagonal in the teacher eigenbasis and
supported on the leading $p$ teacher eigenvectors.
The difficulty here is that the retained student
eigenvalues are not necessarily proportional to those
of the teacher, so the canonical coordinate
of an arbitrary aligned student cannot in
general be expressed as a function of $H_p$
alone. We overcome this difficulty by conditioning
on all but one of the retained Gaussian
coordinates.

Let $Z_p=(X_2,\ldots,X_p)$ and decompose the
retained signal as $H_p=A_1+B_p$, in which
\begin{equation*}
 A_1=\lambda_1(X_1^2-1),\qquad
 B_p=\sum_{j=2}^p\lambda_j(X_j^2-1)
\end{equation*}
For an aligned student with retained eigenvalues
$\mu_1,\ldots,\mu_p$, substituting
$X_1^2-1=(H_p-B_p)/\lambda_1$ into its canonical
coordinate yields
\begin{equation*}
 f_S=\frac{\mu_1}{\lambda_1}H_p+
 \sum_{j=2}^p(\mu_j-\mu_1\lambda_j/\lambda_1)(X_j^2-1)
\end{equation*}
Thus, conditional on $Z_p$, the student
canonical coordinate is an affine function
$bu+c$ of the rescaled retained signal $u=H_p/s_p$,
with coefficients
\begin{equation*}
 b=\frac{s_p\mu_1}{\lambda_1}\geq0,\qquad
 c=\sum_{j=2}^p(\mu_j-\mu_1\lambda_j/\lambda_1)(X_j^2-1)
\end{equation*}
For a given student matrix, the slope $b$ is
fixed, while the intercept $c$ depends on
the conditioned coordinates $Z_p$. To obtain
a lower bound, we relax the constraints imposed
by the original matrix parametrisation and
allow both $b$ and $c$ to be arbitrary measurable
functions of $Z_p$, subject only to $b\geq0$.
Since this enlarges the class of admissible
conditional predictions, the corresponding
minimum expected loss cannot exceed that of
the original rank-constrained problem.

Since $A_1$ is independent of $Z_p$ and $B_p$
is determined by $Z_p$, the conditional density
of $H_p$ given $Z_p$ is $g(h-B_p)$, in which $g$
is the density of $A_1$. Using the density of
a chi-squared random variable with one degree
of freedom, we obtain
\begin{equation}
 g(x)=
 \begin{cases}
 \displaystyle\frac{\exp(-(x+\lambda_1)/(2\lambda_1))}
 {\sqrt{2\pi\lambda_1(x+\lambda_1)}} & x>-\lambda_1\\[6pt]
 0 & x\leq-\lambda_1
 \end{cases}
\end{equation}
In contrast to the unconditional density of $H_p$,
the conditional density is unbounded near the
lower endpoint of its support. Nevertheless,
the a.s. convergence of the infinite
Gaussian quadratic series established above
implies that
\begin{equation*}
 B_p\to B_\infty=\sum_{j=2}^\infty
 \lambda_j(X_j^2-1)\quad \text{a.s.}
\end{equation*}
Since the law of $B_\infty$ is absolutely
continuous, the event $B_\infty=\lambda_1$
has probability zero. On the complement of this
event, $g$ is continuous in a neighbourhood of
$-B_\infty$. Moreover, for any fixed $R<\infty$
\begin{equation*}
 \sup_{|u|\leq R}|s_pu-B_p+B_\infty|
 \leq s_pR+|B_p-B_\infty|\to0\quad \text{a.s.}
\end{equation*}
so it follows that
\begin{equation*}
 \sup_{|u|\leq R}
 |g(s_pu-B_p)-g(-B_\infty)|\to0\quad \text{a.s.}
\end{equation*}

Since the omitted residual is independent of all
the retained Gaussian coordinates, conditioning
on both $Z_p$ and $H_p$ yields the same conditional
teacher probability as conditioning on $H_p$
alone. In terms of the rescaled retained signal
$u=H_p/s_p$, this probability is given by the
distribution function of $-T_p$, with the
appropriate strict threshold convention.
Consequently, the conditional expected canonical
loss of an aligned student, scaled by $s_p$,
is given by
\begin{equation*}
 \int_\R g(s_pu-B_p)\ell_p(u;b,c)\dd u
\end{equation*}
By nonnegativity of the canonical loss, restricting
the domain of integration to $[-R,R]$ and bounding
the conditional density below on this interval
leads to
\begin{align*}
 \int_\R g(s_pu-B_p)\ell_p(u;b,c)\dd u
 &\geq
 \left(\inf_{|u|\leq R}g(s_pu-B_p)\right)
 \int_{-R}^R\ell_p(u;b,c)\dd u\\
 &\geq
 m_{p,R}\inf_{|u|\leq R}g(s_pu-B_p)
\end{align*}
in which $m_{p,R}$ is the minimum value of the
truncated integrated loss over all nonnegative
slopes and arbitrary intercepts, as defined
in Lemma~\ref{lem:truncated}. Since this
lower bound is independent of the student
coefficients, taking the expectation with
respect to $Z_p$ and then the infimum over
all admissible student matrices yields
\begin{equation}
 \frac{\inf_S\mathcal F_{P,P}(S)}{s_p}
 \geq m_{p,R}\,\E\,\left[
 \inf_{|u|\leq R}g(s_pu-B_p)\right]
\end{equation}

To justify the expectation on the right-hand
side, we note that $g=0$  below and at
the lower endpoint of its support, as well as  strictly
decreasing above this endpoint. Therefore
\begin{equation*}
 \inf_{|u|\leq R}g(s_pu-B_p)=
 \begin{cases}
 0 & -s_pR-B_p\leq-\lambda_1\\
 g(s_pR-B_p) & -s_pR-B_p>-\lambda_1
 \end{cases}
\end{equation*}
which establishes measurability of the infimum.
Furthermore, the a.s. uniform convergence
of the conditional densities on $[-R,R]$
established above implies that
\begin{equation*}
 \inf_{|u|\leq R}g(s_pu-B_p)\to g(-B_\infty)\quad \text{a.s.}
\end{equation*}
and since the conditional densities
are nonnegative, Fatou's lemma yields
\begin{equation*}
 \liminf_{p\to\infty}\E\,\left[
 \inf_{|u|\leq R}g(s_pu-B_p)\right]
 \geq\E\, g(-B_\infty)
\end{equation*}

We next establish that the expectation on the
right-hand side is equal to $\rho_0$, the
density of the limiting teacher signal at zero.
Since $A_1$ and $B_\infty$ are independent,
the density of $F_\infty=A_1+B_\infty$ is given
a.e. by the convolution $g*f_B$,
in which $f_B$ denotes the density of $B_\infty$.
Applying the density argument of Lemma~\ref{lem:density}
to the summands with coefficients $\lambda_2$ and
$\lambda_3$ and convolving with the law of the independent
remaining infinite series gives
$\|f_B\|_\infty\leq 1/(2\sqrt{\lambda_2\lambda_3})$. In particular
\begin{align*}
 |(g*f_B)(h)-(g*f_B)(0)|
 &\leq\int_\R|g(h-x)-g(-x)|f_B(x)\dd x\\
 &\leq\|f_B\|_\infty
 \|g(h+\cdot)-g\|_{L^1}
\end{align*}
Now, since $g\in L^1(\R)$, continuity of translations
in $L^1(\R)$ implies that the right-hand converges to 0 as $h\to0$. The same argument
applies at every $h\in\R$ and hence $g*f_B$
is continuous. Since $\rho_\infty$ is also a
continuous density of $F_\infty$, the two
densities must coincide pointwise (not just a.e.). Evaluating
the convolution at zero, we obtain
\begin{equation*}
 \E\, g(-B_\infty)=\int_\R g(-x)f_B(x)\dd x
 =(g*f_B)(0)=\rho_\infty(0)=\rho_0
\end{equation*}
For each fixed $R$, Fatou's lemma implies that for
any $0<\zeta<\rho_0$ the expectation of the conditional
density infimum is at least $\rho_0-\zeta$ for all
sufficiently large $p$. Since $m_{p,R}\geq0$, taking
the lower limit of the product and then letting
$\zeta\downarrow0$ yields
\begin{equation*}
 \liminf_{p\to\infty}
 \frac{\inf_S\mathcal F_{P,P}(S)}{s_p}
 \geq\rho_0\liminf_{p\to\infty}m_{p,R}
\end{equation*}
and letting $R\to\infty$ and applying
Lemma~\ref{lem:truncated}, we obtain
\begin{equation*}
 \liminf_{p\to\infty}
 \frac{\inf_S\mathcal F_{P,P}(S)}{s_p}
 \geq\rho_0 C_\eta^*
\end{equation*}
which matches the upper bound obtained using
the calibrated spectral truncation. Consequently
\begin{equation*}
 \lim_{p\to\infty}
 \frac{\inf_S\mathcal F_{P,P}(S)}{s_p}
 =\rho_0 C_\eta^*
\end{equation*}

Now we subtract the teacher Bayes risk
from the minimum expected canonical loss in
order to obtain the asymptotic population
regret. For finite inverse temperature $\tau_p$,
the teacher Bayes risk is
\begin{equation*}
 H_{\tau_p}^P=\E\, B_P(P(\tau_p F))
\end{equation*}
Writing $F=H_p+s_pG_p$, in which $G_p=R_p/s_p$
is the normalised unresolved component, and
using independence of $H_p$ and $G_p$, we obtain
\begin{equation*}
 H_{\tau_p}^P=\E\,\left[
 \int_\R B_P(P(\tau_p(h+s_pG_p)))\rho_p(h)\dd h
 \right]
\end{equation*}
Introducing the variable $v=\tau_p(h+s_pG_p)$
and using $\eta_p=\tau_ps_p$ gives
\begin{equation}
 \frac{H_{\tau_p}^P}{s_p}
 =\frac{1}{\eta_p}\int_\R B_P(P(v))
 \E\,\left[\rho_p\left(s_p(v/\eta_p-G_p)\right)\right]\dd v
\end{equation}

Suppose first that $\eta_p\to\eta\in(0,\infty)$.
Since $\E \,G_p^2=1$, the sequence of normalised
unresolved components is tight. For each fixed
$v$, it follows from $s_p\to0$ and
$\eta_p\to\eta>0$ that
$s_p(v/\eta_p-G_p)\xrightarrow{\Prob}0$. 
The local uniform convergence of $\rho_p$ to
the limiting density, together with the
uniform density bound, therefore implies that
\begin{equation*}
 \E\,\left[\rho_p\left(s_p(v/\eta_p-G_p)\right)\right]
 \to\rho_0
\end{equation*}
Since $\rho_p\leq K$ and
$\int_\R B_P(P(v))\dd v=v_P$, the dominated
convergence theorem gives
\begin{equation*}
 \lim_{p\to\infty}\frac{H_{\tau_p}^P}{s_p}
 =\frac{\rho_0v_P}{\eta}
\end{equation*}
If instead $\eta_p\to\infty$, the uniform
density bound gives
\begin{equation*}
 0\leq\frac{H_{\tau_p}^P}{s_p}
 \leq\frac{Kv_P}{\eta_p}\to0
\end{equation*}
for finite $\tau_p$. At the deterministic
endpoint $\tau_p=\infty$, the teacher Bayes
risk is identically 0 and hence the same
limiting result holds.

Subtracting the teacher Bayes risk from the
minimum expected canonical loss and using
the expression for $C_\eta^*$, we obtain, for
$0<\eta<\infty$
\begin{align*}
 \lim_{p\to\infty}\frac{\mathcal E^P_{p,\tau_p}}{s_p}
 =\rho_0\left(\sqrt{v_P}\sqrt{1+v_P/\eta^2}
 -v_P/\eta\right)
 =\rho_0\frac{\eta}{1+\sqrt{1+\eta^2/v_P}}
\end{align*}
For $\eta=\infty$, the corresponding limit is
$\rho_0\sqrt{v_P}$. Since these limiting
coefficients are strictly positive, the limits
are equivalent to the sharp asymptotic
equivalents in \eqref{eq:exact-crossover}.
Moreover, since the scaled spectral truncation
attains the same limiting expected canonical
loss as the global minimum and the teacher
Bayes risk is independent of the student
matrix, its population regret has the same
strictly positive leading-order coefficient
as the optimal population regret. This
establishes the relative asymptotic optimality
of the calibration in
\eqref{eq:canonical-spectral-calibration}.
\end{proof}
For logarithmic loss, the canonical latent
distribution is the standard logistic distribution,
for which $v_P=\pi^2/3$. We therefore retain the
notation $\Phi=\Phi_{\sig}$ and write the
corresponding spectral calibration as
\begin{equation}
 \alpha_p=(\tau_p^{-2}+3s_p^2/\pi^2)^{-1/2}
 \label{eq:calibration}
\end{equation}

\begin{corollary}[Global sharp canonical crossover]
\label{cor:canonical-global}
Under the fixed-spectrum assumptions and the
canonical assumptions of Section~\ref{sec:canonical},
suppose that the teacher and student inverse links
coincide and that $P\in C^2(\R)$ with $P'>0$.
Then along every joint asymptotic path with
$p\to\infty$ and $\tau_p\to\infty$, irrespective
of the relative rates at which $s_p$ and
$\tau_p^{-1}$ tend to zero, we have
\begin{equation}
 \mathcal E^P_{p,\tau_p}\sim
 \rho_0\frac{\tau_ps_p^2}
 {1+\sqrt{1+\tau_p^2s_p^2/v_P}}
\end{equation}
The scaled spectral truncation
$S=\alpha_p^P\Lambda_p$, with $\alpha_p^P$
given by \eqref{eq:canonical-spectral-calibration},
is asymptotically optimal in relative excess risk.
For deterministic labels, the corresponding
sharp asymptotic equivalent is
$\rho_0\sqrt{v_P}s_p$, without the additional
differentiability assumptions on $P$.
\end{corollary}
\begin{proof}
For $\eta_p=\tau_ps_p\to\eta\in(0,\infty]$,
the assertion follows directly from
Theorem~\ref{thm:exact-crossover}. It therefore
remains to establish the sharp asymptotic
equivalent in the regular regime $\eta_p\to0$.

By Proposition~\ref{prop:alignment}, we can
restrict the optimisation to student matrices
supported on the leading $p$ teacher eigenvectors.
Let $Z_p=(X_1,\ldots,X_p)$ denote the corresponding
retained Gaussian coordinates. Since the omitted
residual $R_p$ is independent of $Z_p$ and the
retained signal $H_p$ is a function of $Z_p$, we have
\begin{equation*}
 \E\,[P(\tau_p F)\mid Z_p]
 =\E\,[P(\tau_p(H_p+R_p))\mid H_p]
\end{equation*}
Thus the conditional teacher probability given
the full vector of retained coordinates depends
on these coordinates only through the scalar
retained signal $H_p$. Since the optimal
measurable probability prediction under a
strictly proper loss is the conditional teacher
probability, the minimum population regret over
all measurable predictions based on $Z_p$ is
identical to that over all measurable predictions
based on $H_p$ alone.

Every aligned quadratic student defines a
measurable probability prediction based on $Z_p$.
As a consequence, the minimum regret over all such
measurable predictions provides a lower bound
on the optimal rank-constrained population
regret. Conversely, since the calibrated spectral
truncation in \eqref{eq:canonical-spectral-calibration}
is an admissible student matrix, its population
regret provides an upper bound. By
Theorem~\ref{thm:abstract-sharp}, the minimum
population regret over all measurable predictions
based on $H_p$ and the regret of the calibrated
spectral truncation are both asymptotic to
$\rho_0\tau_ps_p^2/2$. Equivalently, by
Theorem~\ref{thm:decomposition}, the additional
regret resulting from restriction to the affine
canonical class is $o(\tau_ps_p^2)$. The matching
upper and lower bounds therefore give
\begin{equation*}
 \mathcal E^P_{p,\tau_p}\sim
 \frac{\rho_0}{2}\tau_ps_p^2
\end{equation*}
and since $\tau_ps_p\to0$, we also have
\begin{equation*}
 \rho_0\frac{\tau_ps_p^2}
 {1+\sqrt{1+\tau_p^2s_p^2/v_P}}
 \sim\frac{\rho_0}{2}\tau_ps_p^2
\end{equation*}
which establishes the required asymptotic
equivalent in the regular regime, together
with the relative asymptotic optimality of
the calibrated spectral truncation.

To extend the result to arbitrary joint
asymptotic paths, note that every subsequence
of $(\eta_p)$ has a further subsequence
converging to some $\eta\in[0,\infty]$.
The ratio of the optimal population regret
to the claimed asymptotic equivalent approaches 1 along each such convergent subsequence,
by the regular-regime result established above
when $\eta=0$ and by
Theorem~\ref{thm:exact-crossover} when
$\eta\in(0,\infty]$. It follows that the ratio
converges to 1 along the original sequence.
The same argument establishes the relative
asymptotic optimality of the calibrated spectral
truncation along arbitrary joint paths.

Finally, the result for deterministic labels
follows directly from
Theorem~\ref{thm:exact-crossover} with
$\tau_p=\infty$ and does not require an
interchange of the zero-temperature limit
with the infimum over admissible student matrices.
\end{proof}

\subsection{The regular regime and a unified asymptotic law}
The sharp asymptotic equivalent established above
for $\eta_p\to\eta\in(0,\infty]$ does not extend
directly to the regular regime $\eta_p\to0$,
since in this regime the leading-order terms in
the minimum expected loss and the teacher Bayes
risk cancel upon subtraction. A separate analysis
is therefore required to determine the sharp
asymptotic equivalent of the population regret
in the regular regime.
\begin{theorem}[Exact regular-regime coefficient]\label{thm:regular}
Under the fixed-spectrum assumptions, if $\tau_p>0$ and $\tau_ps_p\to0$, then
\begin{equation}
\mathcal E_{p,\tau_p}\sim\frac{\tau_p^2s_p^2}{2}\,\E\,[\sig'(\tau_p F_\infty)]
\label{eq:regular}
\end{equation}
In particular, at fixed $\tau>0$ this yields the exact coefficient of $s_p^2$, when $\tau_p\to0$ it gives $\tau_p^2s_p^2/8$, and when $\tau_p\to\infty$ it yields $\rho_0\tau_ps_p^2/2$.
\end{theorem}
\begin{proof}
We first determine the leading-order population
regret of the unadjusted spectral truncation
$S=\tau_p\Lambda_p$, and subsequently establish
that optimising over the full admissible matrix
class cannot reduce this leading-order regret.
Let $D_p$ denote the population regret of this
particular student. Since the expression on
the right-hand side of \eqref{eq:temp-bregman}
was obtained by evaluating the Bregman remainder
at $S=\tau_p\Lambda_p$, we have
\begin{equation}\label{th610}
 \frac{D_p}{\tau_p^2s_p^2}
 =\E\,\left[G_p^2\int_0^1(1-t)
 \sig'(\tau_p H_p+(1-t)\eta_pG_p)\dd t\right]
\end{equation}
in which $G_p=R_p/s_p$ and $\eta_p=\tau_ps_p$.

Consider first a subsequence along which
$\tau_p\to t_0\in[0,\infty)$. For each fixed
$t\in[0,1]$, define
\begin{equation*}
 \Delta_p(t)=
 \sig'(\tau_pH_p+(1-t)\eta_pG_p)-\sig'(\tau_pH_p)
\end{equation*}
Since $\E \,G_p^2=1$ and $\eta_p\to0$, we have
$\eta_pG_p\xrightarrow{\Prob}0$. Uniform continuity
and boundedness of $\sig'$ therefore imply that
$\Delta_p(t)\xrightarrow{\Prob}0$ and that the
sequence is uniformly bounded. To establish
convergence after multiplication by $G_p^2$,
we use the estimate: for any $M>0$
\begin{equation*}
 \E\,[G_p^2|\Delta_p(t)|]
 \leq M^2\,\E\,|\Delta_p(t)|
 +C\,\E\,[G_p^2\one_{\{|G_p|>M\}}]
\end{equation*}
For fixed $M$, $M^2\,\E\,|\Delta_p(t)|\to 0$ in $p$ by boundedness and
convergence in probability of $\Delta_p(t)$,
while the second term converges uniformly to zero
as $M\to\infty$ by uniform integrability of
$G_p^2$. Consequently,
$\E\,[G_p^2|\Delta_p(t)|]\to0$.

Furthermore, since $G_p$ and $H_p$ are independent
and $\E \,G_p^2=1$, we have
\begin{equation*}
 \E\,[G_p^2\sig'(\tau_pH_p)]
 =\E\,\sig'(\tau_pH_p)
\end{equation*}
Since $H_p\to F_\infty$ a.s. and
$\tau_p\to t_0$, continuity and boundedness of
$\sig'$ imply that, as $p\to\infty$
\begin{equation*}
 \E\,\sig'(\tau_pH_p)\to\E\,\sig'(t_0F_\infty) 
\end{equation*}
By the preceding argument, for each fixed
$t\in[0,1]$, we have
\begin{equation*}
 \E\,\left[G_p^2\sig'(\tau_pH_p+(1-t)\eta_pG_p)\right]
 \to\E\,\sig'(t_0F_\infty)
\end{equation*}
Furthermore, since $\sig'\leq1/4$ and
$\E\,G_p^2=1$, we have
\begin{equation*}
 \E\,\left[G_p^2(1-t)\sig'(\tau_pH_p+(1-t)\eta_pG_p)\right]
 \leq\frac{1-t}{4}
\end{equation*}
The right-hand side is independent of $p$ and
integrable over $[0,1]$. Interchanging the
expectation and integration in \eqref{th610}
by Tonelli's theorem and subsequently applying
the dominated convergence theorem yields
\begin{align*}
 \lim_{p\to\infty}\frac{D_p}{\tau_p^2s_p^2}
 =\int_0^1(1-t)\E\,\sig'(t_0F_\infty)\dd t=\frac12\E\,\sig'(t_0F_\infty)
\end{align*}
Moreover, since $\tau_p\to t_0$, continuity
and boundedness of $\sig'$ imply that
\begin{equation*}
 \E\,\sig'(\tau_pF_\infty)
 \to\E\,\sig'(t_0F_\infty)>0
\end{equation*}
Consequently,
\begin{equation*}
 \frac{2D_p}{\tau_p^2s_p^2\,\E\,\sig'(\tau_pF_\infty)}
 \to1
\end{equation*}
or equivalently
\begin{equation*}
 D_p\sim\frac{\tau_p^2s_p^2}{2}\,
 \E\,\sig'(\tau_pF_\infty)
\end{equation*}
along every subsequence for which
$\tau_p\to t_0\in[0,\infty)$.

Consider now a subsequence along which
$\tau_p\to\infty$. In this case, the curvature
of the logarithmic loss becomes concentrated
near the decision boundary, and we exploit
the convergence of the retained signal densities
rather than convergence of the retained signal
itself. Conditioning on $G_p$ and introducing
the variable $v=\tau_ph+(1-t)\eta_pG_p$ in the
integral with respect to the retained signal
$h$ yields
\begin{equation}
 \frac{D_p}{\tau_ps_p^2}
 =\E\,\left[G_p^2\int_0^1(1-t)\int_\R
 \sig'(v)\rho_p(v/\tau_p-(1-t)s_pG_p)
 \dd v\dd t\right]
\end{equation}
For each fixed $v$ and $t$, the argument of
$\rho_p$  converges in probability to 0 
since $\tau_p\to\infty$ and
$\E\,[(s_pG_p)^2]=s_p^2\to0$. By the local uniform
convergence of the retained signal densities
and continuity of the limiting density at
zero, it follows that
\begin{equation*}
 \rho_p(v/\tau_p-(1-t)s_pG_p)\xrightarrow{\Prob}\rho_0
\end{equation*}
Since the densities are uniformly
bounded, the same truncation argument used
above, together with uniform integrability of
$G_p^2$, gives
\begin{equation*}
 \E\,[G_p^2\rho_p(v/\tau_p-(1-t)s_pG_p)]
 \to\rho_0
\end{equation*}
in which we have used $\E \,G_p^2=1$. Moreover,
these expectations are bounded above by
$K\E \,G_p^2=K$, in which $K$ is the uniform bound
on the retained signal densities. The dominated
convergence theorem therefore applies to the
integrals over $t\in[0,1]$ and $v\in\R$, with
integrable majorant $K(1-t)\sig'(v)$, yielding
\begin{equation*}
 \lim_{p\to\infty}\frac{D_p}{\tau_ps_p^2}
 =\rho_0\int_0^1(1-t)\dd t
 \int_\R\sig'(v)\dd v=\frac{\rho_0}{2}
\end{equation*}
Similarly, introducing the variable $v=\tau_ph$
in the expectation with respect to $F_\infty$
leads to
\begin{equation*}
 \tau_p\,\E\,\sig'(\tau_pF_\infty)
 =\int_\R\sig'(v)\rho_\infty(v/\tau_p)\dd v
 \to\rho_0
\end{equation*}
by continuity of $\rho_\infty$ at 0 and
dominated convergence. As a result, also along subsequences for which
$\tau_p\to\infty$, we obtain
\begin{equation*}
 D_p\sim\frac{\tau_p^2s_p^2}{2}
 \E\,\sig'(\tau_pF_\infty)
\end{equation*}
Since every subsequence of
$(\tau_p)$ has a further subsequence converging
in $[0,\infty]$, the equivalent holds along
arbitrary sequences satisfying $\eta_p\to0$.

We must still establish that the leading-order
regret of the unadjusted spectral truncation
cannot be reduced by independently optimising
the retained student eigenvalues. Let
\begin{equation*}
 q_0(h)=\sig(\tau_ph),\qquad
 \bar q(h)=\E\,\sig(\tau_p(h+R_p))
\end{equation*}
denote the probability predicted by the
unadjusted spectral student and the optimal
conditional probability prediction based on
$H_p$, respectively. Since the omitted residual is independent of the
retained Gaussian coordinates, the conditional
teacher probability given these coordinates
depends only on the retained signal $H_p$. Proposition~\ref{prop:alignment}
allows us to restrict the matrix optimisation
to students supported on the leading $p$
teacher eigenvectors, and, therefore, the minimum population
regret over all measurable predictions based
on these coordinates provides a
lower bound on $\mathcal E_{p,\tau_p}$, while
$D_p$ provides an upper bound.

By the conditional KL decomposition, the
difference between $D_p$ and the minimum
population regret over all measurable predictions
based on the retained coordinates is the
expected KL divergence between the Bernoulli
distributions with success probabilities
$\bar q(H_p)$ and $q_0(H_p)$, and thus
\begin{equation}
 0\leq D_p-\mathcal E_{p,\tau_p}
 \leq\E\,\KL(\operatorname{Bern}(\bar q(H_p))
 \|\operatorname{Bern}(q_0(H_p)))
\end{equation}
It therefore suffices to establish that the
decoder regret on the right-hand side is
asymptotically smaller than $D_p$.

To estimate the difference between $\bar q(h)$
and $q_0(h)$ uniformly in $h$, we first establish
a uniform exponential-moment bound for the
normalised unresolved component $G_p$. Setting
\begin{equation*}
 G_p=\sum_{j>p}a_j(X_j^2-1),
 \qquad a_j=\lambda_j/s_p
\end{equation*}
we have $\sum_{j>p}a_j^2=1/2$ and
$\max_{j>p}a_j\leq1/\sqrt{2}$. By independence
of the Gaussian coordinates and the moment
generating function of a chi-squared random
variable with one degree of freedom, we obtain
\begin{equation*}
 \E \,e^{tG_p}
 =\prod_{j>p}e^{-ta_j}(1-2ta_j)^{-1/2}
\end{equation*}
For a sufficiently small $|t|$, we have
$|2ta_j|\leq1/2$ uniformly in $j$ and $p$; 
taking logarithms and using the Taylor estimate
$-x-\log(1-x)\leq Cx^2$ for $|x|\leq1/2$
yields
\begin{equation*}
 \log\,\E\, e^{tG_p}
 =\frac12\sum_{j>p}[-2ta_j-\log(1-2ta_j)]
 \leq Ct^2\sum_{j>p}a_j^2\leq Ct^2
\end{equation*}
Exponentiating the bound on the logarithm of the
moment-generating function yields
$\E \,e^{tG_p}\leq e^{Ct^2}$ for sufficiently
small $|t|$, uniformly in $p$. Since
$e^{t|x|}\leq e^{tx}+e^{-tx}$ for $t>0$,
it follows that
\begin{equation*}
 \E \,e^{t|G_p|}
 \leq \E \,e^{tG_p}+\E \,e^{-tG_p}
 \leq 2e^{Ct^2}
\end{equation*}
Choosing a sufficiently small fixed $\delta>0$, 
therefore, furnishes the uniform exponential-moment bound $\sup_p\E \,e^{\delta|G_p|}<\infty$. Moreover, since
$\eta_p\to0$, we have $\eta_p\leq\delta/2$
for all sufficiently large $p$, and hence
\begin{equation*}
 G_p^2e^{\eta_p|G_p|}
 \leq C_\delta e^{\delta|G_p|}
\end{equation*}
and it follows that $\E\,[G_p^2e^{\eta_p|G_p|}]$
is uniformly bounded for all sufficiently
large $p$.

We now apply Taylor's formula with integral
remainder to the sigmoid function, using
$R_p=s_pG_p$, to obtain
\begin{align*}
 \bar q(h)-q_0(h)
 =\eta_p\E\, G_p\,\sig'(\tau_ph)+\eta_p^2\E\,\left[G_p^2\int_0^1(1-t)
 \sig''(\tau_ph+t\eta_pG_p)\dd t\right]
\end{align*}
in which the first term vanishes since $\E \,G_p=0$.
To estimate the remainder, note that the
identity $\sig''=\sig'(1-2\sig)$ implies that 
\begin{equation*}
 |\sig''|\leq\sig'\qquad \left|\frac{\dd}{\dd x}\log\sig'(x)\right|
 \leq1
\end{equation*}
and integrating the latter inequality between
$x$ and $x+y$ yields
\begin{equation*}
 \sig'(x+y)\leq e^{|y|}\sig'(x)
\end{equation*}
Consequently, the second-order Taylor remainder
satisfies
\begin{align*}
 |\bar q(h)-q_0(h)|
 &\leq\eta_p^2\sig'(\tau_ph)\,
 \E\,\left[G_p^2\int_0^1(1-t)
 e^{t\eta_p|G_p|}\dd t\right]\\
 &\leq\frac{\eta_p^2}{2}\sig'(\tau_ph)\,
 \E\,[G_p^2e^{\eta_p|G_p|}]
\end{align*}
The uniform exponential-moment estimate
established above therefore gives
\begin{equation*}
 |\bar q(h)-q_0(h)|
 \leq C\eta_p^2\sig'(\tau_ph)
\end{equation*}
uniformly in $h$.

For any $a\in[0,1]$ and $b\in(0,1)$, applying
the inequality $\log x\leq x-1$ to the two
terms in the Bernoulli KL divergence leads to
\begin{equation*}
 \KL(\operatorname{Bern}(a)\|\operatorname{Bern}(b))
 \leq\frac{(a-b)^2}{b(1-b)}
\end{equation*}
with the cases $a=0$ and $a=1$ following by
continuity. Taking the values $a=\bar q(h)$ and $b=q_0(h)$
and using the identity
$q_0(h)(1-q_0(h))=\sig'(\tau_ph)$ yields
\begin{equation*}
 \KL(\operatorname{Bern}(\bar q(h))
 \|\operatorname{Bern}(q_0(h)))
 \leq C\eta_p^4\sig'(\tau_ph)
\end{equation*}
Now taking the expectation with respect to $H_p$
and using both $\sig'\leq1/4$ and the uniform
density bound results in
\begin{equation}
 0\leq D_p-\mathcal E_{p,\tau_p}
 \leq C\eta_p^4\E\,\sig'(\tau_pH_p)
 \leq C\eta_p^4\min\{1,K/\tau_p\}
\end{equation}
in which the second inequality follows from  $v=\tau_ph$ and
$\int_\R\sig'(v)\dd v=1$.

Finally, along any subsequence on which
$\tau_p\to t_0\in[0,\infty)$, we have
\begin{equation*}
 D_p\sim\frac{\eta_p^2}{2}\E\,\sig'(t_0F_\infty)
\end{equation*}
while the decoder regret is $O(\eta_p^4)$.
Along any subsequence on which
$\tau_p\to\infty$, we have $D_p\sim\frac{\rho_0\eta_p^2}{2\tau_p}$, 
while the decoder regret is
$O(\eta_p^4/\tau_p)$. In both cases the ratio
of the decoder regret to $D_p$ is
$O(\eta_p^2)$ and hence approaches 0.
By the same subsequence argument used above,
it follows that
\begin{equation*}
 \frac{D_p-\mathcal E_{p,\tau_p}}{D_p}\to0
\end{equation*}
along arbitrary sequences satisfying
$\eta_p\to0$. Thus the optimal rank-constrained
population regret and the population regret
of the unadjusted spectral truncation have
the same leading-order behaviour, which
confirms \eqref{eq:regular}.
\end{proof}

\begin{corollary}[Unified low-temperature asymptotics]\label{cor:unified}
Under the fixed-spectrum assumptions, along every sequence $\tau_p\to\infty$,
\begin{equation}
\mathcal E_{p,\tau_p}\sim\rho_0\frac{\tau_ps_p^2}{1+\sqrt{1+3\tau_p^2s_p^2/\pi^2}}
\label{eq:unified}
\end{equation}
The student~\eqref{eq:calibration} is asymptotically optimal in relative risk. For deterministic labels $\mathcal E_p(\Lambda_d)\sim\pi\rho_0s_p/\sqrt3$.
\end{corollary}
\begin{proof}
Apply Corollary~\ref{cor:canonical-global} to $P=\sig$, for which $v_P=\pi^2/3$ and the calibration is \eqref{eq:calibration}.
\end{proof}

\begin{remark}
For the power-law spectrum considered in
Corollary~\ref{cor:powerlaw}, let
$\beta=\gamma-1/2$ and
$A_\gamma=C_0\sqrt{2/(2\gamma-1)}$, so that
$s_p\sim A_\gamma p^{-\beta}$ as $p\to\infty$
with $d/p\to\infty$. At a fixed positive inverse
temperature, the optimal population regret decays
as $p^{-2\beta}$, while for deterministic labels
we have the sharp asymptotic equivalent
\begin{equation*}
 \mathcal E_{p,\infty}
 \sim\frac{\pi\rho_0A_\gamma}{\sqrt{3}}p^{-\beta}
\end{equation*}
More generally, for an inverse temperature
that varies with the student width according
to $\tau_p\asymp p^\kappa$, with $\kappa\geq0$,
Theorem~\ref{thm:uniform-temp} yields
\begin{equation*}
 \mathcal E_{p,\tau_p}\asymp
 \begin{cases}
  p^{-(2\beta-\kappa)} & 0\leq\kappa\leq\beta\\
  p^{-\beta} & \kappa\geq\beta
 \end{cases}
\end{equation*}
The crossover between the regular and singular
regimes occurs when $\tau s_p\asymp1$, corresponding
to a student width of order
$(\tau A_\gamma)^{1/\beta}$ for sufficiently large
$\tau$ within the range of validity of the
spectral-tail asymptotics.

These results characterise population approximation
laws with respect to student width rather than
scaling laws with respect to training sample size
or total parameter count. In particular, the two
limiting width exponents are obtained within the
same family of observation models and under the
same loss, solely by varying the sharpness of the
teacher response. In contrast, the squared-error
comparison in Section~\ref{sec:results} involves
a change in the loss function.
\end{remark}

\section{The nonconvex optimisation landscape}\label{sec:landscape}
The results above concern the optimum over positive-semidefinite matrices of prescribed rank. We now examine whether the corresponding factorised objective has stationary points that cannot attain that optimum. Throughout this section the dimension and teacher are fixed, $1\le p\le d$, and $0<\tau<\infty$. The teacher link $G$ is continuous and strictly increasing, with values in $(0,1)$, and the student canonical inverse link $P$ satisfies the assumptions of Section~\ref{sec:canonical}. We additionally assume $P\in C^1(\R)$ and $\|P'\|_\infty<\infty$. The matched case is of particular interest, but the argument does not require matching links.

\subsection{Spectral Bregman approximation and factorisation}
\begin{definition}[Spectral potential and population objective]
Define the teacher moment matrix by
\begin{equation}
 T=T_{\tau,G}=\E\,[G(\tau F)(XX^\top-I_d)]
\end{equation}
The spectral potential associated with the canonical
loss and the corresponding population objective
are defined, respectively, by
\begin{equation}
 \mathcal V_P(S)=\E\, A(f_S(X)),\qquad
 \mathcal F_{P,G}(S)=\mathcal V_P(S)-\tr(ST)
\end{equation}
\end{definition}
If $G=P$ and $S_0=\tau\Lambda$, then $T=\nabla\mathcal V_P(S_0)$ and the excess population loss is precisely the matrix Bregman divergence
\begin{equation}
 \E\, D_P(P(f_{S_0}),f_S)=\mathcal V_P(S)-\mathcal V_P(S_0)-\langle\nabla\mathcal V_P(S_0),S-S_0\rangle_{\mathrm F}
\end{equation}
Here $\langle A,B\rangle_{\mathrm F}=\tr(A^\top B)$. This places the rank-constrained problem within spectral matrix nearness \citep{dhillon2007,kulis2009}, with a nonseparable spectral potential determined by the Gaussian population loss. The factorised formulation is
\begin{equation}
 \mathcal J(W)=\mathcal F_{P,G}(WW^\top),\qquad W\in\R^{d\times p}
 \label{eq:factor-objective}
\end{equation}
\begin{lemma}[Existence, smoothness and the moment spectrum]\label{lem:landscape-prelim}
The matrix objective is strictly convex and coercive on symmetric matrices, and the factorised objective is coercive and twice continuously differentiable. The moment matrix $T$ has the same eigenspaces as the teacher, with eigenvalues $t_i$ satisfying $t_i>t_j$ if $\lambda_i>\lambda_j$, and $t_i>0$ if and only if $\lambda_i>0$.
\end{lemma}
\begin{proof}
We first establish the differentiability and strict
convexity of the population objective on the space
of symmetric matrices. For any symmetric matrix
$D$, we have
\begin{equation*}
 f_D(X)=\langle D,XX^\top-I_d\rangle_{\mathrm F}
\end{equation*}
and hence, by the Cauchy-Schwarz inequality
\begin{equation*}
 |f_D(X)|\leq\Frob{D}\Frob{XX^\top-I_d}
\end{equation*}
Since $A'=P$ and $0<P<1$, the first directional
derivative of $A(f_S(X))$ in the direction $D$
satisfies
\begin{equation*}
 |P(f_S(X))f_D(X)|
 \leq\Frob{D}\Frob{XX^\top-I_d}
\end{equation*}
and the mixed second directional derivative in the
directions $D$ and $E$ satisfies
\begin{equation*}
 |P'(f_S(X))f_D(X)f_E(X)|
 \leq\|P'\|_\infty\Frob{D}\Frob{E}\Frob{XX^\top-I_d}^2
\end{equation*}
Both upper bounds are integrable by finiteness
of the fourth moments of the Gaussian coordinates, 
and are independent of $S$. Thus, by
dominated convergence, we can
differentiate under the expectation and the
resulting derivatives are continuous in $S$.
Using the linearity of the term $\tr(ST)$ in
the population objective, we obtain
\begin{align}
 \nabla\mathcal F_{P,G}(S)
 &=\E\,[P(f_S)(XX^\top-I_d)]-T
 \label{eq:matrix-gradient}\\
 D^2\mathcal V_P(S)[D,D]
 &=\E\,[P'(f_S)f_D^2]
 \label{eq:matrix-hessian}
\end{align}
To establish strict convexity, consider two
distinct symmetric matrices $S_1$ and $S_2$.
By the Gaussian moment identity \eqref{eq:gaussian-quadratic-variance}, we obtain
\begin{equation*}
 \E\,[(f_{S_1}-f_{S_2})^2]
 =2\Frob{S_1-S_2}^2>0
\end{equation*}
and so $f_{S_1}$ and $f_{S_2}$ differ on a set
of positive probability. Strict convexity of
$A$, which follows from strict monotonicity
of $P$, therefore implies strict convexity
of $\mathcal V_P$. Subtraction of the linear
term $\tr(ST)$ preserves strict convexity
and hence $\mathcal F_{P,G}$ is strictly convex.

We next establish coercivity of the population
objective. Let $q=G(\tau F)$ and define
$w=\min\{q,1-q\}$. Since the inverse temperature
is finite and $G$ takes values in $(0,1)$, we
have $w>0$ a.s.. Using $A(u)\geq u_+$,
for any $u\in\R$, we obtain
\begin{equation*}
 A(u)-qu\geq u_+-qu
 =\begin{cases}
  (1-q)|u| & u\geq0\\
  q|u| & u<0
 \end{cases}
 \geq w|u|
\end{equation*}
and applying this inequality with $u=f_S(X)$
and taking expectations yields
\begin{equation}
 \mathcal F_{P,G}(S)\geq\E\,[w|f_S|]
\end{equation}

Consider now the function
$\mathcal N(S)=\E\,[w|f_S|]$, which is positively
homogeneous in $S$. To establish continuity,
we use the reverse triangle inequality and
the linearity of $f_S$ in $S$ to obtain
\begin{align*}
 |\mathcal N(S_1)-\mathcal N(S_2)|
 \leq\E\,[w|f_{S_1-S_2}|]
 \leq\E\,|f_{S_1-S_2}|\leq\sqrt{\E\, f_{S_1-S_2}^2}
 =\sqrt{2}\Frob{S_1-S_2}
\end{align*}
in which the final equality follows from
\eqref{eq:gaussian-quadratic-variance}. Furthermore, for
any nonzero symmetric matrix $S$, we have
$\E\, f_S^2=2\Frob{S}^2>0$, so that $|f_S|>0$
on a set of positive probability. Since
$w>0$ a.s., it follows that
$\mathcal N(S)>0$ whenever $S\ne0$.

By continuity and compactness of the unit
sphere in the space of symmetric matrices,
the quantity $c=\min_{\Frob{S}=1}\mathcal N(S)$ is strictly positive. Positive homogeneity
of $\mathcal N$ therefore yields
\begin{equation*}
 \mathcal F_{P,G}(S)\geq\mathcal N(S)
 \geq c\Frob{S}
\end{equation*}
which establishes coercivity of the population
objective, and in which the constant $c$ may depend on the
ambient dimension, the teacher and the inverse
temperature. Since $WW^\top$ has rank at most
$p$, the Cauchy-Schwarz inequality applied
to its nonzero eigenvalues gives
\begin{equation*}
 \|WW^\top\|_{\mathrm F}
 \geq\frac{\tr(WW^\top)}{\sqrt p}
 =\frac{\|W\|_{\mathrm F}^2}{\sqrt p}
\end{equation*}
so that
\begin{equation*}
 \mathcal F_{P,G}(WW^\top)
 \geq\frac{c}{\sqrt p}\|W\|_{\mathrm F}^2
\end{equation*}
which establishes coercivity of the factorised
population objective.

Finally, we examine the spectral properties
of the teacher moment matrix $T$. Strict
ordering of its eigenvalues when the
corresponding teacher eigenvalues are distinct
follows from the exchangeability argument in
Proposition~\ref{prop:alignment} and strict
monotonicity of $G$. To establish positivity
of the moment eigenvalues, fix $i$ and write
$F=\lambda_i U+C$, in which $U=X_i^2-1$ and
$C=\sum_{j\ne i}\lambda_j(X_j^2-1)$ is independent
of $U$. The corresponding eigenvalue of $T$
is given by
\begin{equation*}
 t_i=\E\,[U G(\tau(\lambda_i U+C))]
\end{equation*}
Let $U'$ be an independent copy of $U$,
independent also of $C$. Then, since $\E\,U'=0$,
we have
\begin{equation*}
 \E\,[U'G(\tau(\lambda_i U+C))]=0
\end{equation*}
Using this identity and exchangeability of
$U$ and $U'$ we have
\begin{equation}
 t_i=\frac12\E\,[(U-U')
 \{G(\tau(\lambda_i U+C))-G(\tau(\lambda_i U'+C))\}]
\end{equation}
If $\lambda_i>0$, strict monotonicity of $G$
and positivity of $\tau$ imply that the
integrand is strictly positive if
$U\ne U'$. Since $U$ has a continuous
distribution, it follows that $t_i>0$.
If $\lambda_i=0$, the difference between
the two values of $G$ vanishes identically, 
and hence $t_i=0$. Moreover, equal teacher
eigenvalues yield equal moment eigenvalues
by exchangeability. Together with the strict
ordering for distinct teacher eigenvalues,
this shows  that $T$ and $\Lambda$
have identical eigenspace decompositions.
\end{proof}

\subsection{Stationary matrices and optimal rank}
\begin{theorem}[Strict-saddle landscape]\label{thm:landscape}
Every stationary factor of \eqref{eq:factor-objective} satisfies the commutation relation
\begin{equation*}
 [WW^\top,T]=0
\end{equation*}
in which  $[A,B]=AB-BA$ denotes the matrix commutator. Every local minimum is global, and every nonglobal stationary factor has a strictly negative Hessian eigenvalue. Every global minimum has rank
\begin{equation*}
 r_*=\min(p,\rank\Lambda)
\end{equation*}
and its range is a leading spectral subspace of dimension $r_*$. The coefficients on such a subspace are uniquely determined by the reduced strictly convex objective.
\end{theorem}
\begin{proof}
Since the factorised objective $\mathcal J$ is
continuous and coercive by
Lemma~\ref{lem:landscape-prelim}, it attains its
global minimum. We first derive the conditions
that characterise its stationary points.
Writing $S=WW^\top$, the first variation of $S$
associated with a perturbation $E\in\R^{d\times p}$
of the factor $W$ is $WE^\top+EW^\top$.
The chain rule, symmetry of the matrix gradient
and cyclic invariance of the trace lead to
\begin{align*}
 D\mathcal J(W)[E]
 =\langle\nabla\mathcal F_{P,G}(S),
 WE^\top+EW^\top\rangle_{\mathrm F}=2\langle\nabla\mathcal F_{P,G}(S)W,E\rangle_{\mathrm F}
\end{align*}
Since this identity holds for every perturbation
$E$, the gradient of the factorised objective is
\begin{equation*}
 \nabla\mathcal J(W)=2\nabla\mathcal F_{P,G}(S)W
\end{equation*}
At a stationary point we therefore have
$\nabla\mathcal F_{P,G}(S)W=0$. Multiplying on
the right by $W^\top$ and taking
the transpose of the resulting identity gives
\begin{equation}
 \nabla\mathcal F_{P,G}(S)S
 =S\nabla\mathcal F_{P,G}(S)=0
\end{equation}

We next establish that every stationary matrix
$S$ commutes with the teacher moment matrix $T$.
The commutation property of spectral gradients
\citep{lewis1996} can be derived directly in the
present case. By orthogonal invariance of
$\mathcal V_P$, for any skew-symmetric matrix
$K$ and any $t\in\R$, we have
\begin{equation*}
 \mathcal V_P(e^{tK}Se^{-tK})=\mathcal V_P(S)
\end{equation*}
since $e^{tK}$ is orthogonal. Differentiating
with respect to $t$ at $t=0$ yields
\begin{equation*}
 0=\langle\nabla\mathcal V_P(S),KS-SK\rangle_{\mathrm F}
 =\tr([S,\nabla\mathcal V_P(S)]K)
\end{equation*}
Since the commutator of two symmetric matrices
is skew-symmetric, we may choose
$K=[S,\nabla\mathcal V_P(S)]^\top$, resulting in
\begin{equation*}
 0=\tr([S,\nabla\mathcal V_P(S)]
 [S,\nabla\mathcal V_P(S)]^\top)
 =\Frob{[S,\nabla\mathcal V_P(S)]}^2
\end{equation*}
It therefore follows that $[S,\nabla\mathcal V_P(S)]=0$.
Furthermore, the stationarity condition derived
above implies that $[S,\nabla\mathcal F_{P,G}(S)]=0$.
Since
$\nabla\mathcal F_{P,G}(S)=\nabla\mathcal V_P(S)-T$,
subtracting the two commutation relations yields $ [S,T]=0$. 
Thus, every stationary student matrix can be
simultaneously diagonalised with the teacher
moment matrix and hence, by
Lemma~\ref{lem:landscape-prelim}, with the teacher
matrix $\Lambda$.

In a common orthonormal eigenbasis of $S$, $T$
and $\Lambda$, let
$S=\diag(\mu_1,\ldots,\mu_d)$ and define the
active spectral index set $I=\{j:\mu_j>0\}$,
with $r=|I|=\rank S$. In this basis, the canonical
coordinate of the student is
\begin{equation*}
 f_S(X)=\sum_{j\in I}\mu_j(X_j^2-1)
\end{equation*}
If $i\notin I$, the Gaussian coordinate $X_i$
does not enter into the expression for $f_S(X)$, so $P(f_S)$ is independent of the
centred random variable $X_i^2-1$ and
\eqref{eq:matrix-gradient} implies
\begin{equation*}
 (\nabla\mathcal V_P(S))_{ii}
 =\E\,[P(f_S)(X_i^2-1)]=0
\end{equation*}
If $j\in I$, the covariance argument in
Lemma~\ref{lem:landscape-prelim}, applied with
$P$ in place of $G$, demands
$(\nabla\mathcal V_P(S))_{jj}>0$.
Since $\mu_j>0$, the stationarity condition
$\nabla\mathcal F_{P,G}(S)S=0$ implies that
\begin{equation*}
 (\nabla\mathcal F_{P,G}(S))_{jj}=0
\end{equation*}
and hence $(\nabla\mathcal V_P(S))_{jj}=t_j>0$.
Moreover, reflection symmetry of the Gaussian
distribution implies that
$\nabla\mathcal V_P(S)$ is diagonal in the
chosen eigenbasis, by the same argument used
for the teacher moment matrix in
Proposition~\ref{prop:alignment}. Combining
these observations, we obtain the following
expression for the matrix gradient at a
stationary factor
\begin{equation}
 \nabla\mathcal F_{P,G}(S)
 =-\sum_{i\notin I}t_i e_ie_i^\top
 \label{eq:stationary-gradient}
\end{equation}
In particular, every active student direction
must correspond to a strictly positive
teacher eigenvalue.

We now construct two types of factor perturbation
that yield negative Hessian curvature at
nonglobal stationary points. The first
corresponds to replacing an active spectral
direction by an inactive direction with a
larger teacher moment eigenvalue. Suppose
that $j\in I$, $i\notin I$ and $t_i>t_j$.
Let $R_\theta$ denote an orthogonal rotation
through an angle $\theta$ in the plane spanned
by $e_i$ and $e_j$, acting as the identity
on the orthogonal complement of this plane.
The perturbed factor is $R_\theta W$ and its
corresponding student matrix is
$R_\theta SR_\theta^\top$. By orthogonal
invariance of the spectral potential,
$\mathcal V_P(R_\theta SR_\theta^\top)
=\mathcal V_P(S)$.
The change in the objective therefore arises
entirely from the linear trace term. More
precisely, the contribution $\mu_jt_j$ to
$\tr(ST)$ is replaced by
$\mu_j(t_j\cos^2\theta+t_i\sin^2\theta)$,
while all other contributions remain unchanged.
It follows that
\begin{equation}
 \mathcal J(R_\theta W)-\mathcal J(W)
 =-\mu_j(t_i-t_j)\sin^2\theta
\end{equation}

Since $i\notin I$, the $i$th row of $W$ is
zero, while the squared Euclidean norm of
its $j$th row is $\mu_j$. Consequently, the
tangent to the rotation path at $\theta=0$
satisfies $ \Frob{R'_0W}^2=\mu_j$.  Since $W$ is a stationary point of $\mathcal J$,
the second derivative of the objective along
the rotation path is given entirely by the
Hessian acting on its tangent, i.e.
\begin{equation*}
 \left.\frac{\dd^2}{\dd\theta^2}
 \mathcal J(R_\theta W)\right|_{\theta=0}
 =D^2\mathcal J(W)[R'_0W,R'_0W]
\end{equation*}
The additional term involving the second
derivative of the path vanishes since
$\nabla\mathcal J(W)=0$. Differentiating the
objective difference twice with respect to
$\theta$ at $\theta=0$ therefore yields the
Hessian Rayleigh quotient
\begin{equation*}
 \frac{D^2\mathcal J(W)[R'_0W,R'_0W]}
 {\Frob{R'_0W}^2}
 =-2(t_i-t_j)<0
\end{equation*}
Thus, a stationary factor whose active spectral
directions exclude a direction with a larger
teacher moment eigenvalue has negative Hessian
curvature.

The second type of perturbation corresponds
to activating an additional spectral direction
when the rank of the stationary matrix is
strictly smaller than the admissible rank.
Suppose that $r<p$ and that there exists an
index $i\notin I$ with $t_i>0$. Since
$\rank W=r<p$, there exists a unit vector
$a\in\R^p$ in  $\mathrm{ker}\,W$.
Consider the perturbed factor
$W+\theta e_i a^\top$. Using $Wa=0$ and
$a^\top a=1$, the corresponding student
matrix is
\begin{align*}
 (W+\theta e_i a^\top)(W+\theta e_i a^\top)^\top=S+\theta Wa e_i^\top
 +\theta e_i a^\top W^\top+\theta^2e_ie_i^\top=S+\theta^2e_ie_i^\top
\end{align*}
A first-order Taylor expansion of the matrix
objective at $S$ therefore gives
\begin{equation*}
 \mathcal J(W+\theta e_i a^\top)
 =\mathcal J(W)+\theta^2
 \langle\nabla\mathcal F_{P,G}(S),e_ie_i^\top\rangle_{\mathrm F}
 +o(\theta^2)
\end{equation*}
and using \eqref{eq:stationary-gradient}, we obtain
\begin{equation*}
 \mathcal J(W+\theta e_i a^\top)
 =\mathcal J(W)-t_i\theta^2+o(\theta^2)
\end{equation*}
Since $\Frob{e_i a^\top}=1$, the Hessian
Rayleigh quotient in this direction is
\begin{equation*}
 D^2\mathcal J(W)[e_i a^\top,e_i a^\top]
 =-2t_i<0
\end{equation*}
Hence, a stationary factor with an unused
factor direction and an inactive spectral
direction corresponding to a positive
teacher eigenvalue also has negative
Hessian curvature.

We now establish that every stationary
factor not covered by either of the two
negative-curvature constructions is a global
minimiser. Recall that
$r_*=\min\{p,\rank\Lambda\}$. Since every
active student direction corresponds to a
strictly positive teacher eigenvalue, we
necessarily have $r\leq r_*$. If $r<r_*$,
then $r<p$ and there exists at least one
inactive direction corresponding to a
strictly positive teacher eigenvalue.
By Lemma~\ref{lem:landscape-prelim}, its
moment eigenvalue is also strictly positive,
so that the second perturbation establishes
negative Hessian curvature.

If $r=r_*$ but the active student subspace
is not a leading spectral subspace of the
teacher, there must exist an active index
$j\in I$ and an inactive index $i\notin I$
such that $t_i>t_j$. The first perturbation
then establishes negative Hessian curvature
without changing the student rank.

The remaining stationary factors therefore
have rank $r_*$ and their active subspaces
are leading spectral subspaces of the
teacher, allowing arbitrary choices within
an eigenspace associated with a repeated
teacher eigenvalue at the rank boundary.
On any such subspace, the stationarity
condition implies that the gradient of the
reduced objective with respect to the
retained student eigenvalues vanishes.
Since the reduced objective is strictly
convex, these eigenvalues constitute its
unique minimiser.

If $r_*<p$, then all remaining teacher
eigenvalues and their corresponding moment
eigenvalues are 0. By
\eqref{eq:stationary-gradient}, the gradient
in each of these additional spectral
directions vanishes when the corresponding
student eigenvalue is 0. Convexity of
the reduced objective therefore ensures
that allowing nonzero student eigenvalues
in these directions cannot decrease its
minimum value. Proposition~\ref{prop:alignment}
then implies global optimality over the
full admissible matrix class.

Since every local minimum of the twice
continuously differentiable factorised
objective is stationary and has a positive
semidefinite Hessian, the two negative-curvature
constructions exclude all nonglobal stationary
points as local minima. Hence, every
local minimum is global and every nonglobal
stationary point is a strict saddle, which completes
the proof.
\end{proof}

\begin{corollary}[Stationary matrix classification]\label{cor:stationary-count}
If the $k=\rank\Lambda$ positive teacher
eigenvalues are simple, there are exactly
$\sum_{r=0}^{\min(p,k)}\binom{k}{r}$ distinct
matrices $WW^\top$ arising from stationary
factors, of which precisely one is globally
optimal. More generally, each stationary matrix
is obtained by selecting subspaces of the positive
teacher eigenspaces with a combined dimension
not exceeding $p$ and minimising the reduced
objective over student matrices supported on
the direct sum of the selected subspaces.
Any two factors representing the same stationary
matrix are related by right multiplication by
an orthogonal matrix.
\end{corollary}
\begin{proof}
Choose a common orthonormal eigenbasis of the
teacher matrix and the teacher moment matrix
and let $I$ be a subset of indices corresponding
to positive moment eigenvalues, with $|I|\leq p$.
The reduced objective governing the student
eigenvalues associated with the selected
spectral directions is
\begin{equation*}
 \varphi_I(\mu)=\E\, A\left(\sum_{j\in I}
 \mu_j(X_j^2-1)\right)-\sum_{j\in I}t_j\mu_j
\end{equation*}
By restriction of the matrix objective to
diagonal matrices supported on the selected
spectral subspace, $\varphi_I$ inherits the
coercivity and strict convexity established in
Lemma~\ref{lem:landscape-prelim}. As a consequence,
it attains a unique minimum on the closed
nonnegative orthant $[0,\infty)^{|I|}$.
We claim that the minimiser $\mu^*$ lies in
the interior of this orthant. Indeed, if
$\mu_i^*=0$ for some $i\in I$, the argument
is independent of the Gaussian coordinate
$X_i$ in the expression for the canonical
coordinate of the student. Independence of
the Gaussian coordinates and the identity
$\E\,(X_i^2-1)=0$ therefore yield
\begin{equation*}
 \frac{\partial\varphi_I}{\partial\mu_i}(\mu^*)
 =\E\,\left[P\left(\sum_{j\in I}\mu_j^*
 (X_j^2-1)\right)(X_i^2-1)\right]-t_i=-t_i<0
\end{equation*}
Thus, increasing $\mu_i$ from zero would
strictly decrease the objective while
preserving nonnegativity of the coefficients,
contradicting the minimality of $\mu^*$.
It thus follows that $\mu_j^*>0$ for every $j\in I$
and hence
\begin{equation*}
 \frac{\partial\varphi_I}{\partial\mu_j}(\mu^*)=0
 \qquad j\in I
\end{equation*}
The corresponding student matrix is
\begin{equation*}
 S_I=\sum_{j\in I}\mu_j^*e_je_j^\top
\end{equation*}
Reflection symmetry and the vanishing of
the partial derivatives in the active
spectral directions imply that
$\nabla\mathcal F_{P,G}(S_I)$ has the form
given in \eqref{eq:stationary-gradient}.
Since the range of any factor $W$ satisfying
$WW^\top=S_I$ coincides with the range of
$S_I$, we have
$\nabla\mathcal F_{P,G}(S_I)W=0$, and hence
$\nabla\mathcal J(W)=0$. Thus, every factor
of $S_I$ is stationary.

Conversely, in the proof of
Theorem~\ref{thm:landscape} we established
that every stationary matrix admits a common
orthonormal eigenbasis with the teacher matrix
and that every active student direction
corresponds to a positive teacher moment
eigenvalue. Moreover, the stationarity
conditions imply that the partial derivatives
of the reduced objective associated with
the active spectral directions vanish.
By strict convexity of the reduced objective,
the active student eigenvalues must therefore
coincide with its unique minimiser. This
characterises every stationary matrix,
including the zero matrix obtained by
taking $I=\varnothing$. If a teacher
eigenvalue is repeated, the corresponding
moment eigenvalues are equal. The reduced
objective is invariant under permutations
of the active student coefficients associated
with that teacher eigenvalue, by exchangeability
of the Gaussian coordinates and equality of
the corresponding moment eigenvalues.
Uniqueness of the reduced minimiser therefore
implies equality of these active coefficients.
Since the common orthonormal eigenbasis can
be chosen arbitrarily within each repeated
teacher eigenspace, the active spectral
subspace within each such eigenspace can
also be chosen arbitrarily.

When the $k$ positive teacher eigenvalues
are simple, each subset of $r$ positive
teacher eigendirections uniquely determines
a stationary matrix of rank $r$. Distinct
subsets give distinct stationary matrices
because their ranges differ. Since there
are $\binom{k}{r}$ such subsets of cardinality
$r$ and the admissible rank satisfies
$0\leq r\leq\min(p,k)$, the total number
of distinct stationary matrices is
\begin{equation*}
 \sum_{r=0}^{\min(p,k)}\binom{k}{r}
\end{equation*}
By the global optimality characterisation
in the preceding theorem, precisely one
of these matrices is globally optimal,
namely the stationary matrix whose range
is the leading teacher spectral subspace
of dimension $\min(p,k)$.

Finally, suppose that two factors
$W_1,W_2\in\R^{d\times p}$ represent the
same stationary matrix $S$, so that
$W_1W_1^\top=W_2W_2^\top=S$.
For $r=\rank S>0$, write the thin spectral
decomposition of $S$ in the form
$S=U_r C^2U_r^\top$, in which $U_r$ has
orthonormal columns and $C$ is a positive
diagonal matrix. Since both factors have
the same range as $S$, they can be written as
\begin{equation*}
 W_\ell=U_r C V_\ell^\top,\qquad
 V_\ell^\top V_\ell=I_r,\qquad \ell=1,2
\end{equation*}
in which $V_\ell=W_\ell^\top U_r C^{-1}$
has orthonormal columns in $\R^p$.
Extending the columns of $V_1$ and $V_2$
to orthonormal bases of $\R^p$, we obtain
an orthogonal matrix $Q\in\R^{p\times p}$
such that $QV_2=V_1$, and thus
\begin{equation*}
 W_1Q=U_r C V_1^\top Q
 =U_r C V_2^\top=W_2
\end{equation*}
Therefore, any two factors representing the same
stationary matrix are related by right
multiplication by an orthogonal matrix,
including when $r<p$. For $r=0$, both
factors are necessarily zero and the
assertion holds trivially.
\end{proof}
\begin{remark}
The globally optimal student matrix is unique
if either $p<k$ and there is a strict spectral
gap $\lambda_p>\lambda_{p+1}$ at the retained
rank or $p\geq k$, in which $k=\rank\Lambda$.
In the matched and fully representable case,
the unique optimal matrix is $S=\tau\Lambda$
by strict propriety of the loss and strict
convexity of the population objective.
If $p<k$ and $\lambda_p=\lambda_{p+1}$,
additional nonuniqueness arises from the
freedom to select the retained subspace
within the teacher eigenspace associated
with the repeated eigenvalue. This is
distinct from the nonuniqueness of the
factorisation $S=WW^\top$ under right
orthogonal transformations.
\end{remark}

\subsection{Exact negative Hessian index}
The strict-saddle result can be refined by explicitly decomposing the Hessian. In particular the number of unstable directions is independent of the nonlinear coefficient minimisation once the active spectral subspace is known.
\begin{theorem}[Negative Hessian index]\label{thm:hessian-index}
Let $W$ be a stationary point of the factorised
objective $\mathcal J$ and write $S=WW^\top$.
By Theorem~\ref{thm:landscape}, the matrices
$S$ and $T$ admit a common orthonormal eigenbasis,
with corresponding eigenvalues $\mu_i$ and $t_i$,
respectively. Define the active spectral index
set and its cardinality by
\begin{equation*}
 I=\{i:\mu_i>0\},\qquad r=|I|=\rank S
\end{equation*}
The negative eigenvalues of the Hessian of
$\mathcal J$ at $W$, counted with multiplicity,
consist precisely of $2(t_j-t_i)$ for each pair
$j\in I$, $i\notin I$ satisfying $t_i>t_j$,
together with $-2t_i$ repeated $p-r$ times
for each inactive index $i$ satisfying $t_i>0$.
Therefore, the number of negative Hessian
eigenvalues is
\begin{equation}
 n_-(W)=\#\{(i,j):i\notin I,\ j\in I,\ t_i>t_j\}
 +(p-r)\#\{i\notin I:t_i>0\}
\end{equation}
\end{theorem}
\begin{proof}
Let $E\in\R^{d\times p}$ be an arbitrary
perturbation of the factor $W$ and define
$D=WE^\top+EW^\top$. Along the straight
factor path $W+\theta E$, the corresponding
student matrix is
\begin{equation*}
 S(\theta)=(W+\theta E)(W+\theta E)^\top
 =S+\theta D+\theta^2EE^\top
\end{equation*}
and hence $S'(0)=D$, $S''(0)=2EE^\top$.
Applying the chain rule twice to
$\mathcal J(W+\theta E)=\mathcal F_{P,G}(S(\theta))$
and using \eqref{eq:matrix-hessian}, we obtain
\begin{equation}
 D^2\mathcal J(W)[E,E]
 =\E\,[P'(f_S)f_D^2]
 +2\langle\nabla\mathcal F_{P,G}(S),EE^\top\rangle_{\mathrm F}
 \label{eq:factor-hessian}
\end{equation}

We first choose coordinates adapted to the
active spectral subspace of the stationary
matrix $S$. Orthogonal changes of input
coordinates, accompanied by the corresponding
change of basis for the teacher matrix, and
orthogonal changes of factor coordinates
preserve the objective. Since the induced
transformations of the factor perturbations
are isometries with respect to the Frobenius
inner product, they also preserve the Hessian
eigenvalues. We may therefore relabel the
active spectral indices as $I=\{1,\ldots,r\}$
and choose input and factor coordinates in
which
\begin{equation*}
 W=\begin{pmatrix}
 C&0\\
 0&0
 \end{pmatrix},
 \qquad
 C=\diag(\sqrt{\mu_1},\ldots,\sqrt{\mu_r})
\end{equation*}
Partitioning the perturbation $E$ according
to the active and inactive spectral directions
and the first $r$ and remaining $p-r$ factor
coordinates yields
\begin{equation*}
 E=\begin{pmatrix}
 E_{11}&E_{12}\\
 E_{21}&E_{22}
 \end{pmatrix}
\end{equation*}
in which $E_{11}\in\R^{r\times r}$, $E_{12}\in\R^{r\times(p-r)}$, $E_{21}\in\R^{(d-r)\times r}$ and $E_{22}\in\R^{(d-r)\times(p-r)}$, with empty blocks interpreted in the usual
way when $r=0$ or $r=p$. In these coordinates,
the first variation of the student matrix is
\begin{equation*}
 D=\begin{pmatrix}
 CE_{11}^\top+E_{11}C&CE_{21}^\top\\
 E_{21}C&0
 \end{pmatrix}
\end{equation*}
In particular, neither $E_{12}$ nor $E_{22}$
contributes to the first variation $D$,
since the corresponding columns of $W$
are zero.

We now examine separately the two terms
on the right-hand side of
\eqref{eq:factor-hessian}. By the expression
for the matrix gradient at a stationary
factor in \eqref{eq:stationary-gradient},
the second term is
\begin{equation*}
 2\langle\nabla\mathcal F_{P,G}(S),EE^\top\rangle_{\mathrm F}
 =-2\sum_{i\notin I}t_i\sum_{a=1}^p E_{ia}^2
\end{equation*}
Thus, only the blocks $E_{21}$ and $E_{22}$
contribute to this term, and there are no
mixed terms between their individual entries.

To examine the first term in
\eqref{eq:factor-hessian}, write
$X_I=(X_1,\ldots,X_r)^\top$ and define
$M=CE_{11}^\top+E_{11}C$. From the block
representation of $D$, we have
\begin{equation*}
 f_D(X)=X_I^\top M X_I-\tr M
 +2\sum_{i\notin I,j\in I}\sqrt{\mu_j}E_{ij}X_iX_j
\end{equation*}
The first two terms depend only on $E_{11}$
and the active Gaussian coordinates, while
the remaining term depends on $E_{21}$
and involves both active and inactive
Gaussian coordinates. Defining the quadratic
form associated with $E_{11}$ by
\begin{equation*}
 Q_{11}(E_{11})
 =\E\,[P'(f_S)(X_I^\top M X_I-\tr M)^2]
\end{equation*}
since $P'\geq0$, this quadratic form is
nonnegative.

The mixed contribution of $E_{11}$ and
$E_{21}$ to the first term in
\eqref{eq:factor-hessian} vanishes. Indeed,
for any $i\notin I$ and $j\in I$, the
expectation
\begin{equation*}
 \E\,[P'(f_S)(X_I^\top M X_I-\tr M)X_iX_j]
\end{equation*}
is zero by virtue of reflection symmetry in the
inactive coordinate $X_i$, since all
factors other than $X_i$ are independent
of that coordinate. Furthermore, for
distinct index pairs $(i,j)$ and $(i',j')$,
with $i,i'\notin I$ and $j,j'\in I$, we have
\begin{equation*}
 \E\,[P'(f_S)X_iX_jX_{i'}X_{j'}]=0
\end{equation*}
To see this, note that at least one
Gaussian coordinate occurs to an odd
power in the product
$X_iX_jX_{i'}X_{j'}$, while
\begin{equation*}
 f_S(X)=\sum_{j\in I}\mu_j(X_j^2-1)
\end{equation*}
is invariant under reflection in every
Gaussian coordinate. Reflection symmetry
therefore eliminates all mixed terms
between distinct entries of $E_{21}$.
As a result, the first term in
\eqref{eq:factor-hessian} takes the form
\begin{equation*}
 \E\,[P'(f_S)f_D^2]
 =Q_{11}(E_{11})
 +4\sum_{i\notin I,j\in I}\mu_j
 \E\,[P'(f_S)X_i^2X_j^2]E_{ij}^2
\end{equation*}

Combining the two terms of
\eqref{eq:factor-hessian}, we see that
the Hessian quadratic form restricted
to the block $E_{21}$ is diagonal in
the standard matrix basis. Its coefficient
associated with the entry $(i,j)$ is
\begin{equation*}
 4\mu_j\E\,[P'(f_S)X_i^2X_j^2]-2t_i
\end{equation*}
and this coefficient can be identified explicitly
using the spectral rotation considered in
the proof of Theorem~\ref{thm:landscape}.
In the chosen coordinates, the tangent
to the rotation of an active direction
$e_j$ towards an inactive direction $e_i$
has only one nonzero entry, of magnitude
$\sqrt{\mu_j}$, and at position $(i,j)$.
The corresponding Hessian Rayleigh
quotient was shown to be $2(t_j-t_i)$.
Since the rotation calculation is valid
for any active index $j$ and inactive
index $i$, irrespective of the ordering
of $t_i$ and $t_j$, we obtain
\begin{equation*}
 4\mu_j\E\,[P'(f_S)X_i^2X_j^2]-2t_i
 =2(t_j-t_i)
\end{equation*}

Finally, the block $E_{12}$ contributes
to neither term of \eqref{eq:factor-hessian},
since it does not enter into the first variation
$D$ and its rows correspond to active
spectral directions on which the matrix
gradient vanishes. The block $E_{22}$
contributes only through the second term,
with coefficient $-2t_i$ for each entry
in the $i$th row.

Collecting the contributions from the
four blocks, we obtain the orthogonal
decomposition of the Hessian quadratic form
\begin{align*}
 D^2\mathcal J(W)[E,E]
 =Q_{11}(E_{11})
 +2\sum_{i\notin I,j\in I}(t_j-t_i)E_{ij}^2-2\sum_{i\notin I}t_i
 \sum_{a=r+1}^p E_{ia}^2
\end{align*}
The four block subspaces are mutually
orthogonal with respect to the Frobenius
inner product and the quadratic form
contains no mixed terms between these
subspaces. Hence, the Hessian is block
diagonal with respect to this decomposition.
The restriction to $E_{11}$ is positive
semidefinite and the restriction to
$E_{12}$ is identically zero, so neither
block contributes negative eigenvalues.
The restrictions to $E_{21}$ and $E_{22}$
are diagonal in their respective standard
matrix bases and their eigenvalues are
$2(t_j-t_i)$ for each pair $i\notin I$,
$j\in I$ and $-2t_i$ repeated $p-r$
times for each $i\notin I$, respectively.
Counting the strictly negative eigenvalues
in these two blocks yields
\begin{equation*}
 n_-(W)=\#\{(i,j):i\notin I,\ j\in I,\ t_i>t_j\}
 +(p-r)\#\{i\notin I:t_i>0\}
\end{equation*}
which completes the proof.
\end{proof}

\begin{example}
Consider the case $d=3$, $p=2$ and
$\lambda_1>\lambda_2>\lambda_3>0$.
By Corollary~\ref{cor:stationary-count},
there are seven distinct stationary matrices.
The numbers of negative Hessian eigenvalues
at the corresponding stationary factors,
as determined by Theorem~\ref{thm:hessian-index},
are
\begin{center}
\begin{tabular}{c|rrrrrrr}
Active set $I$ & $\varnothing$ & $\{1\}$ & $\{2\}$ &
$\{3\}$ & $\{1,2\}$ & $\{1,3\}$ & $\{2,3\}$ \\
\hline
$n_-(W)$ & 6 & 2 & 3 & 4 & 0 & 1 & 2
\end{tabular}
\end{center}
At each rank-one stationary matrix, two negative
Hessian eigenvalues arise from perturbations in
the unused factor direction, with any additional
negative eigenvalues arising from spectral
inversions between active and inactive teacher
directions. At the rank-two stationary matrices,
only spectral inversions contribute negative
Hessian eigenvalues. The stationary matrix
associated with the leading active set
$\{1,2\}$ is the unique globally optimal
student matrix.
\end{example}

\begin{remark}
The negative Hessian eigenvalues identified in
Theorem~\ref{thm:hessian-index} are exact, but
their magnitudes are not necessarily bounded away from
zero as the gaps between teacher moment
eigenvalues decrease, or as the inverse
temperature varies. Moreover, the absence of
nonglobal second-order stationary points does
not in itself guarantee convergence of gradient
descent to a global minimum from arbitrary
initial conditions. Such convergence results
require additional hypotheses on the optimisation
algorithm and its initialisation, which are not
considered in the present study. Finally, the
characterisation of the population optimisation
landscape is established for finite positive
inverse temperature and no assertion of coercivity
is made in the deterministic separable case.
\end{remark}

\section{Numerical illustrations}\label{sec:experiments}
Here we present  numerical experiments that 
examine the accuracy of the sharp asymptotic
equivalents established in
Theorems~\ref{thm:exact-crossover} and~\ref{thm:regular}
at finite student rank, in order to characterise the
transition between the regular and singular
spectral scaling regimes and to assess the extent
to which the explicit calibration
\eqref{eq:calibration} approximates the minimum
population regret over the entire admissible
student class. We consider two families of
teacher spectra, with algebraically and
geometrically decaying eigenvalues given,
respectively, by
\begin{equation}
 \lambda_j=j^{-\gamma},\quad \gamma>1/2,
 \qquad
 \lambda_j=r^{j-1},\quad 0<r<1
 \label{eq:num-spectra}
\end{equation}
The geometric spectrum is of particular interest
since the normalised omitted quadratic form does
not converge in distribution to a Gaussian random
variable as the retained rank tends to infinity
with an unbounded number of omitted spectral
directions. This family therefore provides a
numerical assessment of the sharp asymptotic
equivalents in a setting in which a Gaussian
approximation of the unresolved component is
not asymptotically justified. All numerical
calculations are performed at finite positive
inverse temperatures and relate to population
expectations rather than empirical risks
computed from finite training samples.

\subsection{Numerical methodology}

By Proposition~\ref{prop:alignment}, minimisation
of the population objective over the admissible
student class reduces to minimisation over the
nonnegative student coefficients associated with
the leading $p$ teacher eigenvectors. We denote
by $\widehat{\mathcal E}_{p,\tau}$ the resulting
numerical approximation to the optimal population
regret $\mathcal E_{p,\tau}$. We also compute
the population regrets $\mathcal E^{\rm cal}_{p,\tau}$
and $\mathcal E^{\rm sc}_{p,\tau}$ attained by
the explicitly calibrated spectral truncation
and an independently optimised positive scalar
multiple of the spectral truncation, respectively.
Comparison of these three quantities allows us
to distinguish the error arising from the choice
of the overall scale of the student logit from
that arising from the restriction of the
student coefficients to the ratios of the
retained teacher eigenvalues. By construction,
the exact population regrets satisfy
\begin{equation}
 \mathcal E_{p,\tau}\leq\mathcal E^{\rm sc}_{p,\tau}
 \leq\mathcal E^{\rm cal}_{p,\tau}
\end{equation}

The expectations required for the optimisation
of the student coefficients and the subsequent
evaluation of the population regret are computed
by one-dimensional Fourier inversion using the
characteristic function of a centred Gaussian
quadratic form. The derivation of the Fourier
integral representations is given in
Appendix~\ref{app:numerics}. In the numerical
implementation we parametrise the student
coefficients as $\mu_j=\alpha_p a_j$, in which 
$\alpha_p$ is given by \eqref{eq:calibration}.
Restricting initially to coefficients of the
form $a_j=x\lambda_j$, we solve the resulting
scalar stationarity equation by a bracketed
root search over $1/2\leq x\leq2$. The
coefficients obtained from the scalar
optimisation are subsequently used to
initialise a Newton iteration for the
full reduced objective in the variables
$a_1,\ldots,a_p$. The Hessian is diagonally
equilibrated prior to its eigendecomposition
and the Newton step is restricted to preserve
positivity of the coefficients and to limit
their changes relative to the retained teacher
coefficients. A line search based on directional
derivatives is employed to avoid the subtraction
of nearly equal objective values.

Denoting by $g$ and $B$ the gradient and Hessian,
respectively, of the population objective divided
by $\alpha_p$ with respect to the variables $a_j$,
we terminate the Newton iteration when
\begin{equation}
 \frac{\alpha_p g^\top B^{-1}g}
 {\rho_d(0)s_p\Phi(\eta)}<10^{-14}
 \label{eq:num-stop}
\end{equation}
in which $\rho_d$ is the density of the
finite-dimensional teacher signal $F$ and the numerator is the squared Newton decrement
in the original objective scale and the
denominator is the theoretical leading-order
approximation to the optimal population regret
with $\rho_0$ replaced by $\rho_d(0)$.
The gradient at the resulting coefficients
is subsequently evaluated on an independently
refined Fourier grid. It should be emphasised  that the
Newton decrement provides a local convergence
diagnostic rather than a certified upper bound
on the global optimisation error.

In order to avoid cancellation errors associated with
subtracting the teacher Bayes risk from the
student population loss, the population regret
is evaluated directly using the Bregman integral
representation \eqref{eq:num-direct-kl} and
the weighted Fourier integral
\eqref{eq:num-weighted}.  For the main power-law experiments,
composite Gauss-Legendre quadrature with
16 nodes per panel is employed for the
frequency integrals and Gauss-Legendre
quadrature of order 48 or 96 is employed
for integration along the coefficient segment.
The accuracy of the frequency quadrature is
assessed in selected cases by halving the panel
spacing, while an additional calculation with
quadrature of order 192 along the coefficient
segment is performed to assess the sensitivity
to the latter discretisation. The geometric
spectral experiments employ quadrature of
order 96 along the coefficient segment and
frequency refinement level two, with the
exception of the largest-rank case with
$r=0.5$ and $\eta=10$, for which the imposed
limit on the number of frequency panels
necessitates the use of refinement level one.

Throughout the numerical experiments, the
normalisations are computed using the actual
finite-dimensional residual standard deviation
$s_p$ and density value $\rho_d(0)$ rather
than their infinite-spectrum counterparts.
By the Fourier inversion argument used in
the proof of Lemma~\ref{lem:central-density},
applied to the full finite-dimensional teacher
signal, we have $\rho_d(0)\to\rho_0$ as
$d\to\infty$. Replacing $\rho_0$ by $\rho_d(0)$
in the theoretical predictions therefore
preserves the asymptotic equivalents while
accounting for the finite ambient dimension
in the numerical comparisons.

\subsection{The crossover function}

\begin{figure}[htbp]
\centering
\includegraphics[width=\linewidth]{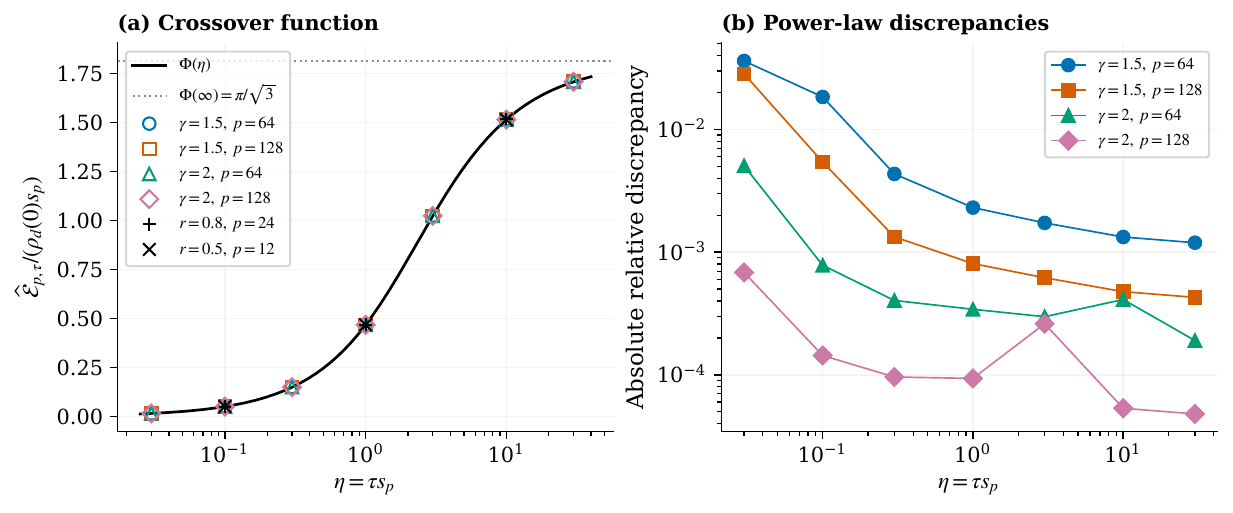}
\caption{Comparison of the numerically computed
normalised population regret
$\widehat{\mathcal E}_{p,\tau}/(\rho_d(0)s_p)$
with the crossover function $\Phi(\eta)$ for
power-law and geometric spectra (left) and
absolute relative discrepancies
$|\widehat{\mathcal E}_{p,\tau}/
(\rho_d(0)s_p\Phi(\eta))-1|$ for the power-law
spectra (right). The horizontal dotted line
represents the theoretical limiting value
$\Phi(\infty)=\pi/\sqrt{3}$ rather than the
result of a deterministic-label computation.}
\label{fig:num-crossover}
\end{figure}

We first consider the power-law spectra with
$\gamma\in\{1.5,2\}$, fixing the ambient dimension
at $d=1024$ and taking student ranks $p=64,128$.
For each combination of $\gamma$ and $p$, we
perform calculations for
$\eta\in\{0.03,0.1,0.3,1,3,10,30\}$, choosing
the inverse temperature according to
$\tau=\eta/s_p$. Figure~\ref{fig:num-crossover}
compares the normalised numerical approximation
to the optimal population regret,
$\widehat{\mathcal E}_{p,\tau}/(\rho_d(0)s_p)$,
with the crossover function $\Phi(\eta)$ given
by \eqref{eq:exact-crossover}. Also included
are the results for the largest student ranks
considered for the geometric spectra discussed
below. The agreement with the crossover
prediction throughout the range of $\eta$
considered is particularly close for $\gamma=2$
and $p=128$, for which the absolute relative
discrepancies are less than $0.07\%$ after
the selected quadrature refinements.

The larger discrepancies at small $\eta$ for
the more slowly decaying power-law spectrum
can be understood in terms of the conditions
under which the asymptotic equivalents were
established. Although small $\eta$ places
the system in the regular regime, it does
not necessarily imply that the inverse
temperature is sufficiently large for the
low-temperature equivalent of
Corollary~\ref{cor:unified} to provide an
accurate approximation at finite rank.
In particular, for $\gamma=1.5$, $p=64$
and $\eta=0.03$, the absolute relative
discrepancy from the crossover prediction
is approximately $3.6\%$, whereas the
corresponding discrepancy from the regular
prediction based on Theorem~\ref{thm:regular}
is approximately $0.11\%$. These results
illustrate the importance of distinguishing
between the regular regime characterised
by small $\tau s_p$ and the joint large-rank
and low-temperature limit.

\subsection{Regular and near-singular spectral scaling}

In order to examine the dependence of the optimal population
regret on student width, we fix the ambient
dimension at $d=4096$ and consider the power-law
spectra with $\gamma\in\{1,1.5,2\}$ for student
ranks $p\in\{8,16,32,64,128,256\}$. Calculations
are performed at fixed inverse temperatures
$\tau=1,10$ and along the near-singular sequence
$\tau_p=30/s_p$. The predictions based on
Theorems~\ref{thm:regular} and
\ref{thm:exact-crossover} used for the regular
and near-singular comparisons, respectively,
are
\begin{equation}
 P_{\rm reg}=\frac{\tau^2s_p^2}{2}\,
 \E\,[\sig'(\tau F)],\qquad
 P_{\rm ns}=\rho_d(0)s_p\Phi(30)
 \label{eq:num-predictions}
\end{equation}
where all expectations and spectral quantities
are evaluated for the finite-dimensional
teacher. The use of these quantities accounts
for the leading effect of truncating the
infinite teacher spectrum when assessing the
accuracy of the theoretical predictions.
Along the sequence with $\eta=30$, the
asymptotic width exponent coincides with that
obtained for deterministic labels, although
the coefficient multiplying $\rho_0s_p$ is
$\Phi(30)\simeq1.70745$ rather than
$\Phi(\infty)=\pi/\sqrt{3}$. Consequently,
these calculations examine the approximation
law in the vicinity of the singular regime
rather than directly at the deterministic
endpoint.

\begin{figure}[htbp]
\centering
\includegraphics[width=\linewidth]{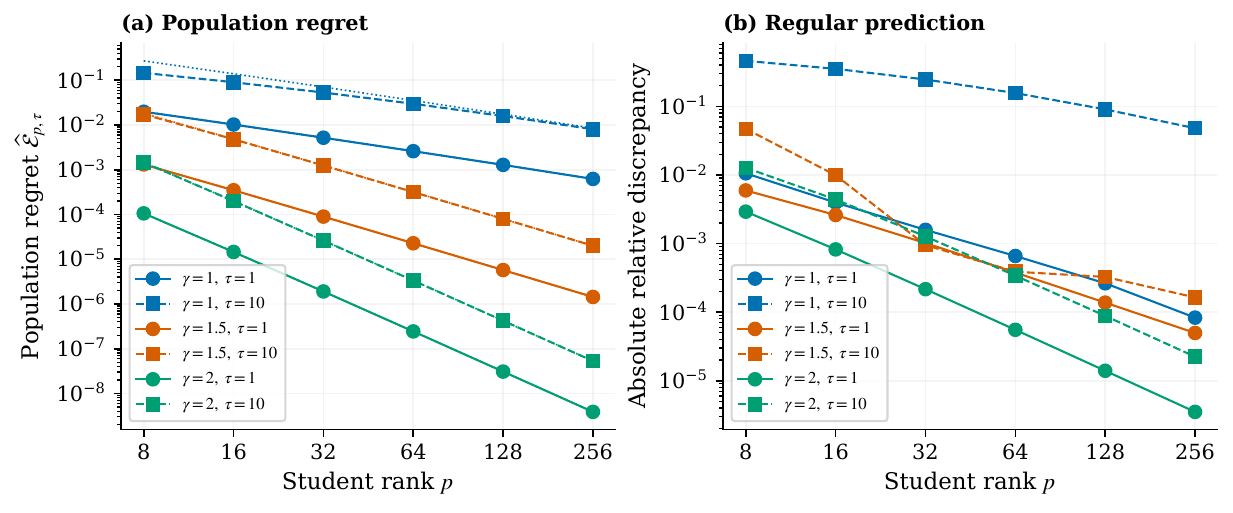}
\caption{Numerically computed population regret
$\widehat{\mathcal E}_{p,\tau}$ at fixed inverse
temperatures $\tau=1,10$ for the three power-law
spectra (left) and absolute relative discrepancies
$|\widehat{\mathcal E}_{p,\tau}/P_{\rm reg}-1|$
from the regular prediction (right). The dotted
curves in the left panel represent the
finite-dimensional predictions $P_{\rm reg}$
given by \eqref{eq:num-predictions}.}
\label{fig:num-regular}
\end{figure}

\begin{figure}[htbp]
\centering
\includegraphics[width=\linewidth]{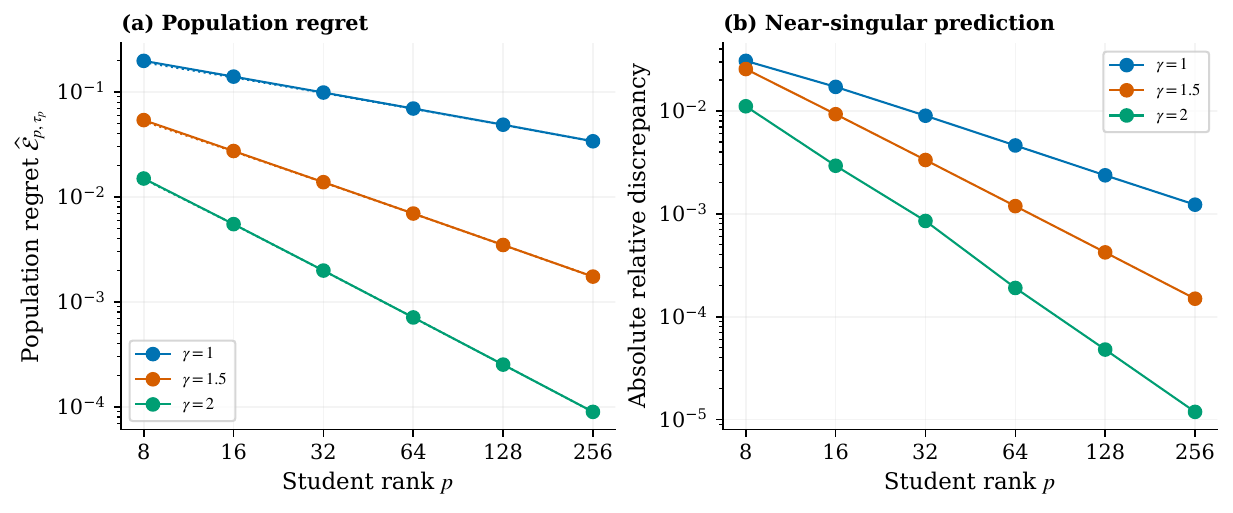}
\caption{Numerically computed population regret
$\widehat{\mathcal E}_{p,\tau_p}$ along the
near-singular sequence $\tau_p=30/s_p$ for the
three power-law spectra (left) and absolute
relative discrepancies
$|\widehat{\mathcal E}_{p,\tau_p}/P_{\rm ns}-1|$
from the near-singular prediction (right).
The dotted curves in the left panel represent
the finite-dimensional predictions $P_{\rm ns}$
given by \eqref{eq:num-predictions}. Refined
frequency quadrature is used for $p=64,128,256$
when $\gamma=2$.}
\label{fig:num-singular}
\end{figure}

Figures~\ref{fig:num-regular} and
\ref{fig:num-singular} show the numerically
computed population regrets and the corresponding
absolute relative discrepancies from the
predictions in \eqref{eq:num-predictions}.
At $p=256$, the relative discrepancy from the
regular prediction is less than $0.017\%$ for
five of the six combinations of $\gamma$ and
$\tau$. The exception is the sequence with
$\gamma=1$ and $\tau=10$, for which the
discrepancy is approximately $4.8\%$ and
$\eta=\tau s_p$ is still approximately $0.85$.
The slower decay of the residual standard
deviation for this spectrum delays the approach
to the asymptotic regular regime. To interpret
the resulting intermediate width dependence,
consider the low-temperature approximation
$\rho_0s_p\Phi(\tau s_p)$ and replace the
residual standard deviation by its pure
power-law equivalent $s_p=A_\gamma p^{-\beta}$,
in which  $\beta=\gamma-1/2$. Logarithmic
differentiation with respect to $p$, treating
the rank as a continuous variable and holding
$\tau$ fixed, yields the effective width exponent
\begin{equation}
 -\frac{\mathrm d\log(s_p\Phi(\tau s_p))}
 {\mathrm d\log p}
 =\beta\left(1+\frac{1}
 {\sqrt{1+3\eta^2/\pi^2}}\right)
 \label{eq:num-effective}
\end{equation}
which varies between $\beta$ in the singular
limit $\eta\to\infty$ and $2\beta$ in the regular
limit $\eta\to0$. This expression provides an
interpretation of the intermediate exponents
observed at finite rank rather than an exact
logarithmic derivative of the optimal
population regret.

\begin{table}[htbp]
\centering\small
\resizebox{\linewidth}{!}{\begin{tabular}{rrrrrrrr}
\hline
$\gamma$ & Reg. theory & Reg. spectral & $\tau=1$ & $\tau=10$ & NS theory & NS spectral & NS risk \\
\hline
1 & 1 & 1.02038 & 1.01968 & 0.90816 & 0.5 & 0.51019 & 0.51386 \\
1.5 & 2 & 1.98882 & 1.98927 & 1.98835 & 1 & 0.99441 & 0.99590 \\
2 & 3 & 2.98062 & 2.98072 & 2.98119 & 1.5 & 1.49031 & 1.49069 \\
\hline
\end{tabular}
}
\caption{Comparison of theoretical and numerically
estimated width exponents. The numerical regret
exponents are obtained as the negatives of the
least-squares slopes of log population regret
against log rank over $p=32,64,128,256$. The
spectral exponents are obtained by the same
procedure applied to $s_p^2$ and $s_p$ for the
regular and near-singular regimes, respectively,
at $d=4096$. NS denotes the near-singular
sequence $\eta=30$.}
\label{tab:num-exponents}
\end{table}

The fitted exponents in Table~\ref{tab:num-exponents}
quantify the change in the width dependence of
the optimal population regret between the two
scaling regimes. With the exception of the
sequence with $\gamma=1$ and $\tau=10$, which
has not reached the asymptotic regular regime
over the range of ranks considered, the fitted
regular regret exponents differ from the
corresponding finite-dimensional squared-tail
exponents by less than $0.001$. For the
near-singular sequences, the differences
between the fitted regret exponents and the
finite-dimensional tail exponents are less
than $0.004$. The comparison with the
finite-dimensional spectral exponents is
important since the fitted slopes of the
spectral tails themselves differ from the
theoretical infinite-spectrum exponents over
the finite range of ranks considered here.

\subsection{Calibration and non-Gaussian residuals}

\begin{figure}[htbp]
\centering
\includegraphics[width=\linewidth]{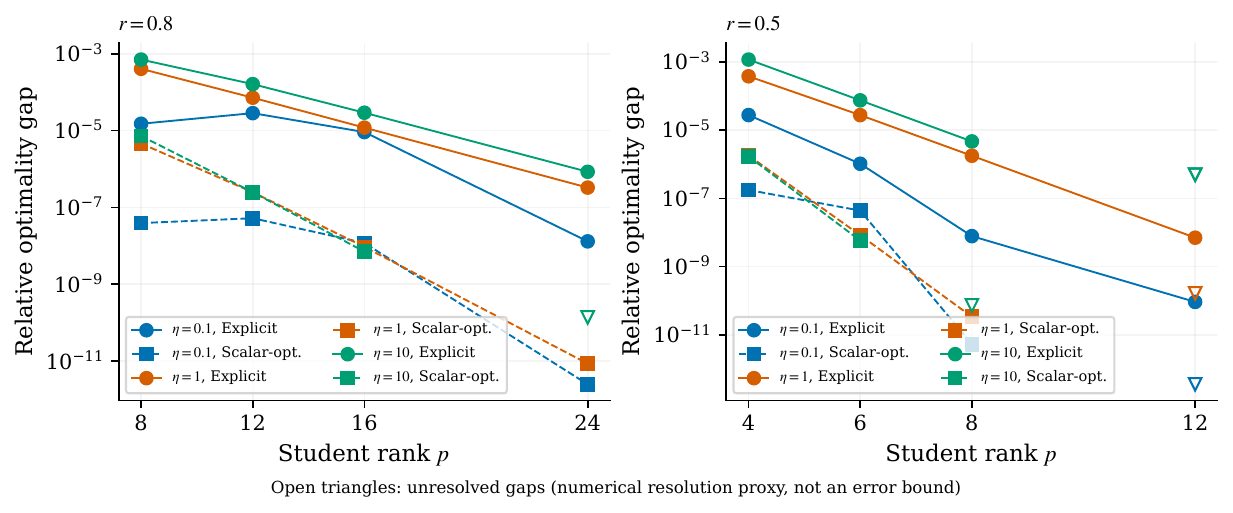}
\caption{Relative optimality gaps of the explicitly
calibrated and independently optimised scalar
multiples of the spectral truncation with respect
to the numerically optimised population regret
for geometric spectra with $r=0.8$ (left) and
$r=0.5$ (right). Open downward triangles denote
numerically unresolved gaps, plotted at the
empirical resolution proxy or the absolute
value of the computed gap, whichever is larger.
The plotted values of the unresolved gaps do
not constitute certified upper bounds.}
\label{fig:num-optimality}
\end{figure}

We next consider the geometric spectra with ambient
dimension $d=128$, taking $r=0.8$ with student
ranks $p\in\{8,12,16,24\}$ and $r=0.5$ with
$p\in\{4,6,8,12\}$. For each combination of
$r$ and $p$, calculations are performed at
$\eta\in\{0.1,1,10\}$.
Figure~\ref{fig:num-optimality} compares the
relative optimality gaps
\begin{equation*}
 \frac{\mathcal E^{\rm cal}_{p,\tau}}
 {\widehat{\mathcal E}_{p,\tau}}-1,\qquad
 \frac{\mathcal E^{\rm sc}_{p,\tau}}
 {\widehat{\mathcal E}_{p,\tau}}-1
\end{equation*}
associated with the explicit calibration and
the independently optimised scalar multiple
of the spectral truncation, respectively.
The gaps generally decrease with increasing
student rank. Moreover, where the gaps can
be reliably resolved, the further reduction
in population regret obtained by allowing
the retained student coefficients to vary
independently is generally considerably
smaller than that obtained by optimising
their common scale. The smallest gaps cannot
be reliably resolved by subtraction of the
directly computed population regrets. To
distinguish these cases from resolved positive
gaps, we employ the empirical numerical
resolution proxy defined in
\eqref{eq:num-resolution-proxy}. Nonpositive gaps
and positive gaps not exceeding this proxy
are marked separately. The resulting
representation is a plotting convention
rather than a rigorous upper bound on the
optimality gaps. In particular, there is
no separate independent quadrature check
of the regret difference between the
scalar-optimised and fully optimised
student models.

For the infinite geometric spectrum, the
variance-normalised omitted quadratic form
has the distribution
\begin{equation}
 G_p\overset{\mathrm d}{=}
 \sqrt{\frac{1-r^2}{2}}
 \sum_{k=0}^\infty r^k(X_{k+1}^2-1)
\end{equation}
which is independent of the retained rank $p$.
For $m\geq2$, the $m$th cumulant of the centred
chi-squared random variable $X_j^2-1$ is
$2^{m-1}(m-1)!$. By independence of the Gaussian
coordinates, additivity of cumulants and
their homogeneity under scalar multiplication,
the cumulants of the normalised residual are
therefore given by
\begin{equation*}
 \kappa_m(G_p)=
 \left(\frac{1-r^2}{2}\right)^{m/2}
 \frac{2^{m-1}(m-1)!}{1-r^m}
\end{equation*}
Since $G_p$ is centred and has unit variance,
its skewness and excess kurtosis are given
by the third and fourth cumulants, respectively,
yielding
\begin{equation}
 \operatorname{Skew}(G_p)
 =\frac{2\sqrt2(1-r^2)^{3/2}}{1-r^3},
 \qquad
 \operatorname{Kurt}_{\rm excess}(G_p)
 =12\frac{1-r^2}{1+r^2}
\end{equation}
In particular, the excess kurtosis is
approximately $2.634$ for $r=0.8$ and $7.2$
for $r=0.5$. Moreover, the fraction of the residual variance
attributable to the first omitted spectral
direction is
\begin{equation*}
 \frac{\lambda_{p+1}^2}
 {\sum_{j=p+1}^\infty\lambda_j^2}=1-r^2
\end{equation*}
independently of $p$. For $r=0.8$ and $r=0.5$,
this fraction is $36\%$ and $75\%$, respectively.

\begin{figure}[htbp]
\centering
\includegraphics[width=\linewidth]{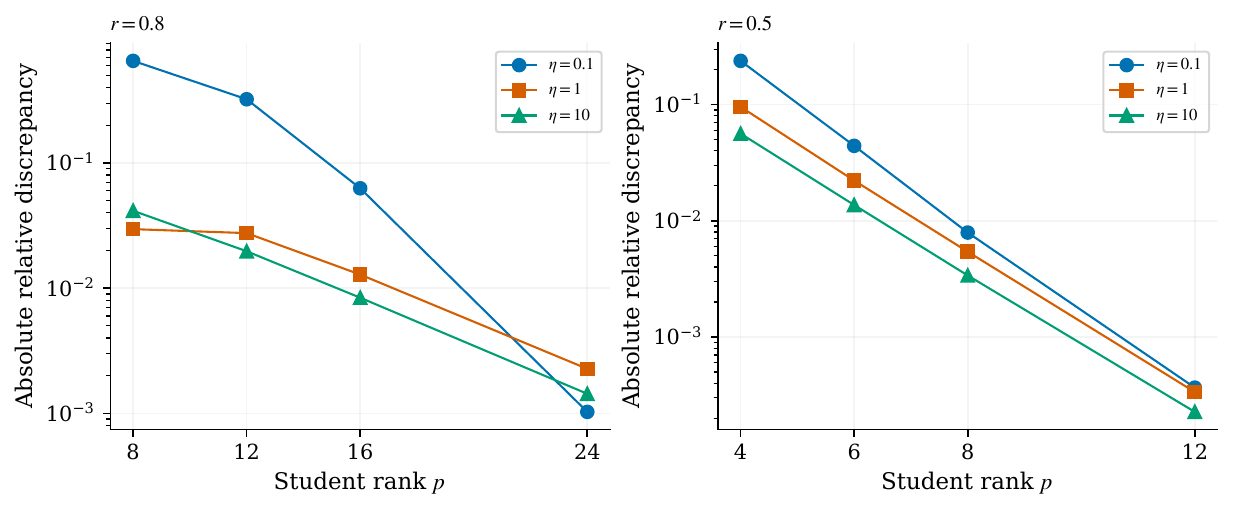}
\caption{Absolute relative discrepancies
$|\widehat{\mathcal E}_{p,\tau}/
(\rho_d(0)s_p\Phi(\eta))-1|$
from the crossover prediction for geometric
spectra with $r=0.8$ (left) and $r=0.5$ (right)
at fixed crossover parameters
$\eta=0.1,1,10$. In both cases the normalised
omitted quadratic form has a non-Gaussian
limiting distribution in the infinite-spectrum
model.}
\label{fig:num-geometric}
\end{figure}

The distribution of the normalised
finite-dimensional residual approaches the
infinite-spectrum distribution as the number
of omitted spectral directions tends to
infinity. Thus, unlike the case in which
the relative contributions of the individual
omitted eigenvalues to the residual variance
become negligible, the normalised residual
for a geometric spectrum remains non-Gaussian
as the retained rank increases in the
infinite-spectrum model.

The decreasing discrepancies in
Figure~\ref{fig:num-geometric} illustrate
an important feature of
Theorem~\ref{thm:exact-crossover}, namely
that the leading-order approximation law
does not require convergence of the
normalised omitted quadratic form in
distribution to a Gaussian random variable.
At the largest retained ranks, the absolute
relative discrepancies from the crossover
prediction are less than $0.23\%$ for $r=0.8$
and $0.037\%$ for $r=0.5$. The relatively
large discrepancies at the smallest ranks
for $\eta=0.1$ occur at inverse temperatures
for which the low-temperature asymptotic
equivalent is not yet expected to provide
an accurate approximation.

\subsection{Numerical accuracy}

The sensitivity of the computed population
regrets to frequency quadrature, Bregman
path quadrature and ambient dimension is
examined in Table~\ref{tab:num-validation}.
The frequency quadrature comparisons reveal
the importance of increasing the frequency
resolution as the population regret becomes
small. In particular, for the near-singular
sequence with $\gamma=2$ and $p=256$,
refinement of the frequency quadrature
changes the computed population regret
by approximately $4.7\%$ and reduces
the absolute relative discrepancy from
the asymptotic prediction to approximately
$1.2\times10^{-5}$. For $p=128$, increasing
the order of the Bregman path quadrature
from 96 to 192 while maintaining the refined
frequency resolution changes the computed
population regret by approximately
$1.1\times10^{-9}$ in relative terms.

\begin{table}[htbp]
\centering\small
\resizebox{\linewidth}{!}{\begin{tabular}{llrrr}
\hline
Test & Change / reference & First value & Second value & Relative change \\
\hline
Fourier refinement, $p=64$ & 1 to 2; path 96 & 0.000711908155634 & 0.000711346748587 & 7.886e-04 \\
Fourier refinement, $p=128$ & 1 to 2; path 96 & 0.000254514733659 & 0.000252940323968 & 6.186e-03 \\
Fourier refinement, $p=256$ & 1 to 2; path 96 & 9.41171593154e-05 & 8.9677682867e-05 & 4.717e-02 \\
Path quadrature, $p=128$ & 96 to 192; Fourier 2 & 0.000252940323968 & 0.000252940324239 & 1.072e-09 \\
Dimension, $\gamma=1$ & $1024$ to $8192$ & 1.71188727 & 1.71142722 & 2.687e-04 \\
Dimension, $\gamma=2$ & $1024$ to $4096$ & 1.70753111 & 1.70753093 & 1.060e-07 \\
Density & Independent reference & 0.0976834074 & 0.0976834074 & 5.329e-15 \\
KL & Independent reference & 0.1790520957 & 0.1790520957 & 2.821e-11 \\
Entropy & Independent reference & 0.3853834122 & 0.3853834122 & 1.882e-11 \\
Moment & Independent reference & 0.1641443419 & 0.1641443419 & 8.605e-12 \\
\hline
\end{tabular}
}
\caption{Numerical verification of the population regret
calculations. The first three rows compare frequency
quadrature refinement levels at fixed Bregman path
quadrature order 96 for $\gamma=2$, $d=4096$ and
$\eta=30$. The fourth row compares path quadrature
orders 96 and 192 at fixed frequency refinement level
two for $p=128$ and the same teacher spectrum and
crossover parameter. The first and second values are
the unnormalised population regrets
$\widehat{\mathcal E}_{p,\tau}$ for the frequency
and path quadrature comparisons and the normalised
population regrets
$\widehat{\mathcal E}_{p,\tau}/(\rho_d(0)s_p)$
for the ambient dimension comparisons at $p=128$
and $\eta=30$. In the final four rows the first
and second values are obtained by Fourier inversion
and independent reference calculations, respectively,
for the equal-weight chi-squared tests described in
Appendix~\ref{app:num-independent}. The relative discrepancies
are computed as $|x_2/x_1-1|$ for the quadrature
refinement and ambient dimension comparisons and
$|x_1/x_2-1|$ for the independent integration tests.
All relative discrepancies are calculated before rounding
the displayed values and do not constitute certified
numerical error estimates.}
\label{tab:num-validation}
\end{table}
\FloatBarrier

The ambient dimension comparisons indicate
that the normalised population regret is
relatively insensitive to the truncation
of the teacher spectrum over the range
of dimensions considered. The relative
changes in the normalised population regret
are approximately $2.7\times10^{-4}$ and
$1.1\times10^{-7}$ for $\gamma=1$ and
$\gamma=2$, respectively. Additional
comparisons with independent integration
methods are included in
Table~\ref{tab:num-validation} to assess
the accuracy of the Fourier integral
representations used in the numerical
calculations. Further details of these
comparisons and the quadrature procedures
are given in Appendix~\ref{app:numerics}.
Collectively, these numerical checks
provide evidence of the accuracy of the
computed population regrets but do not
constitute certified bounds on the
combined optimisation and quadrature error.
\FloatBarrier
\section{Discussion and concluding remarks}\label{sec:discussion}

The analysis presented in this paper identifies a
mechanism by which nonlinear approximation undergoes
a transition from variance-controlled to standard-deviation-controlled
behaviour without a change in the underlying signal
spectrum or the loss function. The transition is
governed by the relative magnitudes of the unresolved
fluctuations and the width of the probabilistic
transition layer of the teacher response. For
independent unresolved fluctuations, localisation
in a neighbourhood of the decision boundary reduces
the population approximation problem to an integrated
threshold-loss optimisation. Under a canonical proper
loss, the integrated objective depends on the threshold
distribution only through its first two moments,
allowing the variational problem to be solved explicitly
in terms of the effective latent variance. The converse
characterisation of this moment dependence identifies
the distinguished role of canonical loss coordinates
in the integrated approximation problem.

The sharp matched-link approximation theorem establishes
an explicit asymptotic equivalent throughout the joint
limit of vanishing residual variance and increasing
inverse temperature, including the regular regime in
which the leading-order contributions to the population
loss and the teacher Bayes risk cancel upon subtraction.
The resulting crossover function connects the quadratic
and linear dependence of the population regret on the
residual standard deviation in the regular and singular
regimes, respectively, with a temperature-dependent
coefficient in the regular regime. Moreover, the
explicit calibration is asymptotically optimal without
requiring convergence of the normalised unresolved
fluctuations to a fixed limiting distribution. Although
the canonical-link and cumulative entropy identities
have antecedents in statistical decision theory, their
application to the joint approximation limit requires
localisation, control of the tails, compactness of
near minimisers and a separate relative-error analysis
in the regular regime.

The decomposition of the population regret into
representation and decoder contributions clarifies
the nature of the universality of the approximation
law. In the matched regular regime, the leading-order
regret is already present for an unrestricted measurable
decoder of the retained representation, while in
the crossover and singular regimes the individual
contributions generally retain information about the
distribution of the unresolved fluctuations. Their
sum is, nevertheless, determined at leading order by
the residual variance for fixed teacher and student
links and a given inverse temperature and limiting
boundary density. The exact transport representation
quantifies the decoder contribution through a squared
Wasserstein distance to the moment-matched canonical
law. The corresponding asymptotic expansion is
additive and need not provide a relative equivalent
when this distance tends to zero. These results
are in relation to a prescribed retained representation and
do not establish its optimality among alternative
representations. Similarly, the additive expansion
for mismatched inverse links does not establish a
sharp relative equivalent after subtraction of the
intrinsic link misspecification in the regular regime.

For Gaussian quadratic teachers, the spectral alignment
principle and the conditional affine relaxation extend
the sharp scalar approximation law to global minimisation
over the entire class of positive-semidefinite
rank-constrained student matrices, without imposing
proportionality between the retained student coefficients
and the corresponding teacher eigenvalues. The uniform
nonasymptotic estimates additionally control the
approximation error for arbitrary Gaussian projections,
while the sharp logistic regular-regime result extends
the analysis to fixed finite inverse temperatures.
For teacher eigenvalues decaying as $j^{-\gamma}$,
the approximation exponent with respect to student
rank changes from $2\gamma-1$ at fixed positive
inverse temperature to $\gamma-1/2$ for deterministic
labels. The numerical experiments corroborate the
crossover prediction, including for geometric spectra
with strongly non-Gaussian normalised residuals, and
demonstrate the accuracy of the explicit calibration
over the range of ranks and inverse temperatures
considered.

A complementary result speaks to the optimisation
landscape of the factorised population objective at
finite positive inverse temperature. Every local
minimum is global and every nonglobal stationary
factor is a strict saddle, even when the teacher
cannot be represented within the student class.
The stationary matrices are characterised by their
active subspaces within the positive teacher eigenspaces
and are enumerated explicitly for a simple positive
teacher spectrum. Furthermore, the negative Hessian
eigenvalues and their multiplicities are determined
entirely by the teacher moment eigenvalues and the
active spectral subspace, independently of the
nonlinear optimisation of the retained coefficients.
The exact negative Hessian index distinguishes the
unstable directions associated with spectral inversions
from those associated with unused factor directions.
Taken together, the approximation and landscape results
characterise both the magnitude of the optimal
population regret and the stationary obstructions
to attaining its minimum.

The uniform nonasymptotic approximation bounds and
the sharp fixed-temperature coefficients remain
specific to logarithmic loss, while the sharp joint
low-temperature approximation law applies to the
broader class of matched canonical losses under
the stated regularity assumptions. The matrix results
rely on the Gaussian input distribution and the
positive-semidefinite structure of the teacher and
student matrices. Moreover, the optimisation landscape
results do not imply convergence of gradient descent
from arbitrary initial conditions or uniform
separation of the negative Hessian eigenvalues from
zero. The approximation results concern population
objectives rather than statistical estimation from
finite training samples and should not be interpreted
as predictions of empirical neural scaling exponents.
The principal mathematical conclusion is that the
approximation rate depends not only on the spectral
decay of the underlying signal but also on the
geometry of the loss and the relative magnitudes of
the unresolved fluctuations and the probabilistic
transition layer. As a consequence, the spectral
approximation properties of the signal alone do not necessarily
determine the approximation exponent of a
nonlinear prediction problem.

\appendix
\section{Additional Gaussian identities}\label{app:gaussian}

For $X\sim\cN(0,I_d)$, independence and the
second and fourth moments of the standard
normal distribution yield Wick's fourth-moment
identity (see also \citep{janson1997})
\begin{equation*}
 \E\,[X_iX_jX_kX_l]
 =\delta_{ij}\delta_{kl}
 +\delta_{ik}\delta_{jl}
 +\delta_{il}\delta_{jk}
\end{equation*}
Expanding the product of two quadratic forms
and summing over the four indices yields,
for symmetric matrices $A$ and $B$
\begin{equation*}
 \E\,[(X^\top AX)(X^\top BX)]
 =\tr A\tr B+2\tr(AB)
\end{equation*}
Since $\E\,[X^\top AX]=\tr A$, subtraction
of the product of the means yields
\begin{equation}
 \E\,[(X^\top AX-\tr A)(X^\top BX-\tr B)]
 =2\tr(AB)
 \label{eq:gaussian-quadratic-covariance}
\end{equation}
Taking $B=A$ in
\eqref{eq:gaussian-quadratic-covariance}
we obtain
\begin{equation}
 \E\,[(X^\top AX-\tr A)^2]=2\Frob{A}^2
 \label{eq:gaussian-quadratic-variance}
\end{equation}
Therefore, the linear mapping
$A\mapsto (X^\top AX-\tr A)/\sqrt{2}$
is an isometric embedding of the space of
real symmetric matrices equipped with the
Frobenius inner product into
$L^2(\cN(0,I_d))$. Under logarithmic loss,
however, the population regret cannot
generally be expressed as a squared matrix
norm and the geometry of the decision
boundary becomes essential in the
low-temperature limit.

\section{Fourier evaluation of the population objective}\label{app:numerics}
\subsection{Characteristic functions and weighted Gaussian moments}

Let $Z_j=X_j^2-1$ and, for a real coefficient
vector $a=(a_1,\ldots,a_n)$, define the
centred Gaussian quadratic form and its
characteristic function by
\begin{equation*}
 Q_a=\sum_{j=1}^n a_jZ_j,\qquad
 \varphi_a(t)=\E\,[e^{itQ_a}]
\end{equation*}
Independence of the Gaussian coordinates
and evaluation of the Gaussian integral result in
\begin{equation}
 \varphi_a(t)=
 \prod_{j=1}^n e^{-ita_j}(1-2ita_j)^{-1/2}
 \label{eq:num-cf}
\end{equation}
in which each square root is chosen continuously
from its value of unity at $t=0$. Define
\begin{equation*}
 D_j=1-2ita_j,\qquad m_j=\frac{2ita_j}{D_j}
\end{equation*}
Differentiating the characteristic function
with respect to the coefficients yields
\begin{equation*}
 \partial_{a_j}\varphi_a(t)
 =it\E\,[Z_j e^{itQ_a}],\qquad
 \partial_{a_k}\partial_{a_j}\varphi_a(t)
 =(it)^2\E\,[Z_jZ_k e^{itQ_a}]
\end{equation*}
in which differentiation under the expectation
is justified by the integrability of $|Z_j|$
and $|Z_jZ_k|$. 
Logarithmic differentiation of
\eqref{eq:num-cf}, on the other hand, yields
\begin{equation*}
 \partial_{a_j}\log\varphi_a(t)
 =-it+\frac{it}{D_j}=it m_j,\qquad
 \partial_{a_k}m_j=\frac{2it\delta_{jk}}{D_j^2}
\end{equation*}
and so
\begin{equation*}
 \partial_{a_k}\partial_{a_j}\varphi_a(t)
 =(it)^2\varphi_a(t)
 \left(m_jm_k+\frac{2\delta_{jk}}{D_j^2}\right)
\end{equation*}
Comparing the two representations of the
first and second derivatives for $t\ne0$
and extending by continuity to $t=0$ then gives
\begin{equation}
 \E\,[Z_j e^{itQ_a}]=\varphi_a(t)m_j,\qquad
 \E\,[Z_jZ_k e^{itQ_a}]
 =\varphi_a(t)
 \left(m_jm_k+\frac{2\delta_{jk}}{D_j^2}\right)
 \label{eq:num-cf-moments}
\end{equation}
These identities hold for arbitrary real
coefficient vectors, including vectors
with negative or zero entries.

\subsection{Fourier representations of the gradient and Hessian}

We use the Fourier transform convention
\begin{equation*}
 \widehat f(t)=\int_\R e^{itx}f(x)\dd x
\end{equation*}
Substituting $y=e^x$ in the Fourier transform
of $\sig'$ and applying the beta integral
and gamma reflection formula
\citep[Equations~5.12.3 and~5.5.3]{dlmf}
yields
\begin{align*}
 \widehat{\sig'}(t)
 =\int_0^\infty\frac{y^{it}}{(1+y)^2}\dd y
 &=B(1+it,1-it)\\
 &=\Gamma(1+it)\Gamma(1-it)\\&
 =\frac{\pi t}{\sinh(\pi t)}
\end{align*}
in which the expression at $t=0$ is interpreted
by continuity. Both $\sig'$ and its Fourier
transform are integrable, so that Fourier
inversion applies pointwise. Defining
$K(t)=t/\sinh(\pi t)$ with $K(0)=1/\pi$
and using the evenness of $\sig'$ yields,
after a change of  variable
\begin{equation}
 \sig'(bx)=\frac1b\int_0^\infty
 K(t/b)\cos(tx)\dd t,\qquad b>0
 \label{eq:num-logistic-fourier}
\end{equation}
Multiplying by $b$ and integrating  from zero to $x$, using
$\sig(0)=1/2$, yields
\begin{equation}
 \sig(bx)-\frac12
 =\int_0^\infty K(t/b)\frac{\sin(tx)}{t}\dd t
 \label{eq:num-logistic-sine}
\end{equation}
Since $\E\,[Z_j]=0$, multiplication of
\eqref{eq:num-logistic-sine} by $Z_j$,
with $x=Q_a$, and taking expectations furnishes
\begin{align}
 M_j(a,b):=\E\,[Z_j\sig(bQ_a)]\notag&=\int_0^\infty\frac{K(t/b)}{t}
 \operatorname{Im}[\varphi_a(t)m_j]\dd t\notag\\&=2a_j\int_0^\infty K(t/b)
 \operatorname{Re}\left(\frac{\varphi_a(t)}{D_j}\right)
 \dd t
 \label{eq:num-moment}
\end{align}
in which  the second equality follows from the
first identity in \eqref{eq:num-cf-moments}
and the third follows from the definition
of $m_j$. The interchange of expectation
and integration is justified by the bound
$|\sin(tQ_a)|\leq t|Q_a|$ near $t=0$,
the integrability of $|Z_jQ_a|$ and the
exponential decay of $K(t/b)$ as $t\to\infty$.

In the retained coefficient variables
introduced in Section~\ref{sec:experiments},
let
\begin{equation*}
 J(a)=\frac{1}{\alpha_p}
 {\CE}_\tau\bigl(\alpha_p\diag(a,0)\bigr)
\end{equation*}
for $a=(a_1,\ldots,a_p)$ and the remaining
$d-p$ student coefficients are zero.
Differentiating the scaled population
objective with respect to the retained
coefficients yields
\begin{equation*}
 g_j=\partial_{a_j}J(a)
 =M_j(a,\alpha_p)-M_j(\lambda,\tau)
\end{equation*}
and
\begin{equation*}
 B_{jk}=\partial_{a_k}\partial_{a_j}J(a)
 =\alpha_p\E\,[Z_jZ_k\sig'(\alpha_pQ_a)]
\end{equation*}
for $1\leq j,k\leq p$. Substituting
\eqref{eq:num-logistic-fourier} and the
second identity in \eqref{eq:num-cf-moments}
into the Hessian expression leads to 
\begin{equation}
 B_{jk}=\int_0^\infty K(t/\alpha_p)
 \operatorname{Re}\left[
 \varphi_a(t)\left(
 m_jm_k+\frac{2\delta_{jk}}{D_j^2}
 \right)\right]\dd t
 \label{eq:num-hessian}
\end{equation}
The interchange of expectation and integration
is justified by the integrability of
$|Z_jZ_k|$ and $K(t/\alpha_p)$.

The difference between the unscaled
population objectives associated with
two retained coefficient vectors $a$ and
$a'$ can also be evaluated without computing
either objective separately. Indeed, by
the fundamental theorem of calculus
\begin{equation}
 \alpha_p[J(a')-J(a)]
 =\alpha_p\int_0^1
 g(a+v(a'-a))^\top(a'-a)\dd v
 \label{eq:num-objective-difference}
\end{equation}
The implementation uses a 12-node
Gauss-Legendre rule to evaluate this
integral as an independent check of the
difference between the directly computed
population regrets.

Let $\Delta_{\rm dir}$ denote the difference
between the fully optimised and explicitly
calibrated population regrets computed by
direct subtraction, and let $\Delta_{\rm path}$
denote the same difference computed using
\eqref{eq:num-objective-difference}.
Furthermore, let $e_{\rm N}$ denote the
reported local Newton error estimate relative
to the predicted population regret, and let
$\epsilon_{\rm mach}$ denote machine epsilon.
The empirical numerical resolution proxy used
in Figure~\ref{fig:num-optimality} is
\begin{equation}
 \max\left\{
 \frac{10|\Delta_{\rm dir}-\Delta_{\rm path}|}
 {\widehat{\mathcal E}_{p,\tau}},
 \ 10\epsilon_{\rm mach},\ e_{\rm N}
 \right\}
 \label{eq:num-resolution-proxy}
\end{equation}
This quantity provides a heuristic diagnostic
of the numerical resolution of the optimality
gaps rather than a certified error bound.
The same proxy is employed for the
scalar-optimised gaps, for which no separate
gradient-path evaluation of the difference
from the fully optimised population regret
is performed.

\subsection{Direct evaluation of the population regret}

For direct evaluation of the population
regret, let $u=\tau\lambda$ denote the teacher
coefficient vector, let $w$ denote the full
student coefficient vector including its
zero entries and define
\begin{equation*}
 \delta=w-u,\qquad c(v)=u+v\delta
\end{equation*}
The teacher and student logits are therefore
$Q_u$ and $Q_w$, respectively. The population
regret associated with the student
$S=\diag(w)$ is
\begin{equation*}
 {\CE}_\tau(S)-H_\tau
 =\E\,[\psi(Q_w)-\psi(Q_u)-\psi'(Q_u)(Q_w-Q_u)]
\end{equation*}
in which $\psi(x)=\log(1+e^x)$.
Since $\psi''=\sig'$, Taylor's theorem
with integral remainder yields
\begin{equation}
 {\CE}_\tau(S)-H_\tau
 =\int_0^1(1-v)
 \E\,[Q_\delta^2\sig'(Q_{c(v)})]\dd v
 \label{eq:num-direct-kl}
\end{equation}
which avoids subtraction of the teacher
Bayes risk from the student population loss.

For a given $v$ with $c(v)\ne0$, set 
$b=\|c(v)\|_2$ and $a=c(v)/b$, so that
$Q_{c(v)}=bQ_a$. The second identity in
\eqref{eq:num-cf-moments} results in
\begin{equation*}
 \E\,[Q_\delta^2e^{itQ_a}]
 =\varphi_a(t)
 \left\{
 \left(\sum_j\delta_jm_j\right)^2
 +2\sum_j\frac{\delta_j^2}{D_j^2}
 \right\}
\end{equation*}
so substituting the Fourier representation
\eqref{eq:num-logistic-fourier} and
interchanging expectation and integration
therefore yields
\begin{equation}
 \E\,[Q_\delta^2\sig'(Q_{c(v)})]
 =\frac1b\int_0^\infty K(t/b)
 \operatorname{Re}\left[
 \varphi_a(t)\left\{
 \left(\sum_j\delta_jm_j\right)^2
 +2\sum_j\frac{\delta_j^2}{D_j^2}
 \right\}\right]\dd t
 \label{eq:num-weighted}
\end{equation}
in which we define $D_j=1-2ita_j$ and $m_j=2ita_j/D_j$.
The interchange is justified by the finite
second moment of $Q_\delta$ and the
integrability of the Fourier kernel.
For $b=0$, the expectation on the left-hand
side of \eqref{eq:num-weighted} reduces to
$\E\,[Q_\delta^2]/4=\|\delta\|_2^2/2$.

The expectation in the regular prediction
$P_{\rm reg}$ defined in
\eqref{eq:num-predictions} is evaluated by
setting $a=\lambda$ and $b=\tau$ in
\eqref{eq:num-logistic-fourier}, which gives
\begin{equation}
 \E\,[\sig'(\tau F)]
 =\frac1\tau\int_0^\infty
 K(t/\tau)\operatorname{Re}\varphi_\lambda(t)
 \dd t
 \label{eq:num-regular-fourier}
\end{equation}
The central density of the finite-dimensional
teacher signal is obtained by Fourier
inversion
\begin{equation}
 \rho_d(0)=\frac1\pi\int_0^\infty
 \operatorname{Re}\varphi_\lambda(t)\dd t
 \label{eq:num-density-fourier}
\end{equation}
provided that $\varphi_\lambda$ is integrable.
As shown below, this condition is satisfied
for the teacher spectra used in the
numerical experiments.

\subsection{Frequency truncation and numerical quadrature}

The truncation error in the density integral
\eqref{eq:num-density-fourier} can be controlled
using the decay of the characteristic function.
The modulus of \eqref{eq:num-cf} yields
\begin{equation*}
 |\varphi_a(t)|
 =\prod_j(1+4t^2a_j^2)^{-1/4}
\end{equation*}
and for $t>0$, define the logarithmic decay
rate of the characteristic function by
\begin{equation*}
 k(t)=-\frac{\mathrm d\log|\varphi_a(t)|}
 {\mathrm d\log t}
 =\frac12\sum_j\frac{4t^2a_j^2}{1+4t^2a_j^2}
\end{equation*}
Since every summand is nondecreasing in
$t$, integration of the logarithmic
derivative from $T>0$ to $t\geq T$ yields
\begin{equation*}
 \log\frac{|\varphi_a(t)|}{|\varphi_a(T)|}
 =-\int_T^t k(u)\frac{\dd u}{u}
 \leq-k(T)\log(t/T)
\end{equation*}
Exponentiation results in the pointwise bound
\begin{equation*}
 |\varphi_a(t)|\leq
 |\varphi_a(T)|(t/T)^{-k(T)},\qquad t\geq T
\end{equation*}
and integrating over $[T,\infty)$ when $k(T)>1$
we obtain
\begin{equation}
 \int_T^\infty|\varphi_a(t)|\dd t
 \leq\frac{T|\varphi_a(T)|}{k(T)-1}
 \label{eq:num-fourier-tail}
\end{equation}
If $n_a$ denotes the number of nonzero
entries of $a$, then $k(T)\to n_a/2$
as $T\to\infty$. As a result,  if $n_a>2$,
the characteristic function is absolutely
integrable and \eqref{eq:num-fourier-tail}
applies for sufficiently large $T$. This
condition is satisfied by the teacher
spectra used in the density calculations.
For $a=\lambda$, the resulting truncation
error bound for the density integral is
\begin{equation}
 \left|
 \rho_d(0)-\frac1\pi\int_0^T
 \operatorname{Re}\varphi_\lambda(t)\dd t
 \right|
 \leq\frac{T|\varphi_\lambda(T)|}
 {\pi(k(T)-1)}
 \label{eq:num-density-tail}
\end{equation}
provided that $k(T)>1$. The bound controls
the error arising from frequency truncation
but not the quadrature error over the
finite interval $[0,T]$.

For the logistic-weighted Fourier integrals,
the additional exponential decay of the
kernel $K(t/b)$ also permits the use of
a truncation criterion based on the
magnitude of this kernel. The frequency
integrals are evaluated by composite
16-node Gauss-Legendre quadrature on
uniform panels, with panel spacing not
exceeding
\begin{equation*}
 \frac{1}{\ell}
 \min\left\{
 \frac{4}{1+\sum_j|a_j|},
 \max(10^{-6},b/3)
 \right\}
\end{equation*}
in which  $\ell$ is the frequency refinement
level and $b=\infty$ is used for the density
integral. The characteristic function
is evaluated through its logarithm to
reduce cancellation of the linear phase
at small frequencies. A fixed frequency
grid is employed during optimisation
and the gradient at the resulting
coefficients is recomputed using a
smaller truncation tolerance and twice
the frequency refinement level.

\subsection{Independent numerical verification}
\label{app:num-independent}

To assess the accuracy of the Fourier
integral representations independently
of the frequency quadrature procedure,
we consider a teacher with eight equal
eigenvalues, of which four are retained
by the student. The teacher inverse
temperature is $\tau=0.7$ and the student
multiplier is $0.4$. In this case the
retained and omitted quadratic forms
can be expressed in terms of independent
centred chi-squared random variables.
The central density of the teacher
signal is obtained directly from the
chi-squared density, while the population
regret, teacher Bayes risk and sigmoid
moment are evaluated by tensor-product
generalised Gauss-Laguerre quadrature
with respect to the independent
chi-squared distributions of the retained
and omitted sums. The results of these
comparisons with the corresponding
Fourier calculations are reported in
Table~\ref{tab:num-validation}.
The independent reference quadrature
provides a check of the Fourier integral
representations and their numerical
evaluation but is not exact for the
nonpolynomial integrands.


\begin{thebibliography}{99}
\bibitem[Amari(2016)]{amari2016}
\textsc{S.-I. Amari}, \emph{Information Geometry and Its Applications}, Springer, 2016.

\bibitem[Arjevani et~al.(2026)]{arjevani2026}
\textsc{Y. Arjevani, J. Bruna, J. Kileel,
E. Polak, and M. Trager},
\emph{Geometry and optimization of shallow
polynomial networks},
SIAM J. Appl. Algebra Geom., 10 (2026),
pp. 174--209,
\url{https://doi.org/10.1137/25M1732994}.

\bibitem[Bahri et~al.(2024)]{bahri2024}
\textsc{Y. Bahri, E. Dyer, J. Kaplan, J. Lee, and U. Sharma}, \emph{Explaining neural scaling laws}, Proc. Natl. Acad. Sci. USA, 121 (2024), e2311878121, \url{https://doi.org/10.1073/pnas.2311878121}.

\bibitem[Banerjee et~al.(2005)]{banerjee2005}
\textsc{A.~Banerjee, S.~Merugu, I.~S. Dhillon, and J.~Ghosh}, \emph{Clustering with Bregman divergences}, J. Mach. Learn. Res., 6 (2005), pp. 1705--1749, \url{https://www.jmlr.org/papers/v6/banerjee05b.html}.

\bibitem[Barkeshli et~al.(2026)]{barkeshli2026}
\textsc{M. Barkeshli, A. Alfarano, and A. Gromov}, \emph{On the origin of neural scaling laws: from random graphs to natural language}, in Proceedings of the 43rd International Conference on Machine Learning, PMLR 306, 2026, pp. 6704--6735, \url{https://proceedings.mlr.press/v306/barkeshli26a.html}.

\bibitem[Ben Arous et~al.(2025)]{benarous2025}
\textsc{G. Ben Arous, M.~A. Erdogdu, N.~M. Vural, and D. Wu}, \emph{Learning quadratic neural networks in high dimensions: SGD dynamics and scaling laws}, in Advances in Neural Information Processing Systems, 38, 2025, \url{https://proceedings.nips.cc/paper_files/paper/2025/hash/d7ce06e9293c3d8e6cb3f80b4157f875-Abstract-Conference.html}.

\bibitem[Bhatia(1997)]{bhatia1997}
\textsc{R. Bhatia}, \emph{Matrix Analysis}, Springer, New York, Chapter III, Corollary III.1.5, 1997, \url{https://doi.org/10.1007/978-1-4612-0653-8}.

\bibitem[Billingsley(1995)]{billingsley1995}
\textsc{P. Billingsley}, \emph{Probability and Measure}, 3rd ed., John Wiley \& Sons, 1995.

\bibitem[Blondel et~al.(2020)]{blondel2020}
\textsc{M. Blondel, A.~F.~T. Martins, and V. Niculae}, \emph{Learning with Fenchel--Young losses}, J. Mach. Learn. Res., 21(35) (2020), pp. 1--69, \url{https://www.jmlr.org/papers/v21/19-021.html}.

\bibitem[Bordelon et~al.(2024)]{bordelon2024}
\textsc{B. Bordelon, A. Atanasov, and C. Pehlevan}, \emph{A dynamical model of neural scaling laws}, in Proceedings of the 41st International Conference on Machine Learning, PMLR 235, 2024, pp. 4345--4382, \url{https://proceedings.mlr.press/v235/bordelon24a.html}.

\bibitem[Bordelon et~al.(2025)]{bordelon2025}
\textsc{B. Bordelon, A. Atanasov, and C. Pehlevan}, \emph{How feature learning can improve neural scaling laws}, in International Conference on Learning Representations, 2025, \url{https://proceedings.iclr.cc/paper_files/paper/2025/hash/8092824cd98b783ccba168446141d822-Abstract-Conference.html}.

\bibitem[Borell(1975)]{borell1975}
\textsc{C. Borell}, \emph{The Brunn--Minkowski inequality in Gauss space}, Invent. Math., 30 (1975), pp. 207--216, \url{https://doi.org/10.1007/BF01425510}.

\bibitem[Cagnetta et~al.(2026)]{cagnetta2026}
\textsc{F. Cagnetta, A. Ravent\'{o}s, S. Ganguli, and M. Wyart}, \emph{Deriving neural scaling laws from the statistics of natural language}, in Proceedings of the 43rd International Conference on Machine Learning, PMLR 306, 2026, pp. 10450--10480, \url{https://proceedings.mlr.press/v306/cagnetta26a.html}.

\bibitem[Carbery and Wright(2001)]{carbery2001}
\textsc{A. Carbery and J. Wright}, \emph{Distributional and $L^q$ norm inequalities for polynomials over convex bodies in $\R^n$}, Math. Res. Lett., 8(3) (2001), pp. 233--248, \url{https://doi.org/10.4310/MRL.2001.v8.n3.a1}.

\bibitem[Chen and Dai(2011)]{chen2011}
\textsc{X. Chen and W. Dai}, \emph{Maximum entropy principle for uncertain variables}, Int. J. Fuzzy Syst., 13 (2011), pp. 232--236.

\bibitem[Chizat et~al.(2019)]{chizat2019}
\textsc{L. Chizat, E. Oyallon, and F. Bach}, \emph{On lazy training in differentiable programming}, in Advances in Neural Information Processing Systems, 32, 2019, \url{https://papers.neurips.cc/paper/2019/hash/ae614c557843b1df326cb29c57225459-Abstract.html}.

\bibitem[Defilippis et~al.(2025)]{defilippis2025spectra}
\textsc{L. Defilippis, Y. Xu, J. Girardin, E. Troiani, V. Erba, L. Zdeborov\'a, B. Loureiro, and F. Krzakala}, \emph{Scaling laws and spectra of shallow neural networks in the feature learning regime}, preprint, arXiv:2509.24882, 2025, \url{https://arxiv.org/abs/2509.24882}.

\bibitem[Defilippis et~al.(2026)]{defilippis2026}
\textsc{L. Defilippis, F. Krzakala, B. Loureiro, and A. Maillard}, \emph{Optimal scaling laws in learning hierarchical multi-index models}, preprint, arXiv:2602.05846, 2026, \url{https://arxiv.org/abs/2602.05846}.

\bibitem[Dhillon and Tropp(2007)]{dhillon2007}
\textsc{I.~S. Dhillon and J.~A. Tropp}, \emph{Matrix nearness problems with Bregman divergences}, SIAM J. Matrix Anal. Appl., 29(4) (2007), pp. 1120--1146, \url{https://doi.org/10.1137/060649021}.

\bibitem[Drissi et~al.(2008)]{drissi2008}
\textsc{N. Drissi, T. Chonavel, and J.-M. Boucher}, \emph{Generalized cumulative residual entropy for distributions with unrestricted supports}, Res. Lett. Signal Process., 2008 (2008), 790607, \url{https://doi.org/10.1155/2008/790607}.

\bibitem[Du and Lee(2018)]{dulee2018}
\textsc{S. Du and J. Lee}, \emph{On the power of over-parametrization in neural networks with quadratic activation}, in Proceedings of the 35th International Conference on Machine Learning, PMLR 80, 2018, pp. 1329--1338, \url{https://proceedings.mlr.press/v80/du18a.html}.

\bibitem[Evans and Gariepy(2015)]{evansgariepy2015}
\textsc{L.~C. Evans and R.~F. Gariepy}, \emph{Measure Theory and Fine Properties of Functions}, revised ed., CRC Press, Section 3.4 (Coarea Formula), 2015.

\bibitem[Gamarnik et~al.(2025)]{gamarnik2025}
\textsc{D. Gamarnik, E.~C. K\i z\i lda\u g, and I. Zadik}, \emph{Stationary points of a shallow neural network with quadratic activations and the global optimality of the gradient descent algorithm}, Math. Oper. Res., 50(1) (2025), pp. 209--251, \url{https://doi.org/10.1287/moor.2021.0082}.

\bibitem[Gasparin et~al.(2025)]{gasparin2025}
\textsc{M. Gasparin, B. Scarpa, and E. Stanghellini}, \emph{Omitting continuous covariates in binary regression models: implications for sensitivity and mediation analysis}, Statist. Neerlandica, 79 (2025), e12369, \url{https://doi.org/10.1111/stan.12369}.

\bibitem[Girardin et~al.(2026)]{girardin2026}
\textsc{J. Girardin, E. Troiani, Y. Xu, V. Erba, F. Krzakala, and L. Zdeborov\'{a}}, \emph{How width and data shape generalization scaling laws in quadratic neural networks}, preprint, arXiv:2606.28242, 2026, \url{https://arxiv.org/abs/2606.28242}.

\bibitem[Gneiting and Raftery(2007)]{gneiting2007}
\textsc{T.~Gneiting and A.~E. Raftery}, \emph{Strictly proper scoring rules, prediction, and estimation}, J. Amer. Statist. Assoc., 102(477) (2007), pp. 359--378, \url{https://doi.org/10.1198/016214506000001437}.

\bibitem[Hoffmann et~al.(2022)]{hoffmann2022}
\textsc{J. Hoffmann, S. Borgeaud, A. Mensch, E. Buchatskaya, T. Cai, E. Rutherford, D. de Las Casas, L. A. Hendricks, J. Welbl, A. Clark, T. Hennigan, E. Noland, K. Millican, G. van den Driessche, B. Damoc, A. Guy, S. Osindero, K. Simonyan, E. Elsen, O. Vinyals, J. Rae, and L. Sifre}, \emph{Training compute-optimal large language models}, in Advances in Neural Information Processing Systems, 35, 2022, pp. 30016--30030, \url{https://proceedings.neurips.cc/paper_files/paper/2022/hash/c1e2faff6f588870935f114ebe04a3e5-Abstract.html}.

\bibitem[Jacot et~al.(2018)]{jacot2018}
\textsc{A. Jacot, F. Gabriel, and C. Hongler}, \emph{Neural tangent kernel: convergence and generalization in neural networks}, in Advances in Neural Information Processing Systems, 31, 2018, \url{https://papers.neurips.cc/paper/2018/hash/5a4be1fa34e62bb8a6ec6b91d2462f5a-Abstract.html}.

\bibitem[Janson(1997)]{janson1997}
\textsc{S. Janson}, \emph{Gaussian Hilbert Spaces}, Cambridge University Press, 1997, \url{https://doi.org/10.1017/CBO9780511526169}.

\bibitem[Kane(2010)]{kane2010}
\textsc{D.~M. Kane}, \emph{The Gaussian surface area and noise sensitivity of degree-$d$ polynomial threshold functions}, in Proceedings of the 25th Annual IEEE Conference on Computational Complexity, 2010, \url{https://arxiv.org/abs/0912.2709}.

\bibitem[Kane(2014)]{kane2014}
\textsc{D.~M. Kane}, \emph{The correct exponent for the Gotsman--Linial conjecture}, Comput. Complexity, 23 (2014), pp. 151--175, \url{https://arxiv.org/abs/1210.1283}.

\bibitem[Kaplan et~al.(2020)]{kaplan2020}
\textsc{J. Kaplan, S. McCandlish, T. Henighan, T.~B. Brown, B. Chess, R. Child, S. Gray, A. Radford, J. Wu, and D. Amodei}, \emph{Scaling laws for neural language models}, preprint, arXiv:2001.08361, 2020, \url{https://arxiv.org/abs/2001.08361}.

\bibitem[Klein and Doll(2020)]{klein2020}
\textsc{I. Klein and M. Doll}, \emph{(Generalized) maximum cumulative direct, residual, and paired $\Phi$ entropy approach}, Entropy, 22(1) (2020), 91, \url{https://doi.org/10.3390/e22010091}.

\bibitem[Klein et~al.(2016)]{klein2016}
\textsc{I. Klein, B. Mangold, and M. Doll}, \emph{Cumulative paired $\phi$-entropy}, Entropy, 18(7) (2016), 248, \url{https://doi.org/10.3390/e18070248}.

\bibitem[Kulis et~al.(2009)]{kulis2009}
\textsc{B. Kulis, M.~A. Sustik, and I.~S. Dhillon}, \emph{Low-rank kernel learning with Bregman matrix divergences}, J. Mach. Learn. Res., 10 (2009), pp. 341--376, \url{https://jmlr.org/papers/v10/kulis09a.html}.

\bibitem[Ledoux(2001)]{ledoux2001}
\textsc{M. Ledoux}, \emph{The Concentration of Measure Phenomenon}, American Mathematical Society, 2001.

\bibitem[Lewis(1996)]{lewis1996}
\textsc{A.~S. Lewis}, \emph{Derivatives of spectral functions}, Math. Oper. Res., 21(3) (1996), pp. 576--588, \url{https://doi.org/10.1287/moor.21.3.576}.

\bibitem[Li et~al.(2019)]{li2019}
\textsc{Q. Li, Z. Zhu, and G. Tang}, \emph{The non-convex geometry of low-rank matrix optimization}, Inf. Inference, 8(1) (2019), pp. 51--96, \url{https://doi.org/10.1093/imaiai/iay003}.

\bibitem[Martin et~al.(2024)]{martin2024}
\textsc{S. Martin, F. Bach, and G. Biroli}, \emph{On the impact of overparameterization on the training of a shallow neural network in high dimensions}, Proc. Mach. Learn. Res., 238 (2024), pp. 3655--3663, \url{https://proceedings.mlr.press/v238/martin24a.html}.

\bibitem[Martin et~al.(2026)]{martin2026}
\textsc{S. Martin, G. Biroli, and F. Bach}, \emph{High-dimensional analysis of gradient flow for extensive-width quadratic neural networks}, J. Mach. Learn. Res., 27(136) (2026), pp. 1--182, \url{https://www.jmlr.org/beta/papers/v27/26-0672.html}.

\bibitem[NIST]{dlmf}
\textsc{National Institute of Standards and Technology}, \emph{Digital Library of Mathematical Functions}, Chapter 5 (Gamma Function), \url{https://dlmf.nist.gov/5}.

\bibitem[Neuhaus and Jewell(1993)]{neuhaus1993}
\textsc{J.~M. Neuhaus and N.~P. Jewell}, \emph{A geometric approach to assess bias due to omitted covariates in generalized linear models}, Biometrika, 80(4) (1993), pp. 807--815, \url{https://academic.oup.com/biomet/article-abstract/80/4/807/241159}.

\bibitem[Nguyen and Barbier(2026)]{nguyen2026}
\textsc{M.-T. Nguyen and J. Barbier}, \emph{Sharp feature-learning transitions and Bayes-optimal neural scaling laws in extensive-width networks}, preprint, arXiv:2605.10395, 2026, \url{https://arxiv.org/abs/2605.10395}.

\bibitem[Nikolaou et~al.(2026)]{nikolaou2026}
\textsc{K. Nikolaou, J. Scheunemann, S. Krippendorf, S. Tovey, and C. Holm}, \emph{Spectral reach: understanding neural scaling as progress into the spectral tail}, in Proceedings of the 43rd International Conference on Machine Learning, PMLR 306, 2026, pp. 93232--93261, \url{https://proceedings.mlr.press/v306/nikolaou26a.html}.

\bibitem[Pesenti and Vanduffel(2024)]{pesenti2024}
\textsc{S.~M. Pesenti and S. Vanduffel}, \emph{Optimal transport divergences induced by scoring functions}, Oper. Res. Lett., 57 (2024), 107146, \url{https://arxiv.org/abs/2311.12183}.

\bibitem[Reid and Williamson(2010)]{reid2010}
\textsc{M.~D. Reid and R.~C. Williamson}, \emph{Composite binary losses}, J. Mach. Learn. Res., 11 (2010), pp. 2387--2422, \url{https://www.jmlr.org/papers/v11/reid10a.html}.

\bibitem[Reid and Williamson(2011)]{reid2011}
\textsc{M.~D. Reid and R.~C. Williamson}, \emph{Information, divergence and risk for binary experiments}, J. Mach. Learn. Res., 12 (2011), pp. 731--817, \url{https://www.jmlr.org/papers/v12/reid11a.html}.

\bibitem[Santambrogio(2015)]{santambrogio2015}
\textsc{F. Santambrogio}, \emph{Optimal Transport for Applied Mathematicians}, Birkh\"auser, Cham, 2015, \url{https://doi.org/10.1007/978-3-319-20828-2}.

\bibitem[Sudakov and Tsirelson(1978)]{sudakov1978}
\textsc{V.~N. Sudakov and B.~S. Tsirelson}, \emph{Extremal properties of half-spaces for spherically invariant measures}, J. Soviet Math., 9 (1978), pp. 9--18.

\bibitem[Tsybakov(2004)]{tsybakov2004}
\textsc{A.~B. Tsybakov}, \emph{Optimal aggregation of classifiers in statistical learning}, Ann. Statist., 32(1) (2004), pp. 135--166.

\bibitem[Tsybakov(2009)]{tsybakov2009}
\textsc{A.~B. Tsybakov}, \emph{Introduction to Nonparametric Estimation}, Springer, New York, 2009, \url{https://doi.org/10.1007/b13794}.

\bibitem[Villani(2009)]{villani2009}
\textsc{C. Villani}, \emph{Optimal Transport: Old and New}, Springer, Berlin, Heidelberg, 2009, \url{https://doi.org/10.1007/978-3-540-71050-9}.

\bibitem[Wortsman-Zurich et~al.(2026)]{wortsman2026}
\textsc{A. Wortsman-Zurich, H. Tabanelli, Y. Dandi, F. Krzakala, and B. Loureiro}, \emph{Scaling laws from sequential feature recovery: a solvable hierarchical model}, preprint, arXiv:2605.14567, 2026, \url{https://arxiv.org/abs/2605.14567}.

\bibitem[Zeger et~al.(1988)]{zeger1988}
\textsc{S.~L. Zeger, K.-Y. Liang, and P.~S. Albert}, \emph{Models for longitudinal data: a generalized estimating equation approach}, Biometrics, 44(4) (1988), pp. 1049--1060, \url{https://pubmed.ncbi.nlm.nih.gov/3233245/}.

\bibitem[Zou et~al.(2026)]{zou2026}
\textsc{J. Zou, Z. Gong, Y. Su, H. Tang, and Y. Liu}, \emph{Effective frontiers: a unification of neural scaling laws}, preprint, arXiv:2602.02593, 2026, \url{https://arxiv.org/abs/2602.02593}.

\end{thebibliography}
\end{document}